\documentclass[runningheads]{llncs}
\usepackage{graphicx}
\usepackage{comment}
\usepackage{amsmath,amssymb} % define this before the line numbering.
\usepackage{color}

\usepackage{epsfig}
\usepackage{graphicx}
\usepackage{amsmath}
\usepackage{amssymb}
\usepackage{nicefrac}
\usepackage [autostyle, english = american]{csquotes}
\MakeOuterQuote{"}

\usepackage[breaklinks=true,bookmarks=false]{hyperref}
\usepackage{booktabs}
\usepackage{multirow}
\usepackage{bbm}
\usepackage{array}
\usepackage{graphbox}
\usepackage{color}
\usepackage{colortbl}
\usepackage{makecell}
\graphicspath{{images/}}
\usepackage{wrapfig}
\newcommand{\customparagraph}{\textbf}
\newlength{\newl}
\newlength{\newlloss}

\usepackage[linesnumbered,algoruled,boxed,lined]{algorithm2e}
\SetKwProg{Fn}{Function}{}{end}

\newcommand{\norm}[1]{\left\|#1\right\|}

\newcommand{\floor}[1]{\left\lfloor#1\right\rfloor}

\newcommand{\Id}{\mathbbm{1}}

\def\argmax{\mathop{\rm arg\,max}\limits}%    a math operator.
\def\minop{\mathop{\rm min}\limits}
\def\maxop{\mathop{\rm max}\limits}

\def\R{\mathbb{R}}
\def\N{\mathbb{N}}

\def\G{\mathcal{G}}

\def\R{\mathbb{R}}

\def\N{\mathbb{N}}

\def\F{\mathcal{F}}

\newcommand\commentout[1]{}

\makeatletter
\def\blfootnote{\xdef\@thefnmark{}\@footnotetext}
\makeatother

\begin{document}
	% \linenumbers
	\pagestyle{headings}
	\mainmatter
	\def\ECCVSubNumber{4362}  % Insert your submission number here
	
	\title{Square Attack: a query-efficient black-box adversarial attack via random search} % Replace with your title
	
	\begin{comment}
	\titlerunning{ECCV-20 submission ID \ECCVSubNumber} 
	\authorrunning{ECCV-20 submission ID \ECCVSubNumber} 
	\author{Anonymous ECCV submission}
	\institute{Paper ID \ECCVSubNumber}
	\end{comment}
	%******************
	
	%\begin{comment}
	\titlerunning{Square Attack: a query-efficient black-box adversarial attack}
	\author{Maksym Andriushchenko$^*$\inst{1}
		\and
		Francesco Croce$^*$\inst{2}
		\and \\
		Nicolas Flammarion\inst{1}
		\and 
		Matthias Hein \inst{2}}
	
	\authorrunning{M. Andriushchenko et al.}
	\institute{EPFL \and
		University of T{\"u}bingen
	}
	%\end{comment}
	%******************
	\maketitle
	
	\begin{abstract}
		We propose the \textit{Square Attack}, a score-based black-box $l_2$- and $l_\infty$-adversarial attack that does not rely on local gradient information and thus is not affected by gradient masking. Square Attack is based on a randomized search scheme which selects localized square-shaped updates at random positions so that at each iteration the perturbation is situated approximately at the boundary of the feasible set. 
		Our method is significantly more query efficient and achieves a higher success rate compared to the state-of-the-art methods, especially in the untargeted setting. 
		In particular, on ImageNet we improve the average query efficiency in the untargeted setting for various deep networks by a factor of at least $1.8$ and up to $3$ compared to the recent state-of-the-art $l_\infty$-attack of Al-Dujaili \& O’Reilly (2020). Moreover, although our attack is \textit{black-box}, it can also outperform gradient-based \textit{white-box} attacks on the standard benchmarks achieving a new state-of-the-art in terms of the success rate. The code of our attack is available at \url{https://github.com/max-andr/square-attack}.
		
		%\keywords{Adversarial robustness, black-box attacks, deep learning.}
	\end{abstract}
	
	\blfootnote{$^*$Equal contribution.}
	
	\section{Introduction}
	\begin{wrapfigure}{R}{0.4\columnwidth}
		\centering
		%\setlength{\tabcolsep}{0pt}
		%%\begin{tabular}{c}
		\includegraphics[width=0.4\columnwidth]{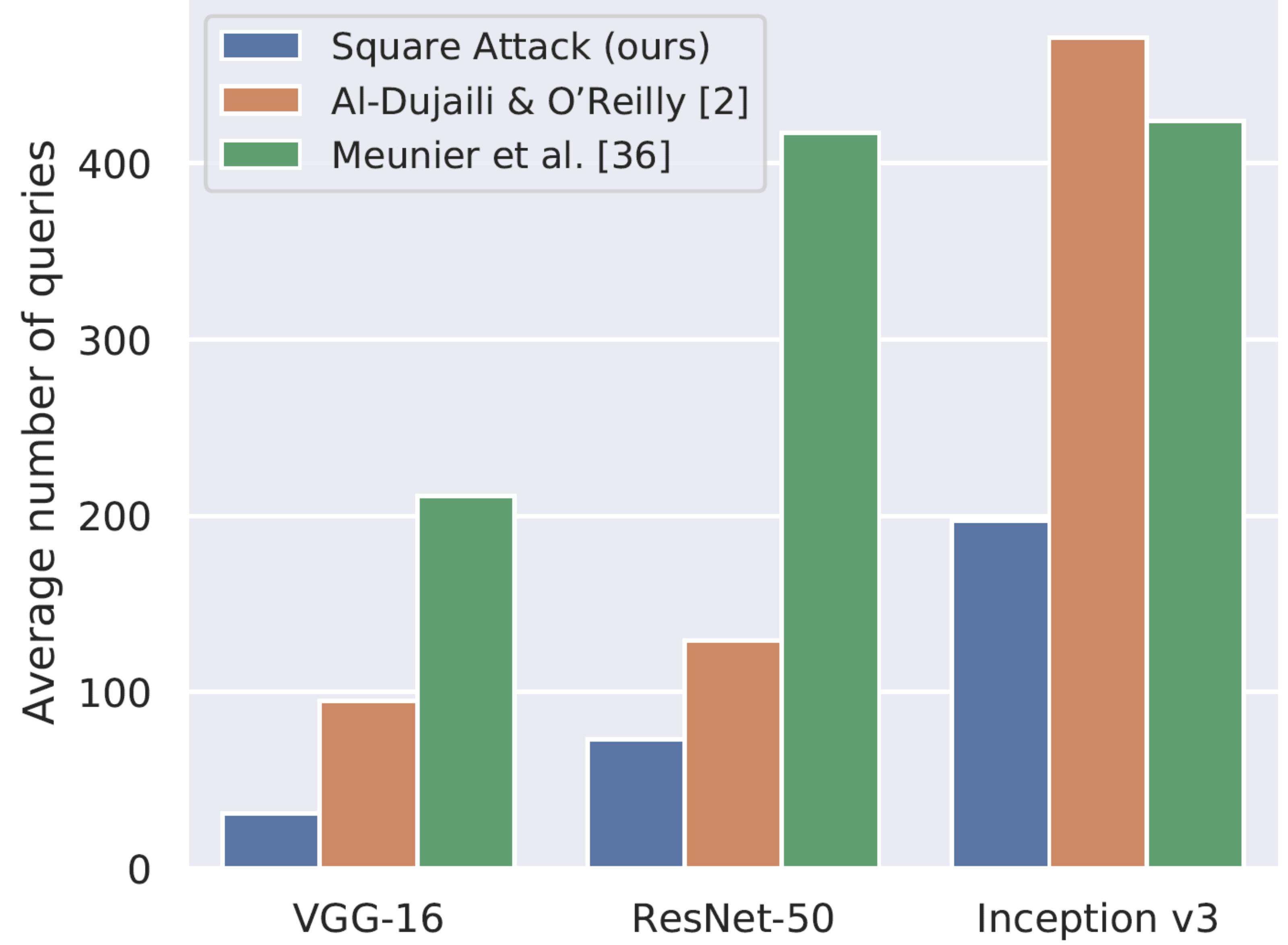} 
		%\includegraphics[height=\linewidth]{bar_plot_teaser_reshaped.pdf} 
		%\end{tabular}
		\caption{%\textbf{Left:} Adversarial example generated for ResNet-50 by the \textit{Square Attack} in the threat model $\norm{\delta}_\infty \leq 0.05$. The $l_\infty$ attack uses as initial perturbation stripes and then does random search using updates with squares. \textbf{Right:} avg.
			Avg. number of queries of successful untargeted $l_\infty$-attacks on three ImageNet models for three score-based black-box attacks. Square Attack outperforms all other attacks
			%regarding query efficiency and success rate.
			by large margin
		}
		\label{fig:teaser_adv_examples}
	\end{wrapfigure}
	\commentout{
		\begin{figure}[t]
			\centering \small\setlength{\newl}{0.22\columnwidth}\setlength{\tabcolsep}{1pt}
			\begin{tabular}{ccc|c}
				\scriptsize \textbf{original} & \scriptsize \textbf{perturbation} & \scriptsize \textbf{adversarial} & \multirow{2}{*}{\includegraphics[align=c, width=0.3\columnwidth]{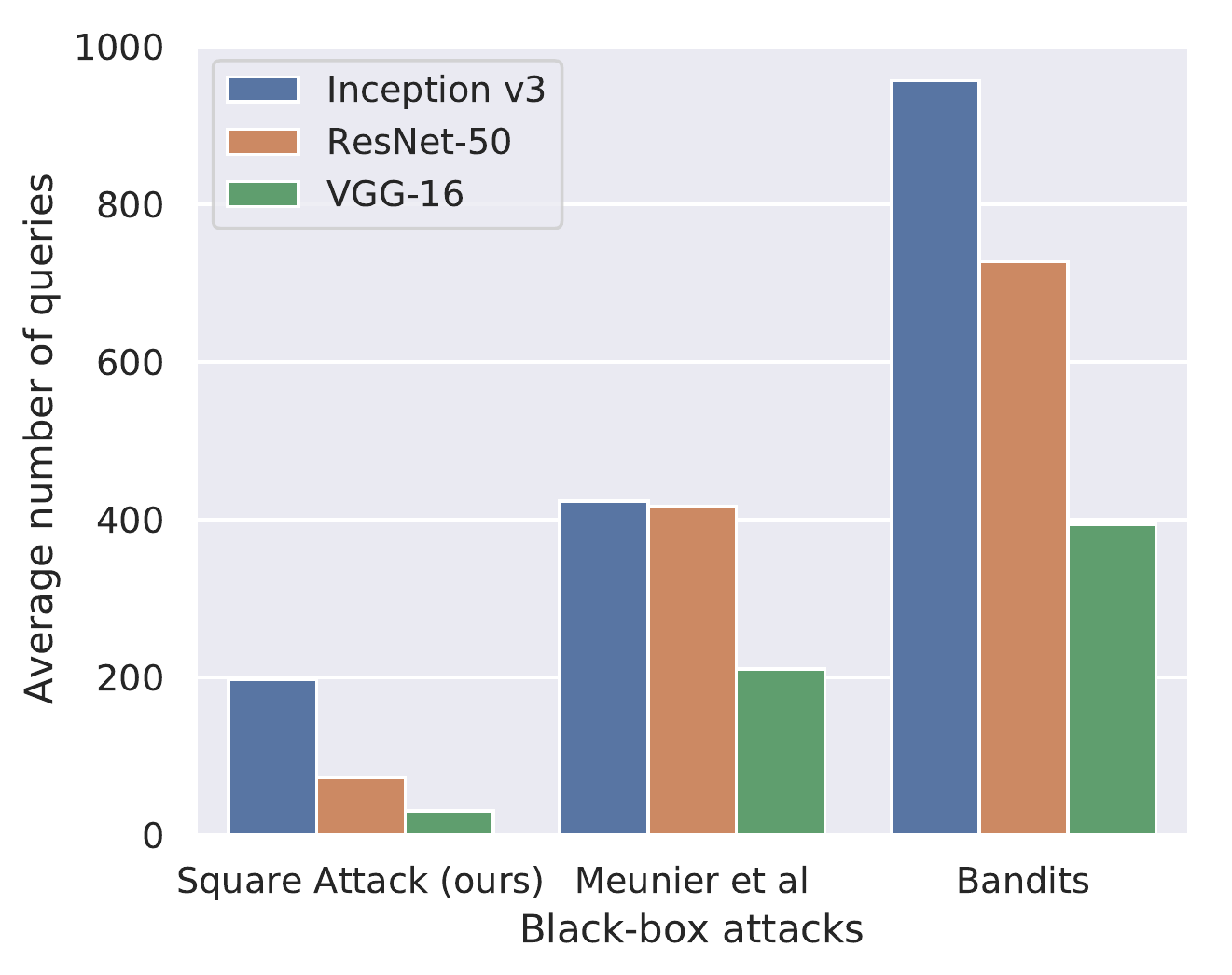}}\\
				\includegraphics[align=c, width=\newl,clip, trim=20mm 10mm 20mm 5mm]{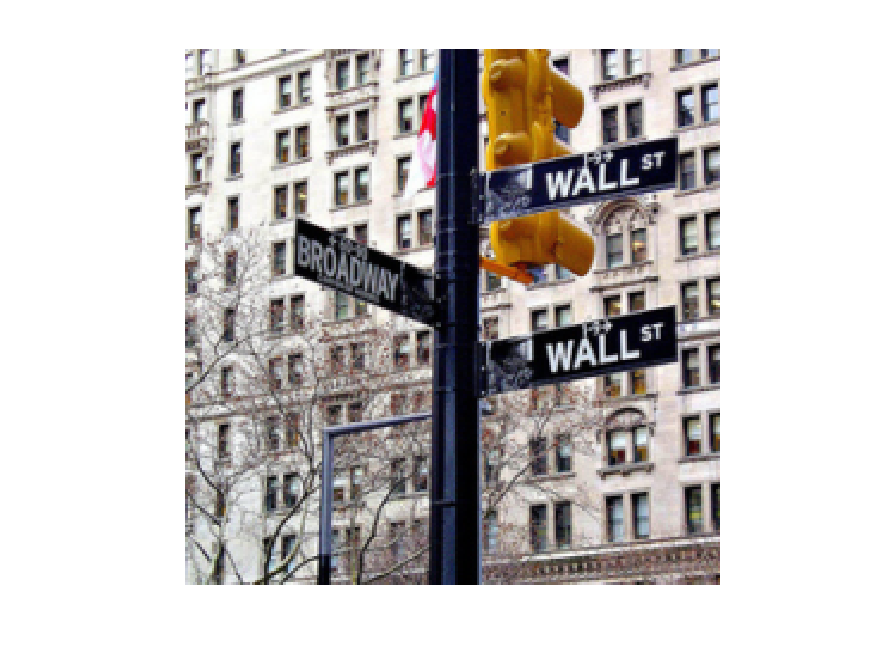} & \includegraphics[align=c, width=\newl,clip, trim=20mm 10mm 20mm 5mm]{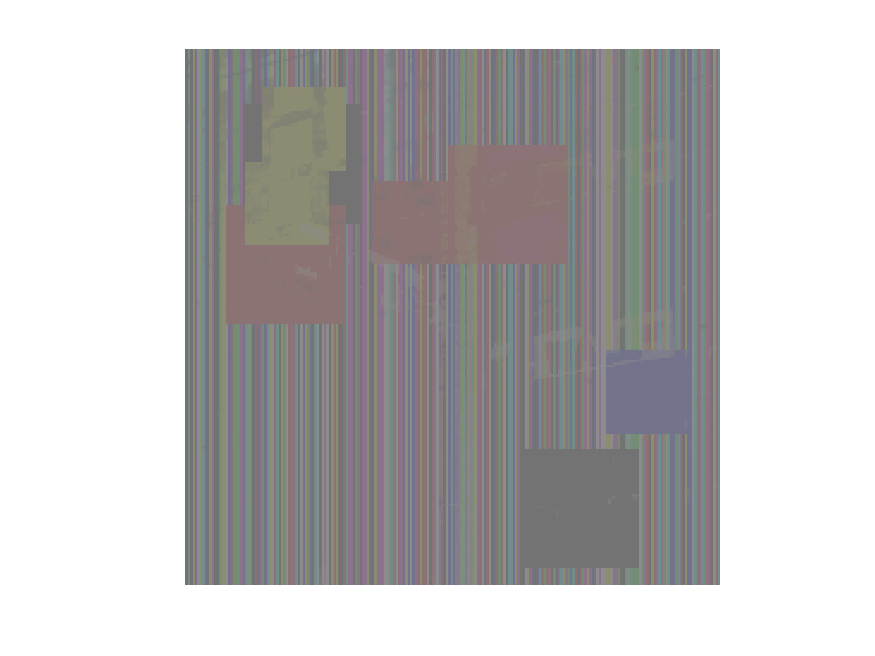}&
				\includegraphics[align=c, width=\newl,clip, trim=20mm 10mm 20mm 5mm]{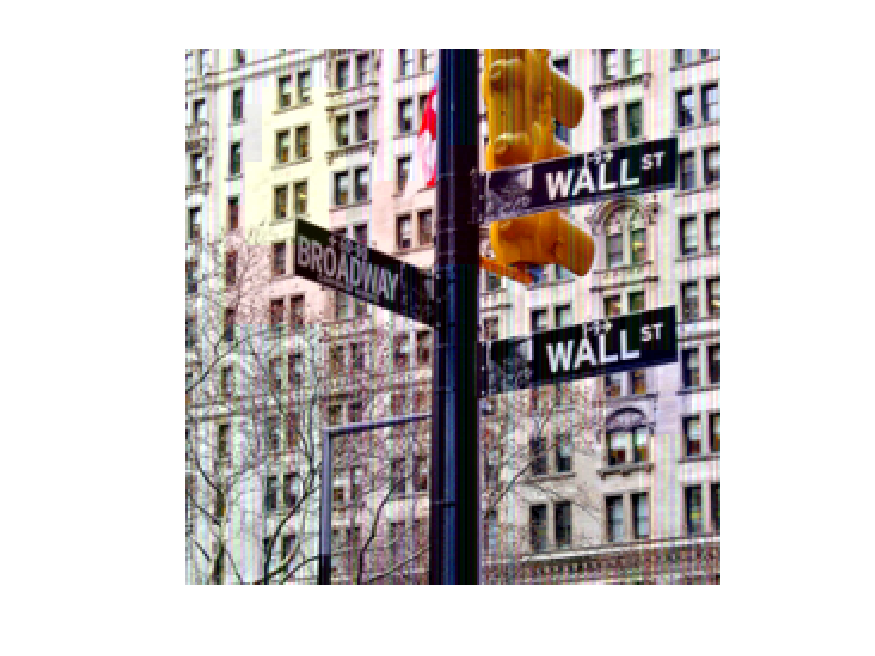} & \\
			\end{tabular}
			\caption{\textbf{Left}: an illustration of the adversarial perturbations produced by the $l_\infty$-Square Attack (first three pictures). \textbf{Rightmost}: comparison of the query consumption of our attack to that of other state-of-the-art attacks. The Square Attack needs from $2$ to $7$ times fewer queries to fool the same network.}%\label{fig:teaser_adv_examples}
	\end{figure}}
	
	Adversarial examples are of particular concern when it comes to applications of machine learning which are safety-critical. Many defenses against adversarial examples have been proposed 
	\cite{GuRig2015,ZheEtAl2016,PapEtAl2016a,BasEtAl2016,madry2018towards,AkhtarARXIV2018,BiggioPR2018} but with limited success, as new more powerful attacks could break many of them \cite{CarWag2017,AthEtAl2018,MosEtAl18,CheEtAl2018,ZheEtAl2019}.
	In particular, gradient obfuscation or masking \cite{AthEtAl2018,MosEtAl18} is often the reason why seemingly robust models turn out to be non-robust in the end. Gradient-based attacks are most often affected by this phenomenon (white-box attacks but also black-box attacks based on finite difference approximations \cite{MosEtAl18}). 
	Thus it is important to have attacks which are based on different principles.
	Black-box attacks have recently become more popular
	\cite{narodytska2017simple,brendel2017decision,SuVarKou19} as their attack strategies are quite different from the ones employed for 
	adversarial training, where often PGD-type attacks \cite{madry2018towards} are used. However, a big weakness currently is that these black-box attacks need to query the classifier too many times before they find adversarial examples, and their success rate is sometimes significantly lower than
	that of white-box attacks.
	
	In this paper we propose Square Attack, a score-based adversarial attack, i.e. it can query the probability distribution over the classes predicted by a classifier but has no access to the underlying model. The Square Attack exploits random search\footnote{It is an iterative procedure different from random sampling inside the feasible region.}
	~\cite{rastrigin1963convergence,schumer1968adaptive} which is one of the simplest approaches for black-box optimization. 
	Due to a particular sampling distribution, it requires significantly fewer queries compared to the state-of-the-art black-box methods (see Fig.~\ref{fig:teaser_adv_examples}) in the score-based threat model while outperforming them
	in terms of \textit{success rate}, i.e. the percentage of successful adversarial examples. 
	This is achieved by a combination of a particular initialization strategy and our square-shaped updates. We motivate why these updates are particularly suited to attack neural networks and provide convergence guarantees for a variant of our method. 
	In an extensive evaluation with untargeted and targeted attacks, three datasets (MNIST, CIFAR-10, ImageNet), normal and robust models, we show that Square Attack outperforms
	state-of-the-art methods in the $l_2$- and $l_\infty$-threat model.

	\section{Related Work}\label{sec:related_work}
	We discuss black-box attacks with $l_2$- and $l_\infty$-perturbations since our attack focuses on this setting.
	Although attacks for other norms, e.g. $l_0$, exist \cite{narodytska2017simple,croce2019sparse}, they are often algorithmically different due to the geometry of the perturbations.
	
	\customparagraph{$l_2$- and $l_\infty$-score-based attacks.}
	Score-based black-box attacks have only access to the score predicted by a classifier for each class for a given input.
	Most of
	such attacks in the literature are based on gradient estimation through finite differences.
	The first papers in this direction \cite{bhagoji2018practical,ilyas2018black,uesato2018adversarial} propose %iterative attacks
	attacks which approximate the gradient by sampling from some noise distribution around the point.
	While this
	approach can be successful,
	it requires many queries of the classifier, particularly in high-dimensional input spaces
	as in image classification. Thus, improved techniques reduce the dimension of the search space via using the principal components of the data \cite{bhagoji2018practical}, searching for perturbations in the latent space of an auto-encoder \cite{tu2019autozoom} or using a low-dimensional noise distribution \cite{ilyas2019prior}.
	Other attacks exploit evolutionary strategies or random search, e.g. \cite{alzantot2018genattack} use a genetic algorithm to generate adversarial examples and %claim to 
	alleviate gradient masking
	as they can
	reduce the robust accuracy on randomization- and discretization-based defenses.
	The $l_2$-attack of \cite{guo2019simple} can be seen as a variant of random search which chooses the search directions in an orthonormal basis and tests up to two candidate updates at each step.
	However, their algorithm can have suboptimal query efficiency since it adds
	at every step
	only small (in $l_2$-norm) modifications,
	and suboptimal updates cannot be undone as they are orthogonal to each other.
	A recent line of work has pursued black-box attacks which are based on the observation that successful adversarial perturbations are attained at corners of the $l_\infty$-ball intersected with the image space $[0,1]^d$ \cite{MooEtAl2019,AlDujaili2019ThereAN,MeuEtAl2019}.
	Searching over the corners allows to apply discrete optimization techniques to generate adversarial attacks, significantly improving the query efficiency.
	Both \cite{MooEtAl2019} and \cite{AlDujaili2019ThereAN} divide the image according to some coarse grid, perform local search in this lower dimensional space allowing componentwise changes only of $-\epsilon$ and $\epsilon$, then refine the grid and repeat the scheme. In \cite{AlDujaili2019ThereAN} such a procedure is motivated as an estimation of the gradient signs.
	Recently, \cite{MeuEtAl2019} proposed several attacks based on %different
	evolutionary algorithms,
	using discrete and continuous optimization, achieving nearly state-of-the-art query efficiency for the $l_\infty$-norm. 
	In order to reduce the dimensionality of the search space, they use the ``tiling trick'' of \cite{ilyas2019prior} where they divide the perturbation into a set of squares and modify the values in these squares with evolutionary algorithms. 
	A related idea also appeared earlier in \cite{fawzi2016measuring} where they introduced black rectangle-shaped perturbations for generating adversarial occlusions.
	In \cite{MeuEtAl2019}, as in \cite{ilyas2019prior}, both size and position of the squares are fixed at the beginning and not optimized.
	Despite their effectiveness for the $l_\infty$-norm, these discrete optimization based attacks are not straightforward to adapt to the $l_2$-norm.
	Finally, approaches based on Bayesian optimization exist, e.g. \cite{shukla2019blackbox}, %combine it with the ``tiling trick'',
	but show competitive performance only in a low-query regime.

	\customparagraph{Different threat and knowledge models.}
	We focus  on \textit{$l_p$-norm-bounded} adversarial perturbations (for other perturbations such as rotations, translations, occlusions in the black-box setting see, e.g., \cite{fawzi2016measuring}). 
	Perturbations with \textit{minimal} $l_p$-norm
	are considered in \cite{chen2017zoo,tu2019autozoom}
	but require significantly more queries than norm-bounded ones. Thus we do not compare to them, except for \cite{guo2019simple} which has competitive query efficiency while aiming at small perturbations. 
	
	In other cases the attacker has a different knowledge of the classifier.
	A more restrictive scenario, considered by
	\textit{decision-based} attacks \cite{brendel2017decision,cheng2018query,guo2018low,brunner2018guessing,chen2019boundary}, is when the attacker can query only the decision of the classifier, but not the predicted scores.
	Other works use more permissive threat models,
	e.g., when the attacker already has a substitute model similar to the target one \cite{papernot2016transferability,yan2019subspace,cheng2019improving,du2019query,suya2019hybrid} and thus can generate adversarial examples for the substitute model and then transfer them to the target model.
	Related to this, \cite{yan2019subspace} suggest to refine this approach by running a black-box gradient estimation attack in a subspace spanned by the gradients of substitute models.
	However, the gain in query efficiency given by such extra knowledge does not account for the computational cost required to train the substitute models, particularly high on ImageNet-scale. Finally, \cite{li2019nattack} use extra information on the target data distribution to train a model that predicts adversarial images that are then refined by gradient estimation attacks.

	\section{Square Attack}
	\label{sec:attack}
	In the following we recall the definitions of the adversarial examples in the threat model we consider and 
	present our black-box attacks for the $l_\infty$- and $l_2$-norms.
	
	\subsection{Adversarial Examples in the $l_p$-threat Model}
	Let $f : [0,1]^d \rightarrow \R^K$ be a classifier, where $d$ is the input dimension, $K$ the number of classes and $f_k(x)$ is the predicted score that $x$ belongs to class $k$. The classifier assigns class $\argmax_{k=1,\ldots,K} f_k(x)$ to the input $x$. 
	The goal of an \textit{untargeted} attack is to change the correctly predicted class $y$ for the point $x$. A point $\hat{x}$ is called an \textit{adversarial example} with an $l_p$-norm bound of $\epsilon$ for $x$ if \[ \argmax_{k=1,\ldots,K} f_k(\hat{x}) \neq y, \quad \norm{\hat{x}- x}_p\leq \epsilon \quad\textrm{and} \quad \hat{x} \in [0,1]^d,\]
	where we have added the additional constraint that $\hat{x}$ is an image.  
	The task of finding $\hat{x}$ can be rephrased as solving the constrained optimization problem
	\begin{equation}
	\label{eq:opt_problem}
	\minop_{\hat{x} \in [0,1]^d} L(f(\hat{x}), y),\quad \textrm{s.t.} \quad \norm{\hat{x} - x}_p \leq \epsilon,
	\end{equation}
	for a loss $L$. In our experiments, we use
	$L(f(\hat{x}), y) = f_y(\hat{x}) - \max_{k \neq y} f_k(\hat{x})$.
	
	The goal of \textit{targeted} attacks is instead to change the decision of the classifier to a particular class $t$, i.e., to find $\hat{x}$ so that $\mathop{\rm arg\,max}_k f_k(\hat{x}) = t$ under the same constraints on $\hat{x}$. We further discuss the targeted attacks in Sup.~\ref{sec:targeted_app}.
	
	\subsection{General Algorithmic Scheme of the Square Attack} \label{sec:general_scheme}
	Square Attack is based on \textit{random search} which is a well known iterative technique in optimization introduced by Rastrigin in 1963~\cite{rastrigin1963convergence}.
	The main idea of the algorithm is to sample a random update $\delta$ at each iteration, and to add this update to the current iterate $\hat{x}$ if it improves the objective function. 
	Despite its simplicity, random search performs well in many situations \cite{zabinsky2010random} and does not depend on gradient information from the objective function $g$.
	\begin{algorithm}[t]
		\caption{The Square Attack via random search}
		\label{alg:random_search}
		\KwIn{classifier $f$, point $x \in \R^d$, image size $w$, number of color channels $c$, $l_p$-radius $\epsilon$, label $y \in \{1, \dots, K\}$, number of iterations $N$}
		\KwOut{approximate minimizer $\hat{x} \in \R^d$ of the problem stated in Eq.~\eqref{eq:opt_problem}}    
		$\hat{x} \gets init(x), \ \ l^* \gets L(f(x), y), \ \ i\gets 1$ \\
		\While{$i < N$ \textbf{and} $\hat{x}$ is not adversarial}{ %${\rm argmax}_{i=1, \ldots, K} s_i = y$}{ %\tcc{For targeted: ${\rm argmax}_{i=1, \ldots, K} s_i \neq y$}
			$h^{(i)}\gets$ side length of the square to modify (according to some schedule)\\
			$\delta \sim P(\epsilon, h^{(i)}, w, c, \hat{x}, x)$ (see Alg.~\ref{alg:linf_sampling_distribution} and \ref{alg:l2_sampling_distribution} for the sampling distributions)\\
			$\hat{x}_{\textrm{new}} \gets$ Project $\hat{x} + \delta$ onto $\{z \in\R^d:\norm{z-x}_p\leq \epsilon\} \cap [0, 1]^d$  \\
			$l_{\textrm{new}} \gets L(f(\hat{x}_{\textrm{new}}), y)$ \\
			\lIf{$l_{\textrm{new}} < l^*$}{
				$\hat{x} \gets \hat{x}_{\textrm{new}},\; l^* \gets l_{\textrm{new}}$
			}
			$i \gets i + 1$
		}
	\end{algorithm}
	
	Many variants of random search have been introduced \cite{matyas1965random,schumer1968adaptive,schrack1976optimized}, which differ mainly in how the random perturbation is chosen at each iteration (the original scheme samples uniformly on a hypersphere of fixed radius). 
	For our goal of crafting adversarial examples we come up with two sampling distributions specific to the $l_\infty$- and the $l_2$-attack (Sec.~\ref{sec:linf_square} and Sec.~\ref{sec:sampl_l2}), which we integrate in the classic random search procedure. These sampling distributions are motivated by both
	how images are processed by neural networks with convolutional filters and the shape of the $l_p$-balls for different $p$. 
	Additionally, since the considered objective is non-convex when using neural networks, a good initialization is particularly important. We then introduce a specific one for better query efficiency.
	
	Our proposed scheme differs from classical random search by the fact that the perturbations 
	$\hat{x}-x$
	are constructed such that for every iteration they lie on the boundary of the $l_\infty$- or $l_2$-ball before projection onto the image domain $[0,1]^d$. Thus we are using the perturbation budget almost maximally at each step. Moreover, the changes are localized in the image in the sense that at each step we modify just a small fraction of contiguous pixels shaped into \textbf{squares}.
	Our overall scheme is presented in Algorithm~\ref{alg:random_search}. First, the algorithm picks the side length $h^{(i)}$  of the square to be modified (step 3), which is decreasing according to an
	a priori fixed schedule. This is in analogy to the step-size reduction in gradient-based optimization. Then in step~4
	we sample a new update $\delta$ and add it to the current iterate (step 5).
	If the resulting loss (obtained in step 6) is smaller than the best loss so far, the change is accepted otherwise discarded. Since we are interested in query efficiency, the algorithm stops as soon as an adversarial example is found. 
	The time complexity of the algorithm is dominated by the evaluation of $f(\hat{x}_{\textrm{new}})$, which is performed at most $N$ times, with $N$ total number of iterations.
	We plot the resulting adversarial perturbations in Fig.~\ref{fig:vis} and additionally in Sup.~\ref{app:additional_exp_results} where we also show imperceptible perturbations.
	
	We note that previous works \cite{ilyas2019prior,MooEtAl2019,MeuEtAl2019} generate perturbations containing squares. %which are however significantly different from ours both in spirit and in practice.
	%In fact, in those cases the square with the constant perturbations are the results of a dimension reduction, and the grid on which those lay is a priori fixed. Conversely,
	However, while those use a fixed grid on which the squares are constrained, we optimize the position of the squares as well as the color, making our attack more flexible and effective. %which are motivated by the structure of the convolutional filters
	Moreover, unlike previous works, we motivate squared perturbations with the structure of the convolutional filters (see Sec.~\ref{sec:theory}).
	
	\customparagraph{Size of the squares.} Given images of size $w\times w$, let $p\in[0,1]$ be the percentage of elements of $x$ to be modified. The length $h$ of the side of the squares used is given by the closest positive integer to $\sqrt{p \cdot w^2}$ (and $h \geq 3$ for the $l_2$-attack). Then, the initial $p$ is the only free parameter of our scheme. With $N=10000$ iterations available, we halve the value of $p$ at $i\in\{10, 50, 200, 1000, 2000, 4000, \allowbreak 6000, \allowbreak 8000\}$ iterations. For different $N$ we rescale the schedule accordingly.

	\subsection{The $l_\infty$-Square Attack}\label{sec:linf_square}
	\customparagraph{Initialization.}
	As initialization we use vertical stripes of width one where the color of each stripe is sampled uniformly at random from $\{-\epsilon, \epsilon\}^c$ ($c$ number of color channels). 
	We found that convolutional networks are particularly sensitive to such perturbations, see also \cite{yin2019fourier} for a detailed discussion on the sensitivity of neural networks to various types of high frequency perturbations.
	
	\customparagraph{Sampling distribution.}
	Similar to \cite{MooEtAl2019} we observe that successful $l_\infty$-perturbations usually have values $\pm \epsilon$ in all the components (note that this does not hold perfectly due to the image
	constraints $\hat{x}\in [0,1]^d$). In particular, it holds
	\[ \hat{x}_i \in \{ \max\{0,x_i-\epsilon\},\min\{1,x_i+\epsilon\}\}.\]
	\begin{wrapfigure}{r}{0.53\textwidth}
		\begin{algorithm}[H]
			\caption{Sampling distribution $P$ for $l_\infty$-norm} 
			\label{alg:linf_sampling_distribution}
			\KwIn{maximal norm $\epsilon$, window size $h$, image size $w$, color channels $c$}
			\KwOut{New update $\delta$}
			$\delta \gets $ array of zeros of size $w\times w\times c$\\
			sample uniformly $r, s \in \{0, \ldots, w - h\} \subset \N$\\
			\For{$i=1,\ldots, c$}{%sample value
				$\rho \gets Uniform(\{-2\epsilon, 2\epsilon\})$ \\
				$\delta_{r+1: r+h,\;s+1:s+h,\;i} \gets \rho \cdot \mathbbm{1}_{h\times h}$ %\tcc{assign the same value to all locations}
			}
		\end{algorithm}
	\end{wrapfigure}
	Our sampling distribution $P$ for the $l_\infty$-norm described in Algorithm~\ref{alg:linf_sampling_distribution} selects sparse updates of $\hat{x}$ with $\norm{\delta}_0 = h \cdot h \cdot c$ where $\delta \in \{-2\epsilon, 0, 2\epsilon\}^d$ and the non-zero elements are grouped to form a square. In this way, after the projection onto the $l_\infty$-ball of radius $\epsilon$ (Step 5 of Algorithm~\ref{alg:random_search}) all components $i$ for which $\epsilon \leq x_i \leq 1-\epsilon$ satisfy $\hat{x}_i \in \{x_i-\epsilon, x_i+\epsilon\}$, i.e. differ from the original point $x$ in each element either by $\epsilon$ or $-\epsilon$. Thus $\hat{x} - x$ is situated at one of the corners of the $l_\infty$-ball (modulo the components which are close to the boundary).
	Note that all projections are done by clipping.
	Moreover, we fix the elements of $\delta$ belonging to the same color channel to have the same sign, since we observed that neural networks are particularly sensitive to such perturbations (see Sec.~\ref{sec:ablation}).

	\subsection{The $l_2$-Square Attack}
	\label{sec:sampl_l2}
	
	\begin{wrapfigure}{r}{0.5\columnwidth}
		\includegraphics[scale=0.1]{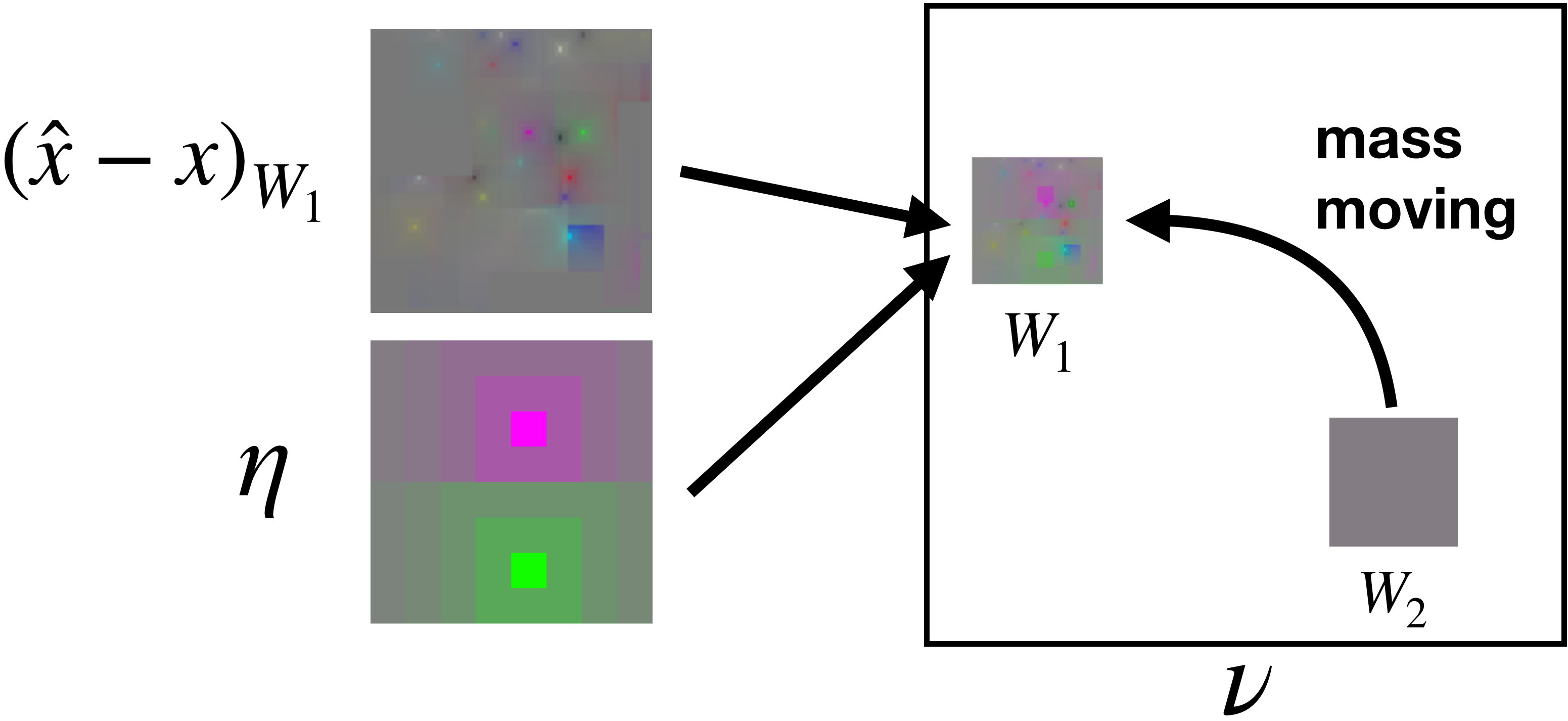}
		\caption{Perturbation of the $l_2$-attack} 
		\label{fig:l2_pert}
	\end{wrapfigure}
	\customparagraph{Initialization.} The $l_2$-perturbation is initialized by generating a $5 \times 5$ grid-like tiling by squares of the image, where the perturbation on each tile
	has the shape described next in the sampling distribution. The resulting perturbation $\hat{x}-x$ is rescaled to have $l_2$-norm $\epsilon$ and the resulting $\hat{x}$ is projected onto $[0,1]^d$ by clipping.
	
	\customparagraph{Sampling distribution.} First, let us notice that the adversarial perturbations typically found for the $l_2$-norm tend to be much more localized than those for the $l_\infty$-norm~\cite{tsipras2019robustness}, in the sense that large changes are applied on some pixels of the original image, while many others are minimally modified.
	To mimic this feature we introduce a new update $\eta$ which has two "centers" with large absolute value and opposite signs, while the other components have lower absolute values as one gets farther away from the centers, but never reaching zero (see Fig.~\ref{fig:l2_pert} for one example with $h=8$ of the resulting update $\eta$).
	In this way the modifications are localized and with high contrast between the different halves, which we found to improve the query efficiency.
	Concretely, we define  $\eta^{(h_1,h_2)}\in\R^{h_1\times h_2}$ (for some $h_1, h_2 \in \N_+$ such that $h_1\geq h_2$) for every $1 \leq r \leq h_1,
	\, 1 \leq s \leq h_2$ as 
	\begin{align*}
	\eta^{(h_1,h_2)}_{r,s} = \sum_{k=0}^{M(r,s)} \frac{1}{(n+1-k)^2}, \;\textrm{ with }\; n=\floor{\frac{h_1}{2}},
	\end{align*}
	and $M(r,s) =n- \max\{|r-\floor{\frac{h_1}{2}}-1|,|s-\floor{\frac{h_2}{2}}-1|\}$. The intermediate square update $\eta\in \R^{h\times h}$ is then selected uniformly at random  from either \begin{equation}\label{eq:l2_pgp} \eta = \left( \eta^{(h,k)}, -\eta^{(h,h-k)}\right),\quad \textrm{with}\; k = \floor{h/2},
	\end{equation}
	or its transpose (corresponding to a rotation of $90^\circ$).

	\begin{algorithm}[t]
		\caption{Sampling distribution $P$ for $l_2$-norm} 
		\label{alg:l2_sampling_distribution}
		\KwIn{maximal norm $\epsilon$, window size $h$, image size $w$, number of color channels $c$, current image $\hat{x}$, original image $x$}
		\KwOut{New update $\delta$}
		$\nu \gets \hat{x} - x$\\
		sample uniformly $r_1, s_1, r_2, s_2 \in \{0, \ldots, w - h\}$\\
		$W_1:= r_1+1:r_1+h,s_1+1:s_1+h$, $W_2:=r_2+1:r_2+h,s_2+1:s_2+h$ \\
		$\epsilon_{\textrm{unused}}^2 \gets \epsilon^2 - \norm{\nu}_2^2$, \ \ $\eta^* \gets \nicefrac{\eta}{\norm{\eta}_2}$ with $\eta$ as in \eqref{eq:l2_pgp}\\
		\For{$i=1,\ldots, c$}{$\rho \gets  Uniform(\{-1, 1\})$ \\
			$\nu_{\textrm{temp}} \gets \rho \eta^* + \nicefrac{\nu_{W_1,i}}{\norm{\nu_{W_1,i}}_2}$\\
			$\epsilon^i_{\textrm{avail}} \gets \sqrt{\norm{\nu_{W_1 \cup W_2,i}}_2^2+\nicefrac{\epsilon_{\textrm{unused}}^2}{c}}$\\
			$\nu_{W_2,i} \gets 0$, \ \ $\nu_{W_1,i} \gets (\nicefrac{\nu_{\textrm{temp}}}{\norm{\nu_{\textrm{temp}}}_2})\epsilon^i_{\textrm{avail}}$\\}
		$\delta \gets x +\nu - \hat{x}$
	\end{algorithm}

	\setlength{\newl}{0.19\columnwidth}
	\begin{figure*}[t] \centering\small \begin{tabular}{c| c  | c  }\textbf{original}& \multicolumn{1}{c|}{\textbf{$l_\infty$-attack} - $\epsilon_\infty=0.05$}&
			\multicolumn{1}{c}{\textbf{$l_2$-attack} - $\epsilon_2=5$} \\
			\commentout{
				\includegraphics[width=\newl,clip, trim=20mm 10mm 20mm 5mm]{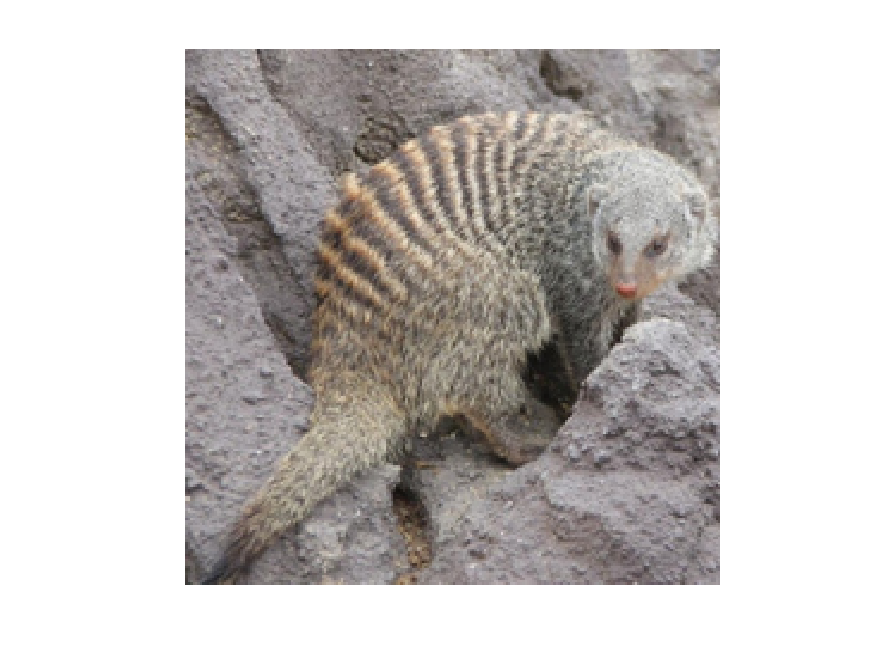} &
				\includegraphics[width=\newl,clip, trim=20mm 10mm 20mm 5mm]{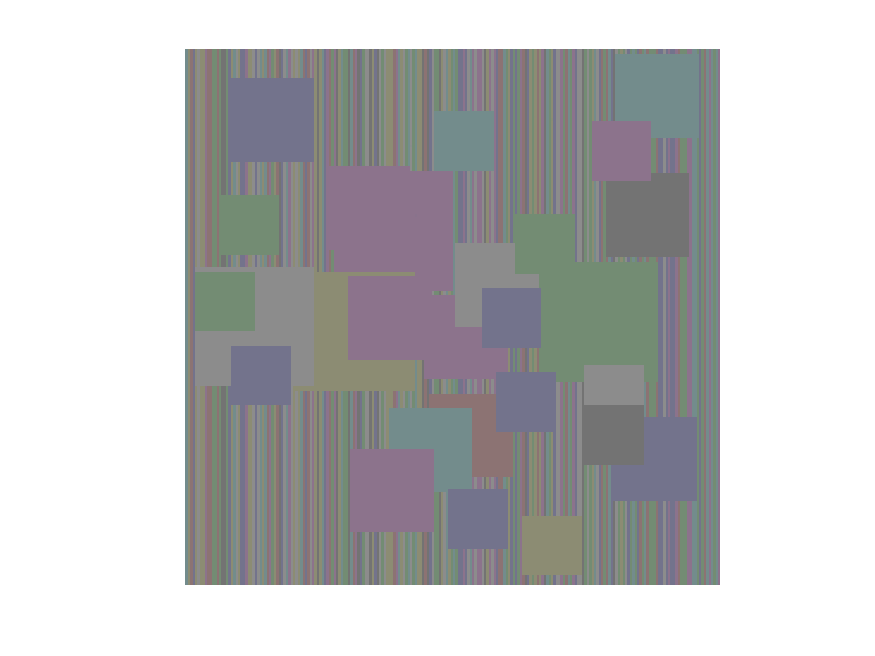}
				\includegraphics[width=\newl,clip, trim=20mm 10mm 20mm 5mm]{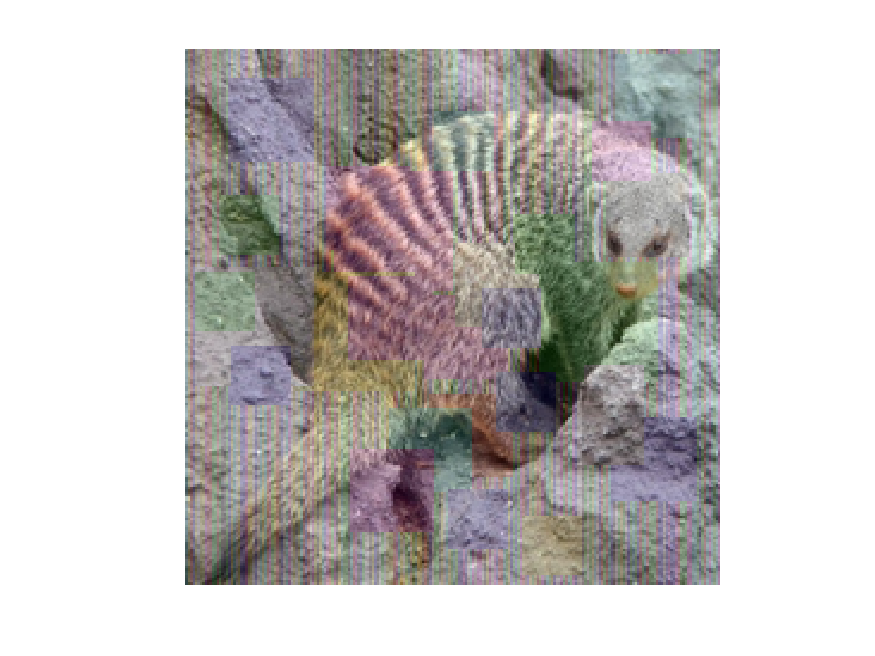}&
				\includegraphics[width=\newl,clip, trim=20mm 10mm 20mm 5mm]{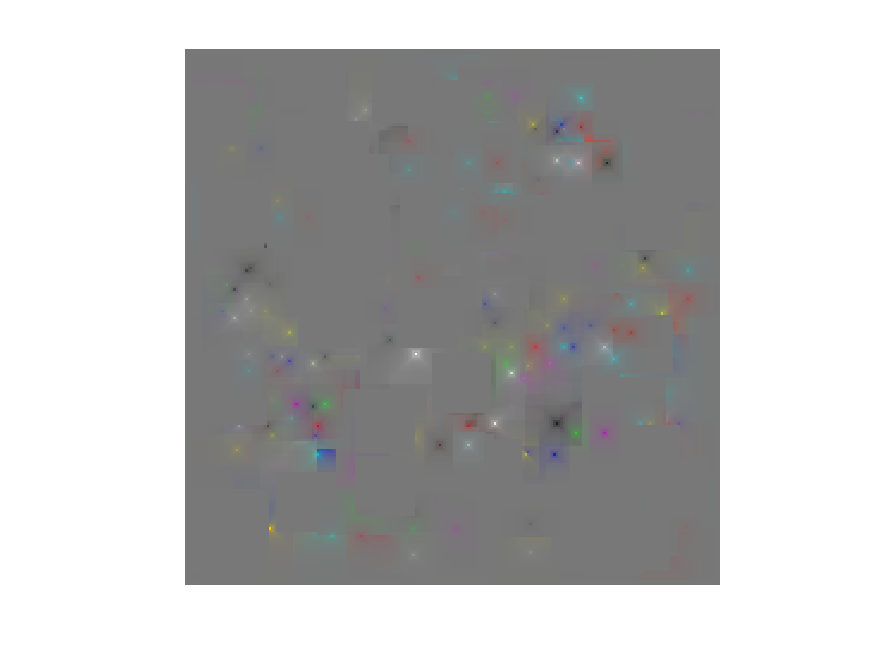}
				\includegraphics[width=\newl,clip, trim=20mm 10mm 20mm 5mm]{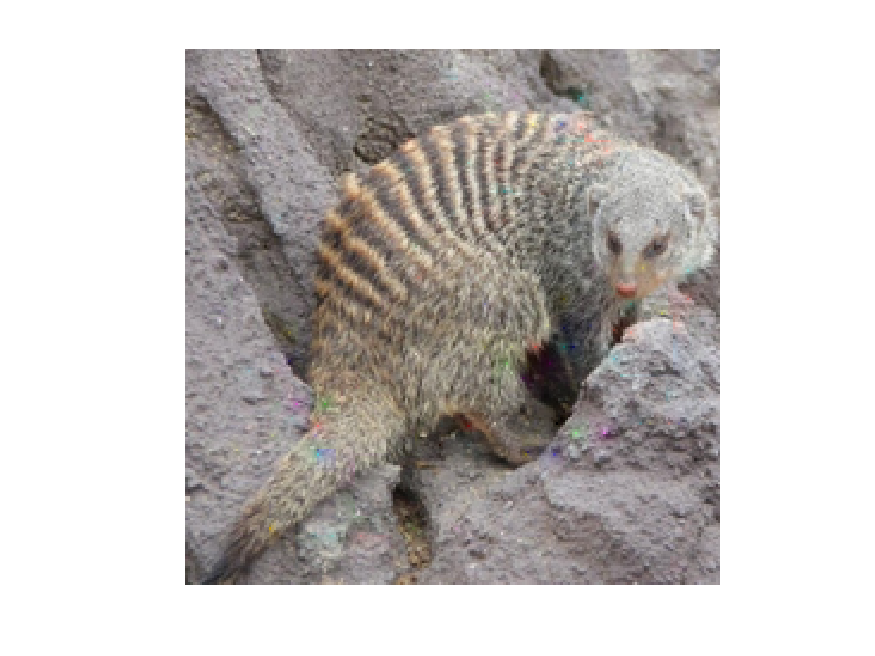}  \\}
			\commentout{
				\includegraphics[width=\newl,clip, trim=20mm 10mm 20mm 5mm]{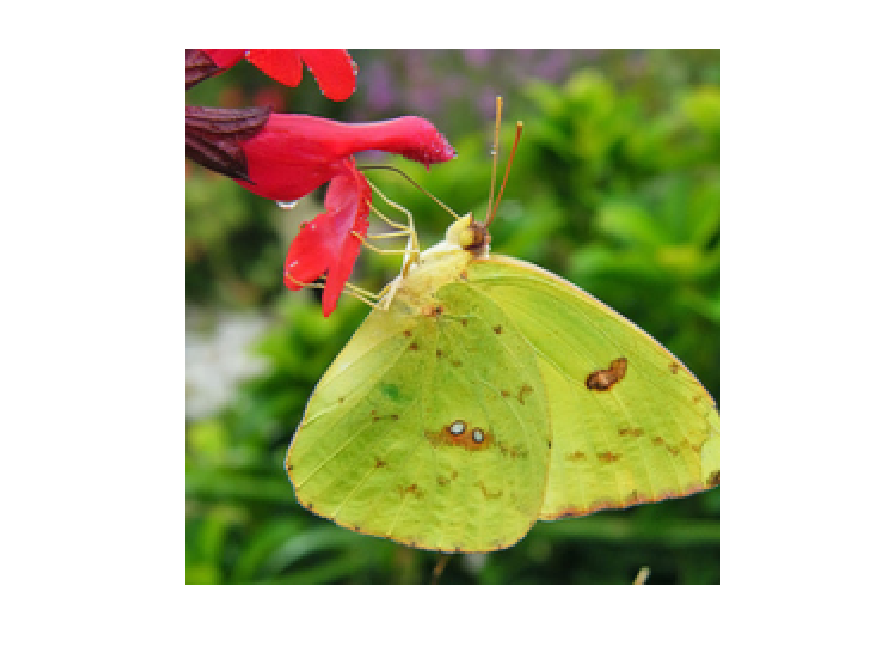} &
				\includegraphics[width=\newl,clip, trim=20mm 10mm 20mm 5mm]{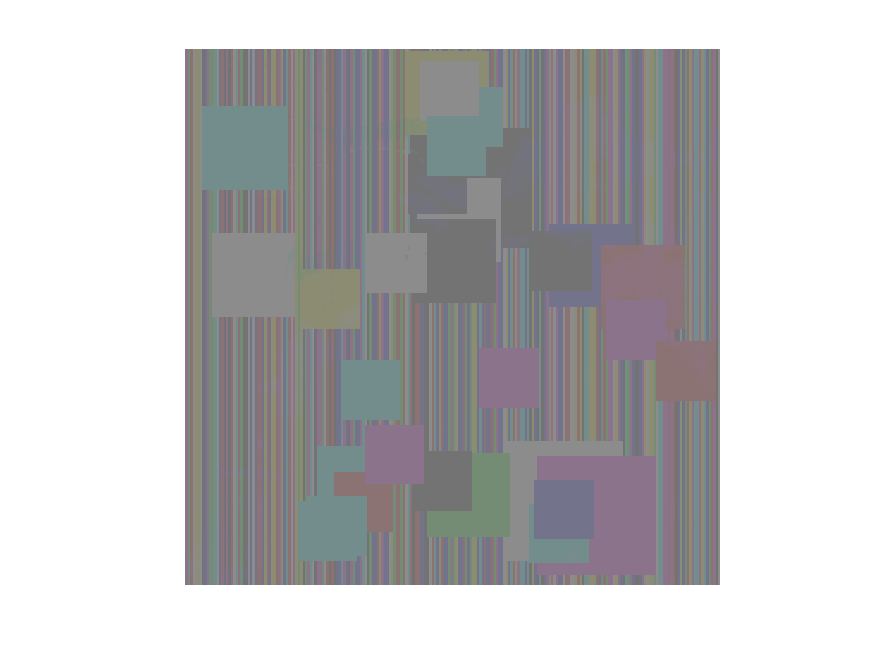}
				\includegraphics[width=\newl,clip, trim=20mm 10mm 20mm 5mm]{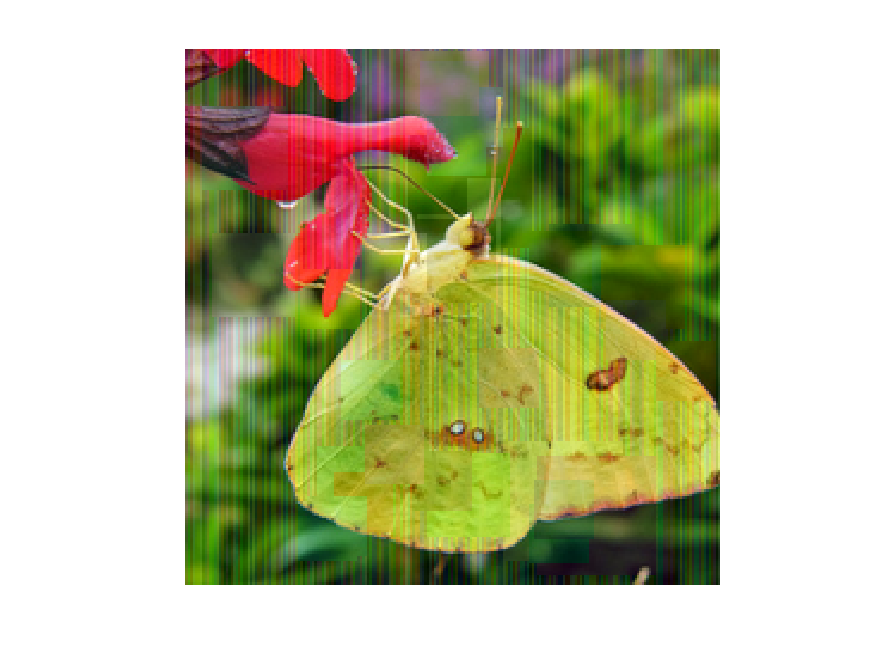}&
				\includegraphics[width=\newl,clip, trim=20mm 10mm 20mm 5mm]{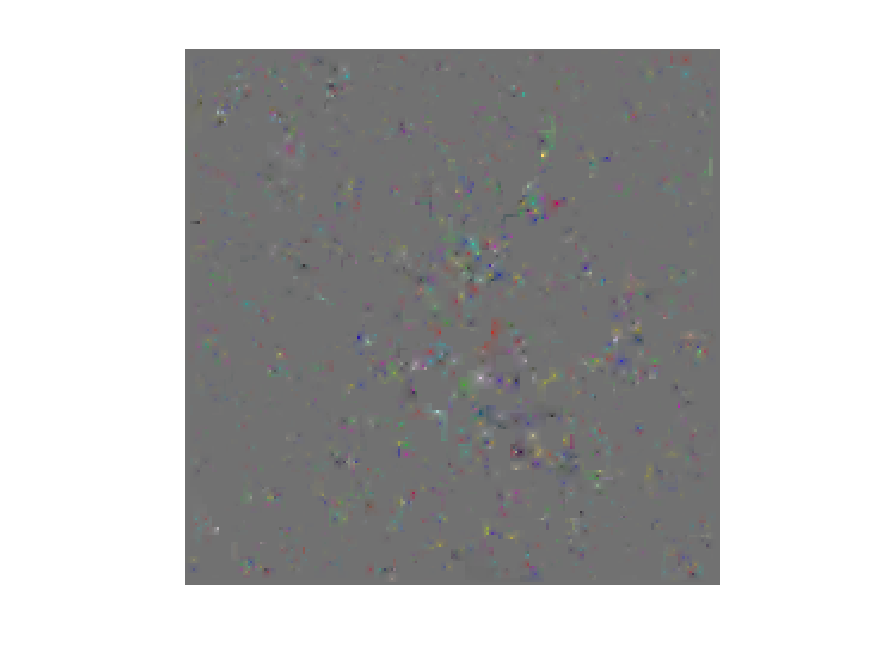} 
				\includegraphics[width=\newl,clip, trim=20mm 10mm 20mm 5mm]{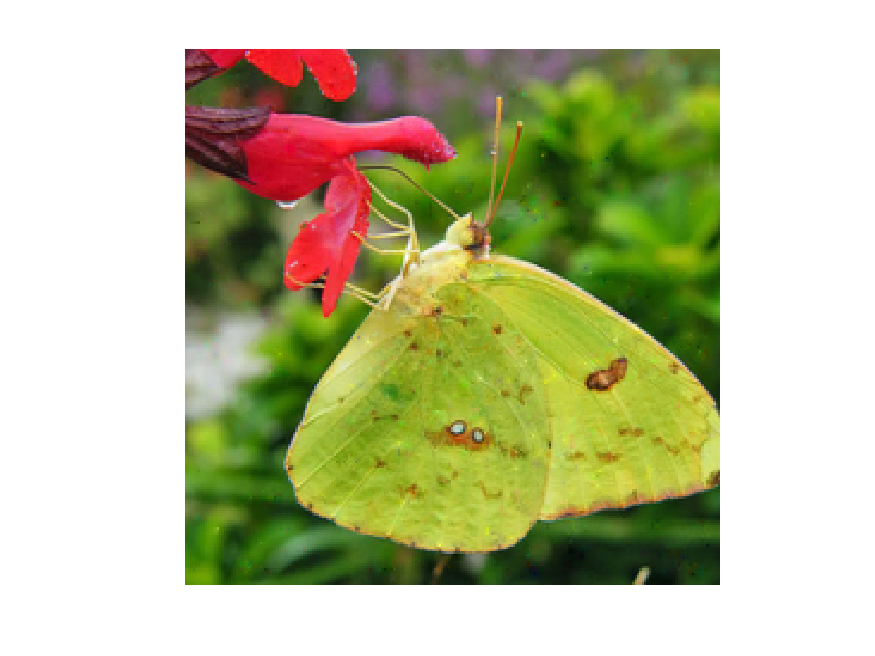}\\}
			%\commentout{
			\includegraphics[width=\newl,clip, trim=20mm 10mm 20mm 5mm]{l2_img_17_orig} & \includegraphics[width=\newl,clip, trim=20mm 10mm 20mm 5mm]{linf_img_17_diff_2}
			\includegraphics[width=\newl,clip, trim=20mm 10mm 20mm 5mm]{linf_img17_adv}& 
			\includegraphics[width=\newl,clip, trim=20mm 10mm 20mm 5mm]{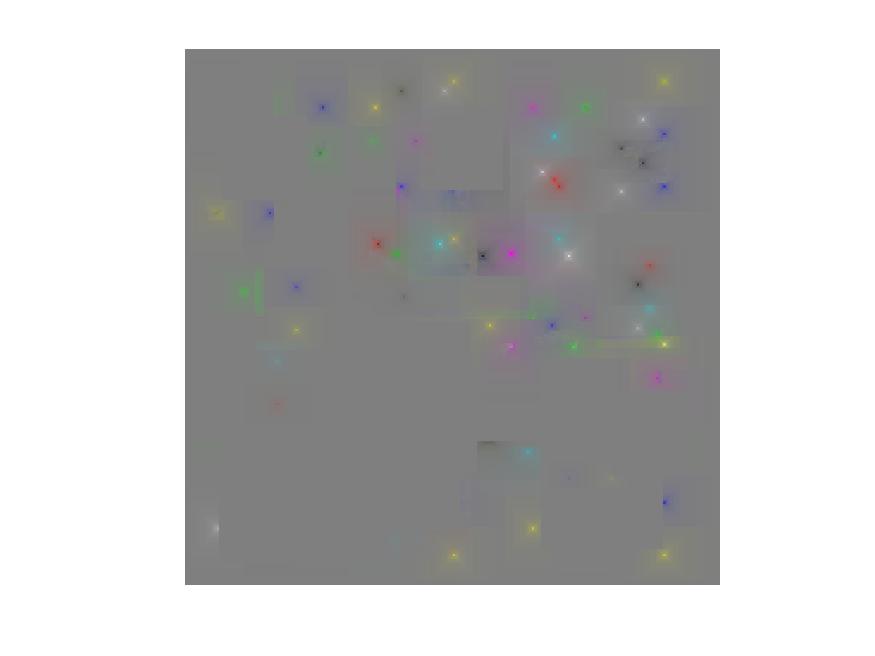}
			\includegraphics[width=\newl,clip, trim=20mm 10mm 20mm 5mm]{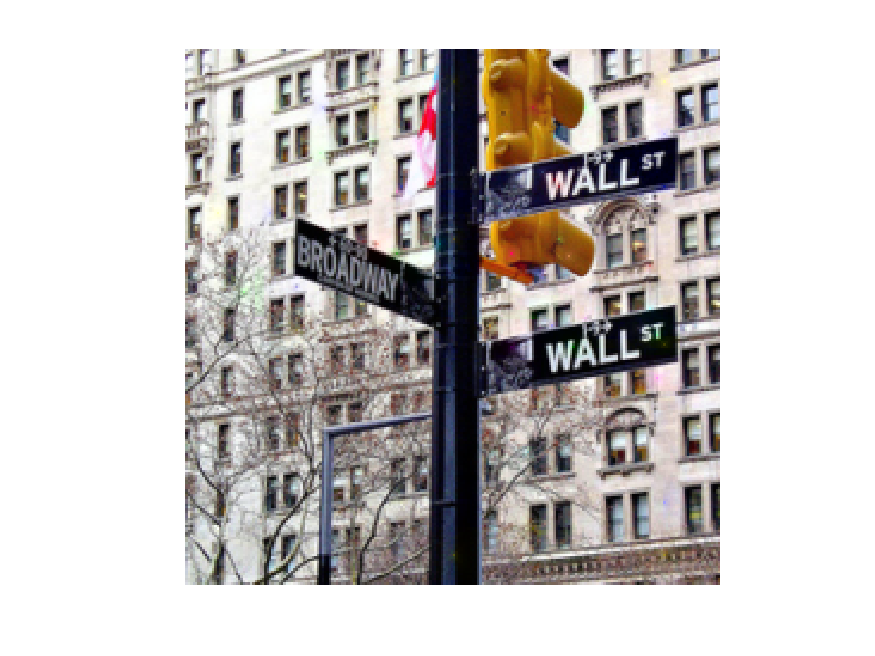}%}
		\end{tabular}
		\caption{Visualization of the adversarial perturbations and examples found by the $l_\infty$- and $l_2$-versions of the Square Attack on ResNet-50} \label{fig:vis}
	\end{figure*}
	
	Second, unlike $l_\infty$-constraints, $l_2$-constraints do not allow to perturb each component independently from the others as the overall $l_2$-norm must be kept
	smaller than $\epsilon$.
	% $\leq\epsilon$. 
	Therefore,
	%if we want
	to modify a perturbation $\hat x - x$ of norm $\epsilon$ with localized changes while staying on the hypersphere, we have to "move the mass" of $\hat x -x$ from one location to another.
	%Therefore, if we have $\norm{\delta}_2=\epsilon$ and we want to change $\delta$ through local changes but staying on the hypersphere, we have to "move the mass" of $\delta$ from one location to another. 
	Thus, our scheme consists in randomly selecting two squared windows in the current perturbation $\nu=\hat x-x$, namely $\nu_{W_1}$ and $\nu_{W_2}$, setting $\nu_{W_2}=0$ and using the budget of $\norm{\nu_{W_2}}_2$ to increase the total perturbation of $\nu_{W_1}$. Note that the perturbation of $W_1$ is then a combination of the existing perturbation plus the new generated $\eta$. We report the details of this scheme in Algorithm~\ref{alg:l2_sampling_distribution} where step~4 allows to utilize the budget of $l_2$-norm lost after the projection onto $[0,1]^d$.  The update $\delta$ output by the algorithm is such that the next iterate $\hat{x}_{\textrm{new}}=\hat{x}+\delta$  (before projection onto $[0,1]^d$ by clipping) belongs to the hypersphere $B_2(x,\epsilon)$
	%of center $x$ and radius $\varepsilon$
	%\norm{\hat{x}_{\textrm{new}} - x}_2=\epsilon$,
	as stated in the following proposition.
	%(proven in Supp~\ref{app:prol2}).
	%
	%\clearpage

	\begin{proposition}
		\label{pro:l2}
		Let $\delta$ be the output of Algorithm \ref{alg:l2_sampling_distribution}. Then $\norm{\hat{x} + \delta - x}_2=\epsilon$.
	\end{proposition}

	\section{Theoretical and Empirical Justification of the Method}
	\label{sec:theory}
	We provide high-level theoretical justifications and empirical evidence regarding the algorithmic choices in Square Attack, with focus on the $l_\infty$-version (the $l_2$-version is significantly harder to analyze).
	
	\subsection{Convergence Analysis of Random Search}\label{sec:convergence}
	First, we want to study the convergence of the random search algorithm when considering an $L$-smooth objective function $g$ (such as neural networks with activation functions like softplus, swish, ELU, etc) on the whole space $\mathbb{R}^d$ (without projection\footnote{ Nonconvex constrained optimization under noisy oracles is notoriously harder~\cite{davis2019stochastic}.}) under the following assumptions on the update $\delta_t$ drawn from the sampling distribution $P_t$:
	\begin{equation}\label{eq:asspt}
	\mathbb{E} \Vert \delta_t \Vert_2^2\leq \gamma_t^2 C \ \text{and } \ \mathbb{E} \vert \langle  \delta_t, v\rangle \vert \geq \tilde C \gamma_t \Vert v \Vert_2, \ \forall v\in \R^d,
	\end{equation}
	where $\gamma_t$ is the step size at iteration $t$, $C, \tilde C > 0$ some constants and $\langle \cdot, \cdot\rangle$ denotes the inner product.
	We obtain the following result, similar to existing convergence rates for zeroth-order methods~\cite{NemYud83,nesterov2017random,duchi2015optimal}:
	\begin{proposition}\label{pro:conv} Suppose that $\mathbb E[\delta_t]=0$ and the assumptions in Eq.~\eqref{eq:asspt} hold. Then for step-sizes $\gamma_t=\gamma/\sqrt{T}$, we have
		\[	
		\min_{t=0,\dots,T}\! \mathbb E\Vert \nabla g (x_t)\Vert_2 \!\leq \! \frac{2}{\gamma \tilde C \sqrt{T}} \bigg( g(x_0)- 
		\mathbb E g(x_{T+1})
		%g(x_\star)
		+\frac{\gamma^2 CL}{2} \bigg).
		\]
	\end{proposition}
	This basically shows for $T$ large enough one can make the gradient arbitrary small, meaning that the random search algorithm converges to a critical point of $g$ (one cannot hope for much stronger results
	in non-convex optimization without stronger conditions). 
	
	Unfortunately, the second assumption in Eq.~\eqref{eq:asspt} does not directly hold for our sampling distribution $P$ for the $l_\infty$-norm (see Sup.~\ref{app:assno}),
	but holds for a similar one, $P^{\text{multiple}}$, where each component of the update $\delta$ is drawn uniformly at random from $\{-2\epsilon,2\epsilon\}$.
	In fact we show in Sup.~\ref{app:assyes}, using the Khintchine inequality~\cite{haagerup1981best}, that %(see Sup.~\ref{app:assyes})
	\begin{equation*}
	\mathbb{E} \Vert \delta_t \Vert_2^2\leq   4 c \varepsilon^2 h^2 \ \text{and } \ \mathbb{E} \vert \langle  \delta_t, v\rangle \vert \geq \frac{\sqrt{2}c\varepsilon  h^2}{d} \Vert v \Vert_2, \ \forall v\in \R^d.
	\end{equation*}
	Moreover, while $P^{\text{multiple}}$ performs worse than the distribution used in Algorithm~\ref{alg:linf_sampling_distribution}, we show in Sec.~\ref{sec:ablation} that it already reaches state-of-the-art results.

	\subsection{Why Squares?}
	\label{sec:why_squares}
	Previous works~\cite{MooEtAl2019,MeuEtAl2019} build their $l_\infty$-attacks by iteratively adding square modifications. Likewise we change square-shaped regions of the image for both our $l_\infty$- and $l_2$-attacks---with the difference that we can sample any square subset of the input, while the grid of the possible squares is fixed in~\cite{MooEtAl2019,MeuEtAl2019}. This leads naturally to wonder why squares are superior to other shapes, e.g. rectangles.
	
	Let us consider the $l_\infty$-threat model, with bound $\epsilon$, input space $\R^{d \times d}$ and a convolutional filter $w \in \R^{s \times s}$ with entries unknown to the attacker. Let $\delta\in\R^{d \times d}$ be the sparse update with $\norm{\delta}_0=k\geq s^2$ and $\norm{\delta}_\infty \leq \epsilon$.
	We denote by $S(a,b)$ the index set of the rectangular support of $\delta$ with $|S(a,b)|=k$ and shape $a\times b$.
	We want to provide intuition why sparse square-shaped updates are superior to rectangular ones in the sense of reaching a maximal change in the activations of the first convolutional layer.

	Let $z = \delta \ast w \in \R^{d\times d}$ denote the output of the convolutional layer for the update $\delta$. The $l_\infty$-norm of $z$ is the maximal componentwise change of the convolutional layer:
	\begin{align*}
	\norm{z}_\infty&=\maxop_{u,v}|z_{u,v}|=\maxop_{u,v} \Big|\sum_{i,j=1}^{s} \delta_{u-\lfloor \frac{s}{2} \rfloor+i, v -\lfloor \frac{s}{2} \rfloor+ j}\cdot w_{i,j}\Big|\\
	&
	\leq \maxop_{u,v}
	\epsilon \sum_{i,j} |w_{i,j}| \Id_{(u-\lfloor \frac{s}{2} \rfloor+i,v -\lfloor \frac{s}{2} \rfloor+ j
		) \in S(a,b)},
	\end{align*}
	where elements with indices exceeding the size of the matrix are set to zero.
	Note that the indicator function attains 1 only for the non-zero elements of $\delta$ involved in the convolution to get $z_{u,v}$.
	Thus, to have the largest upper bound possible on $|z_{u,v}|$, for some $(u,v)$, we need the largest possible amount of components of $\delta$ with indices in
	\[C(u,v) =\left\lbrace(u-\lfloor \frac{s}{2} \rfloor+i,v -\lfloor \frac{s}{2} \rfloor+ j):\; i,j=1,\ldots,s\right\rbrace\] to be non-zero (that is in $S(a,b)$).
	
	Therefore, it is desirable to have the shape $S(a,b)$  of the perturbation $\delta$ selected so to maximize the number $N$ of convolutional filters $w\in \mathbb{R}^{s\times s}$ which fit into the rectangle $a\times b$.
	Let $\F$ be the family of the objects that can be defined as the union of axis-aligned rectangles with vertices on $\N^2$, and $\G \subset \F$ be the squares of $\F$ of shape $s\times s$ with $s\geq 2$. We have the following proposition:
	\begin{proposition}\label{prop:squares_1}
		Among the elements of $\F$ with area $k\geq s^2$, those which contain the largest number of elements of $\G$ have \begin{equation} N^* = (a -s+1)(b-s+1) +(r - s +1)^+\end{equation} of them, with $a=\floor{\sqrt{k}}$, $b=\floor{\frac{k}{a}}$, $r = k - ab$ and $z^+=\max\{z,0\}$.
	\end{proposition}
	This proposition states that, if we can
	%select
	modify only $k$ elements of $\delta$,
	%to modify,
	then shaping them to form (approximately) a square allows to maximize the number of
	pairs $(u,v)$ for which $|S(a,b)\cap C(u,v)|=s^2$.
	If $k=l^2$ then $a=b=l$ are the optimal values for the shape of the perturbation update, i.e. the shape is exactly a square.

	\subsection{Ablation Study}\label{sec:ablation}
	
	\definecolor{Gray}{rgb}{0.9, 0.9, 1}
	\begin{table}[t] \centering
		\caption{Ablation study of the $l_\infty$-Square Attack which shows how the individual 
			design decisions improve the performance.
			%with various algorithmic choices. 
			The fourth row corresponds to the method for which we have shown convergence guarantees in Sec.~\ref{sec:convergence}. The last row 
			corresponds to our final $l_\infty$-attack. $c$ indicates the number of color channels, $h$ the length of the side of the squares, so that "\# random sign" $c$ represents updates with constant sign for each color, while $c\cdot h^2$ updates with signs sampled independently of each other
			%represents our final version of the attack used in the
			%experiments.
		}
		\label{tab:ablation_study_main}
		{\small
			% \textbf{$l_\infty$ ablation study}
			\setlength{\tabcolsep}{4.0pt}
			\begin{tabular}{c c c | c c c}
				\hline
				Update & \# random & \multirow{2}{*}{Initialization} & Failure & Avg. & Median\\
				shape  & signs     &                                 & rate    & queries & queries\\
				\hline
				random & $c \cdot h^2$  & vert. stripes  &  \textbf{0.0\%} & 401 & 48\\
				random & $c \cdot h^2$  & uniform rand.  &  \textbf{0.0\%} & 393 & 132\\
				random & $c$            & vert. stripes  &  \textbf{0.0\%} & 339 & 53\\
				\rowcolor{Gray}
				square & $c \cdot h^2$ & vert. stripes   &  \textbf{0.0\%} & 153 & 15\\
				% square & $1$ & vert. stripes             &  \textbf{0.0\%} & 129 & 18\\
				rectangle & $c$ & vert. stripes          &  \textbf{0.0\%} & 93 & 16\\
				square & $c$ & uniform rand.             &  \textbf{0.0\%} & 91 & 26\\
				% square & $c$ & rand. squares             &  \textbf{0.0\%} & 90 & 20\\
				% square & $c$ & horiz. stripes            &  \textbf{0.0\%} & 83 & 18\\
				\rowcolor{Gray}
				square & $c$ & vert. stripes             &  \textbf{0.0\%} & \textbf{73} & \textbf{11}\\
				\hline
		\end{tabular}}
	\end{table}
	
	We perform an ablation study to show how the individual design decisions for the sampling distribution of the random search improve the performance of $l_\infty$-Square Attack, confirming the theoretical arguments
	above.
	The comparison is done for an $l_\infty$-threat model of radius $\epsilon=0.05$ on $1,000$ test points for a ResNet-50 model trained normally on ImageNet (see Sec.~\ref{sec:exp} for details) with a query limit of $10,000$ and results are shown in Table~\ref{tab:ablation_study_main}.
	Our sampling distribution is special in two aspects: i) we use localized update shapes in form of squares and ii) the update is constant in each color channel.  First, one can observe that
	our update shape ``square'' performs better than ``rectangle'' as we discussed in the previous section, and it is significantly better than ``random'' (the same amount of pixels is perturbed, but selected randomly in the image). This holds both for $c$ (constant sign per color channel) and $c\cdot h^2$ (every pixel and color channel is changed independently of each other), with an improvement in terms of average queries of 339 to 73 and 401 to 153 respectively. Moreover, with updates of the same shape, the constant sign over color channels is better than selecting it uniformly at random (improvement in average queries: 401 to 339 and 153 to 73). In total the algorithm with ``square-$c$'' needs more than $5\times$ less average queries than ``random-$c\cdot h^2$'', showing that our sampling distribution is the key to the high query efficiency of Square Attack.
	
	The last innovation of our random search scheme is the initialization, crucial element of every non-convex optimization algorithm. Our method 
	(``square-$c$'') with the vertical stripes initialization
	improves over a uniform initialization on average by $\approx25\%$ and, especially, median number of queries (more than halved).
	
	We want to also highlight that the sampling distribution ``square-$c\cdot h^2$'' for which we shown convergence guarantees in Sec.~\ref{sec:convergence} performs already better in terms of the success rate and the median number of queries than the state of the art (see Sec.~\ref{sec:exp}). For a more detailed ablation, also for our $l_2$-attack, see Sup.~\ref{app:ablation_study}.

	\section{Experiments}\label{sec:exp}
	In this section we show the effectiveness of the Square Attack. Here we concentrate on \textbf{untargeted} attacks since our primary goal is query efficient robustness evaluation, while the \textbf{targeted} attacks are postponed to the supplement. First, we follow the standard setup \cite{ilyas2019prior,MeuEtAl2019} of comparing black-box attacks on three ImageNet models in terms of success rate and query efficiency for the $l_\infty$- and $l_2$-untargeted attacks (Sec.~\ref{sec:exps_imagenet}). 
	Second, we show that our \textit{black-box} attack can even outperform \textit{white-box} PGD attacks on several models (Sec.~\ref{sec:challenging_tasks}). 
	Finally, in the supplement we provide more experimental details (Sup.~\ref{app:exp_details}), 
	a stability study of our attack for different parameters (Sup.~\ref{app:ablation_study}) and random seeds (Sup.~\ref{app:stability}), 
	and additional results including the experiments for targeted attacks (Sup.~\ref{app:additional_exp_results}).

	\subsection{Evaluation on ImageNet}\label{sec:exps_imagenet}
	
	\begin{table*}[t]
		\caption{\label{tab:main_linf}
			Results of \textbf{untargeted} attacks on ImageNet with a limit of 10,000 queries.
			For the $l_\infty$-attack we set the norm bound $\epsilon=0.05$ and for the $l_2$-attack $\epsilon=5$.
			Models: normally trained \textbf{I}: Inception v3, \textbf{R}: ResNet-50, \textbf{V}: VGG-16-BN. The Square Attack outperforms for both threat models all other methods in terms of success rate and query efficiency. The missing entries correspond to the results taken from the original paper where some models were not reported
		}
		\centering
		\small
		\begin{tabular}{c cccccccccccccc}
			\hline
			\multirow{2}{*}{\textbf{Norm}}& \multirow{2}{*}{\textbf{Attack}} & \multicolumn{3}{c}{\textbf{Failure rate}} & \phantom{x} & \multicolumn{3}{c}{\textbf{Avg. queries}} & \phantom{x} & \multicolumn{3}{c}{\textbf{Med. queries}} \\ 
			\cmidrule{3-5} \cmidrule{7-9} \cmidrule{11-13}& & I & R & V && I & R & V && I & R & V \\
			\hline
			% 		\multirow{6}{*}{$l_\infty$} &Bandits \cite{ilyas2019prior} (from the paper) &    1117 & 722 & 370 &&          - &   - &   - &&             4.6\% &  3.4\% &  8.4\% \\
			\multirow{6}{*}{$l_\infty$} &Bandits \cite{ilyas2019prior} &  3.4\% &  1.4\% &  2.0\% &&    957  & 727 & 394 &&        218 & 136 &  36    \\
			&Parsimonious \cite{MooEtAl2019} &                            1.5\% &  -   &  - &&          722  &   - &   - &&        237 &   - &   -     \\
			&DFO$_c$--CMA \cite{MeuEtAl2019} &                 0.8\% &  \textbf{0.0\%} &  0.1\% &&    630  & 270 & 219 &&        259 & 143 & 107  \\
			&DFO$_d$--Diag. CMA \cite{MeuEtAl2019} &           2.3\% &  1.2\% &  0.5\% &&             424  & 417 & 211 &&        \textbf{20} & 20 &  2 &&  \\
			&SignHunter \cite{AlDujaili2019ThereAN} &                      1.0\% &  0.1\%   & 0.3\% &&  471  & 129 & 95  &&         95 &  39 &  43   \\
			&\textbf{Square Attack}&                             \textbf{0.3\%} &  \textbf{0.0\%} &  \textbf{0.0\%} &&    \textbf{197} & \textbf{73}  & \textbf{31}  &&   24 & \textbf{11} &  \textbf{1}  \\
			\hline%\hline
			\multirow{3}{*}{$l_2$} 
			% &SignHunter \cite{AlDujaili2019ThereAN} & 23.0\% & 3.4\%& 0.0\% &&     2235 & 1320 & 803 &&   951 & 354 & 293  \\
			&Bandits \cite{ilyas2019prior} &  9.8\% & 6.8\%& 10.2\% &&    1486 & 939 & 511 &&   660 &   392 &   196  \\
			&SimBA-DCT \cite{guo2019simple}&                     35.5\% & 12.7\% & 7.9\% &&  \textbf{651}& \textbf{582} & 452 & & 564 &   467 &   360  \\
			&\textbf{Square Attack} &                  \textbf{7.1\%} &  \textbf{0.7\%} &  \textbf{0.8\%} &&     1100 &616&\textbf{377}  &&   \textbf{385} & \textbf{170} &  \textbf{109} \\
			\hline
		\end{tabular}
	\end{table*}
	
	\commentout{
		\begin{table*}[t]
			\centering
			\small
			\begin{tabular}{cccccccccccccc}
				\hline
				\multirow{2}{*}{\textbf{Attack}} & \multicolumn{3}{c}{\textbf{Avg. queries}} & \phantom{x} & \multicolumn{3}{c}{\textbf{Median queries}} & \phantom{x} & \multicolumn{3}{c}{\textbf{Failure rate}} \\ 
				\cmidrule{2-4} \cmidrule{6-8} \cmidrule{10-12} & I & R & V && I & R & V && I & R & V \\
				\hline
				Bandits \cite{ilyas2019prior} &                       1486.42 & 939.19 & 511.27 &&          660 &   392 &   196 &&             9.83\% &  6.82\% &  10.20\% \\
				SimBA-DCT \cite{guo2019simple}&                     \textbf{650.68}  &   \textbf{581.75} &   452.44 &&       564 &   467 &   360 &&             35.52\%    &    12.71\%   &    7.92\%   \\
				\textbf{Square Attack (ours)} &                     1099.60 & 615.55  & \textbf{377.30}  &&   \textbf{385} & 169.5 &  109 &&   \textbf{7.08\%} &  \textbf{0.66\%} &  \textbf{0.81\%} \\
				\hline
			\end{tabular}
			\caption{\label{tab:main_l2}
				Results for $l_2$-attacks under 10,000 queries limit with $\epsilon=5$ on ImageNet. Models: standardly trained \textbf{I}: Inception v3, \textbf{R}: ResNet-50, \textbf{V}: VGG-16-BN.}
	\end{table*}}
	
	We compare the Square Attack to state-of-the-art score-based black-box attacks (without any extra information such as surrogate models)
	on three pretrained models in PyTorch
	(Inception v3, ResNet-50, VGG-16-BN) using 1,000 images from the ImageNet validation set. Unless mentioned otherwise, we use the code from the other papers with their suggested parameters. 
	As it is standard in the literature, we give a budget of 10,000 queries per point to find an adversarial perturbation of $l_p$-norm at most $\epsilon$. We report the \textit{average} and \textit{median} number of queries each attack requires to craft an adversarial example, together with the \textit{failure rate}. 
	All query statistics are computed only for successful attacks on the points which were originally correctly classified.
	
\begin{figure}[t] \centering 
		\setlength{\newl}{0.25\columnwidth}
		\setlength{\tabcolsep}{0.5pt}\scriptsize
		\begin{tabular}{c c c c}
			& \textbf{Inception v3} & \textbf{ResNet-50} & \textbf{VGG-16-BN}\\
			\commentout{
				\textbf{$l_\infty$ attacks} & 
				\includegraphics[align=c, width=\newl]{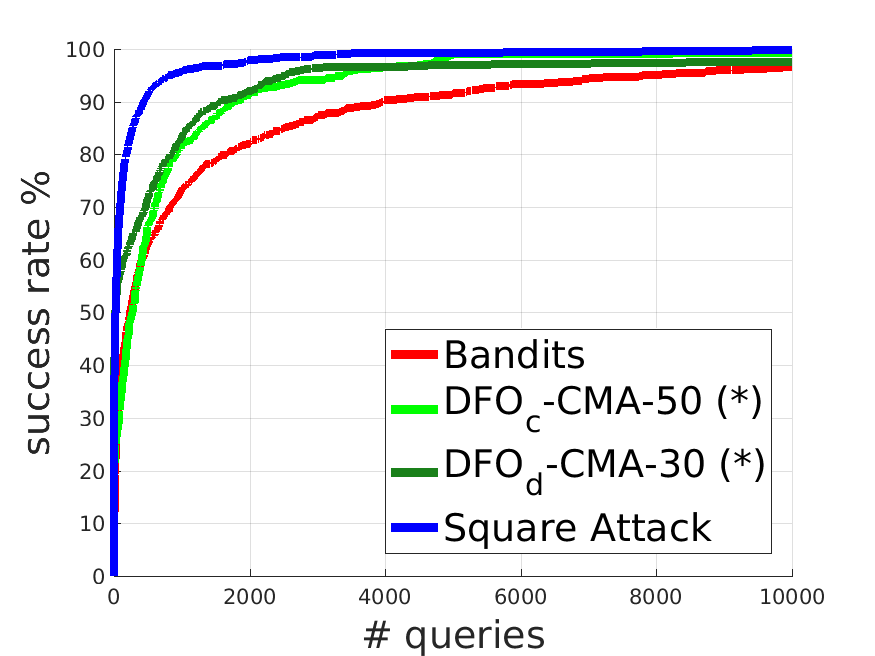}&
				\includegraphics[align=c, width=\newl]{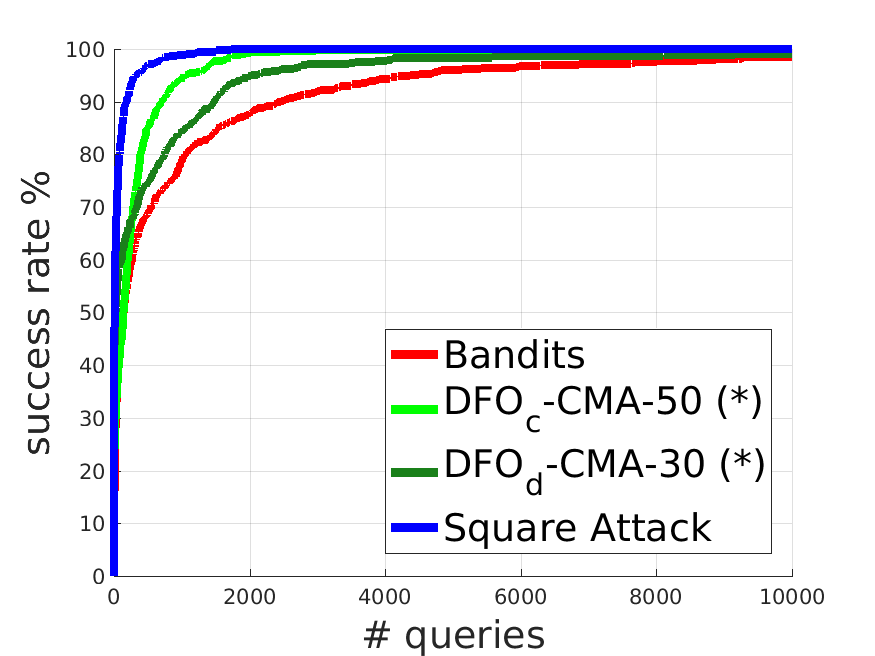}&
				\includegraphics[align=c, width=\newl]{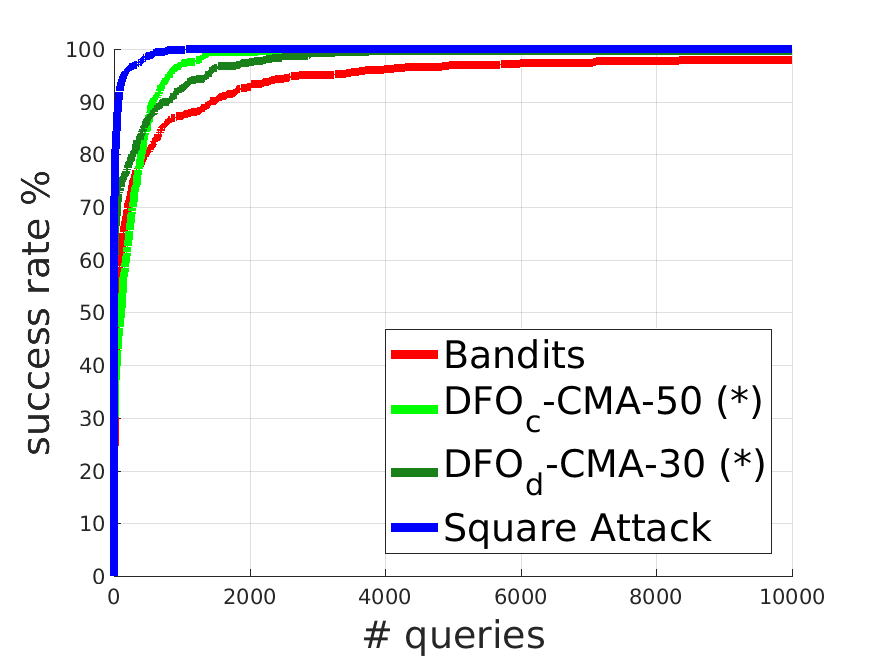}\\}
			\makecell{\textbf{$l_\infty$ attacks} \\ \textbf{low query regime}}&
			\includegraphics[align=c,width=\newl]{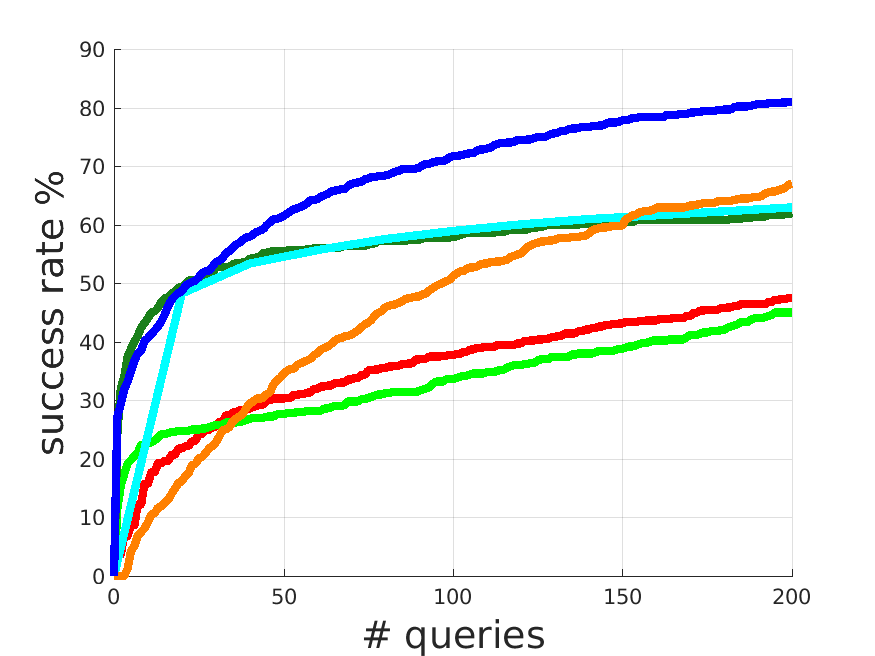}&
			\includegraphics[align=c, width=\newl]{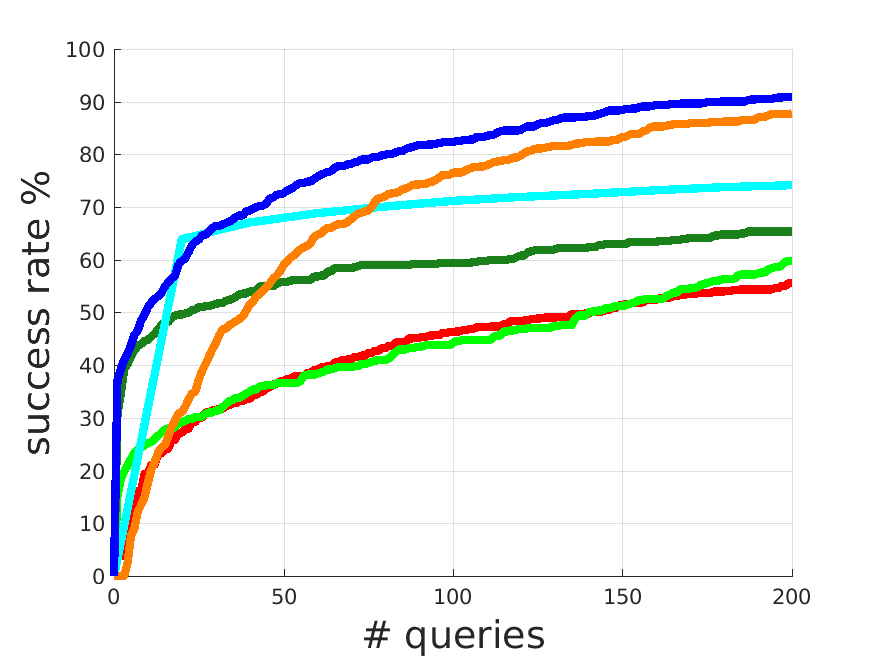}&
			\includegraphics[align=c, width=\newl]{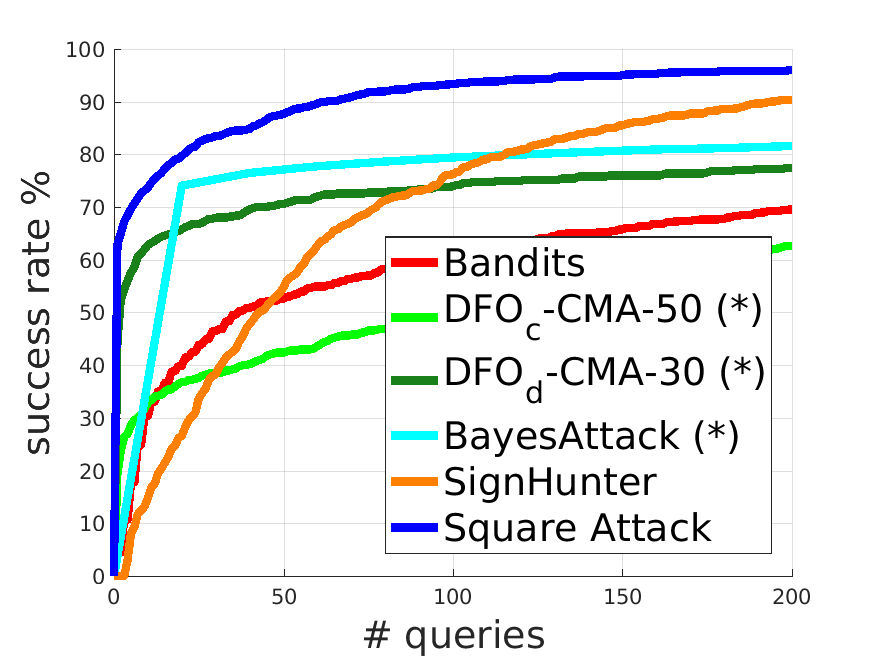}\\
			\makecell{\textbf{$l_2$-attacks} \\ \textbf{low query regime}}&
			\includegraphics[align=c, width=\newl]{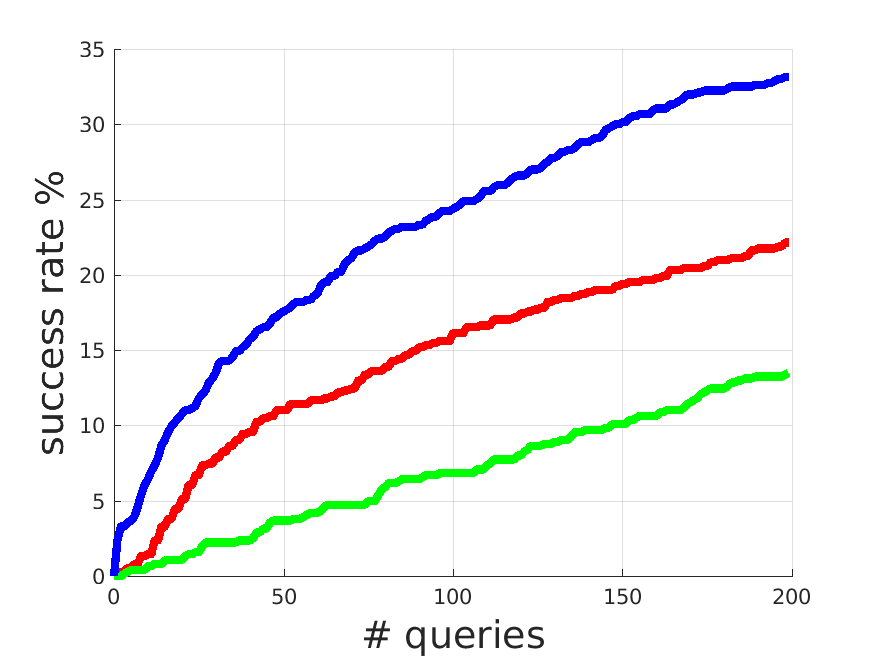}&
			\includegraphics[align=c, width=\newl]{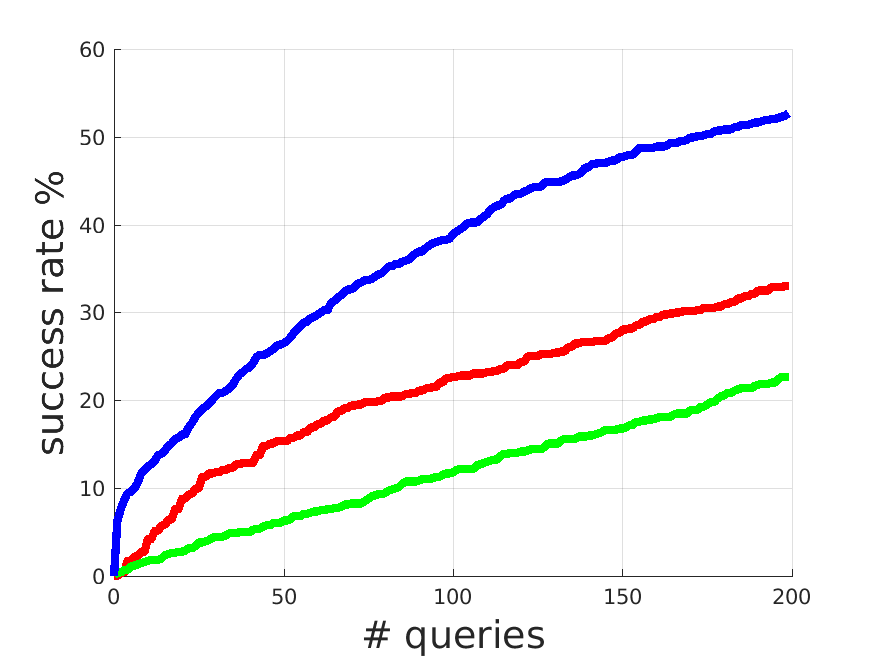}&
			\includegraphics[align=c, width=\newl]{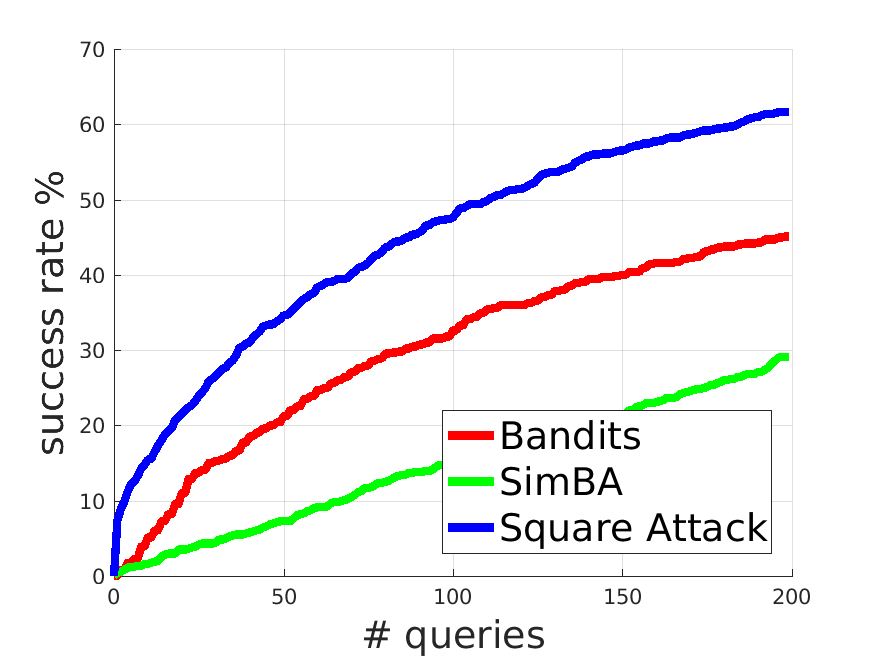}
		\end{tabular}
		\caption{Success rate in the low-query regime (up to 200 queries). $^*$ denotes the results obtained via personal communication with the authors and evaluated on 500 and 10,000 randomly sampled points for BayesAttack \cite{shukla2019blackbox} and DFO \cite{MeuEtAl2019} methods, respectively 
		}\label{fig:low_query} 
	\end{figure}

	Tables~\ref{tab:main_linf} and \ref{tab:jointly_succ_l2} show that the Square Attack, despite its simplicity, achieves in all the cases (models and norms) the \textbf{lowest failure rate}, ($<1\%$ everywhere except for the $l_2$-attack on Inception v3), and almost always requires  \textbf{fewer queries} than the competitors to succeed.
	Fig.~\ref{fig:low_query} shows the progression of the success rate of the attacks over the first 200 queries. Even in the low query regime the Square Attack outperforms the competitors for both norms. 
	Finally, we highlight that the only hyperparameter of our attack, $p$, regulating the size of the squares, is set for all the models to $0.05$ for $l_\infty$ and $0.1$ for $l_2$-perturbations.
	
	\customparagraph{$l_\infty$-attacks.}
	We compare our attack to Bandits~\cite{ilyas2019adversarial}, Parsimonious~\cite{MooEtAl2019}, $\textrm{DFO}_c$ / $\textrm{DFO}_d$~\cite{MeuEtAl2019}, and SignHunter~\cite{AlDujaili2019ThereAN}. 
	In Table \ref{tab:main_linf} we report the results of the $l_\infty$-attacks with norm bound of $\epsilon=0.05$.
	The Square Attack always has the lowest failure rate, notably 0.0\% in 2 out of 3 cases, and
	the lowest query consumption.
	Interestingly, our attack has median equal 1 on VGG-16-BN, meaning that the proposed initialization is particularly effective for this model.
	
	\begin{wraptable}{r}{0.64\columnwidth}
		\caption{Query statistics for untargeted $l_2$-attacks computed for the points for which all three attacks are successful for fair comparison}\label{tab:jointly_succ_l2}
		\centering\small
		\setlength{\tabcolsep}{4pt}
		\begin{tabular}{c ccc  |ccc}\hline\multirow{2}{*}{\textbf{Attack}} & \multicolumn{3}{c}{\textbf{Avg. queries}} & \multicolumn{3}{c}{\textbf{Med. queries}}\\ \cmidrule{2-4} \cmidrule{5-7}  & I & R & V & I & R & V\\
			\hline
			Bandits \cite{ilyas2019prior}  & 536 & 635 & 398 & 368 & 314 & 177\\
			SimBA-DCT \cite{guo2019simple} & 647 & 563 & 421 & 552 & 446 & 332\\
			\textbf{Square Attack} & \textbf{352}& \textbf{287} & \textbf{217} & \textbf{181}&\textbf{116} &\textbf{80}\\ \hline
		\end{tabular}
	\end{wraptable}
	The closest competitor in terms of the \textit{average} number of queries is SignHunter \cite{AlDujaili2019ThereAN}, which still needs on average between 1.8 and 3 times more queries to find adversarial examples and has a higher failure rate than our attack. Moreover, the median number of queries of SignHunter is much worse than for our method (e.g. 43 vs 1 on VGG).
	We note that although $\textrm{DFO}_c$--CMA \cite{MeuEtAl2019} is competitive to our attack in terms of \textit{median} queries, 
	it has a significantly higher failure rate and between 2 and 7 times worse average number of queries.
	Additionally, our method is also more effective in the low-query regime (Fig.~\ref{fig:low_query}) than other methods (including \cite{shukla2019blackbox}) on all the models.
	
	\customparagraph{$l_2$-attacks.}
	We compare our attack to Bandits \cite{ilyas2019prior} and SimBA \cite{guo2019simple} for $\epsilon=5$, while we do not consider SignHunter \cite{AlDujaili2019ThereAN} since it is not as competitive as for the $l_\infty$-norm, and in particular worse than Bandits on ImageNet (see Fig.~2 in \cite{AlDujaili2019ThereAN}).
	% We consider the SimBA attack successful in the same setting as all other attacks, i.e., when the $l_2$-norm of the adversarial perturbation is smaller than $\epsilon$ (we set $\epsilon=5$).

	As Table~\ref{tab:main_linf} and Fig.~\ref{fig:low_query} show, the Square Attack outperforms by a large margin the other methods in terms of failure rate,
	and achieves the lowest median number of queries for all the models and the lowest average one for VGG-16-BN.
	However, since it has a significantly lower failure rate, the statistics of the Square Attack are biased by the "hard" cases where the competitors fail. Then, we recompute the same statistics
	%about query consumption
	considering only the points where all the attacks are successful (Table \ref{tab:jointly_succ_l2}). In this case, our method improves by at least $1.5\times$ the average and by at least $2\times$ the median number of queries. % for the same set of images.

	\subsection{Square Attack Can be More Accurate than White-box Attacks}
	\label{sec:challenging_tasks}
	
	\begin{wraptable}{R}{0.48\columnwidth}\centering \small
		\setlength{\tabcolsep}{4.5pt}
		\caption{On the robust models of \cite{madry2018towards} and \cite{ZhaEtAl2019} on MNIST  $l_\infty$-Square Attack with $\epsilon=0.3$ achieves state-of-the-art (SOTA) results outperforming white-box attacks\label{tab:adv_training_evaluation_linf}}
		\begin{tabular}{c |c c}
			\hline
			\multirow{2}{*}{\textbf{Model}} & \multicolumn{2}{c}{\textbf{Robust accuracy}}\\ & SOTA & \textbf{Square}\\ \hline
			Madry et al. \cite{madry2018towards} &\textbf{88.13\%}&88.25\% \\ TRADES \cite{ZhaEtAl2019} & 93.33\% &\textbf{92.58\%} \\ \hline
		\end{tabular}
	\end{wraptable}
	%Here we show that the Square Attack can outperform white-box attacks on problems challenging for white-box PGD and other black-box methods. Since we consider both SOTA robust models and 
	Here we test our attack on problems which are challenging for both white-box PGD and other black-box attacks. 
	We use for evaluation \textit{robust accuracy}, defined as the worst-case accuracy of a classifier when an attack perturbs each input in some $l_p$-ball. %of radius $\epsilon$.
	We show that our algorithm outperforms the competitors both on state-of-the-art robust models and defenses that induce different types of gradient masking. Thus, our attack is useful to evaluate robustness
	% , being effective in a variety of situations without particular modifications and 
	without introducing adaptive attacks designed for each model separately.

	\customparagraph{Outperforming white-box attacks on robust models.}
	The models obtained with the adversarial training of \cite{madry2018towards} and TRADES \cite{ZhaEtAl2019} are standard benchmarks to test adversarial attacks, %and compare defenses,
	which means that many papers have tried to reduce their robust accuracy, without limit on the computational budget and primarily via white-box attacks. We test our $l_\infty$-Square Attack on these robust models on MNIST at $\epsilon=0.3$, using $p=0.8$, 20k queries and 50 random restarts, i.e., we run our attack 50 times and consider it successful if any of the runs finds an adversarial example (Table~\ref{tab:adv_training_evaluation_linf}). On the model of Madry et al \cite{madry2018towards} Square Attack is only $0.12\%$ far from the \textit{white-box} state-of-the-art, achieving the second best result %among \textit{all} attacks 
	(also outperforming the 91.47\% of SignHunter \cite{AlDujaili2019ThereAN} by a large margin).
	On the TRADES benchmark \cite{ZheEtAl2019}, our method obtains a new SOTA of $92.58\%$ robust accuracy outperforming the white-box attack of \cite{CroHei2019}.
	Additionally, the subsequent work of \cite{croce2020reliable} uses the Square Attack as part of their \textit{AutoAttack} where they show that the Square Attack outperforms other white-box attacks on 9 out of 9 MNIST models they evaluated.
	Thus, our black-box attack can be also useful for robustness evaluation of new defenses in the setting where gradient-based attacks require many restarts and iterations. % to be successful.

	\customparagraph{Resistance to gradient masking.} In Table~\ref{tab:adv_training_evaluation_l2} we report the robust accuracy at different thresholds $\epsilon$ of the $l_\infty$-adversarially trained models on MNIST of \cite{madry2018towards} for the $l_2$-threat model. It is known that the PGD
	is ineffective since it suffers from gradient masking \cite{TraBon2019}.
	Unlike PGD and other black-box attacks, our Square Attack does not suffer from gradient masking and yields robust accuracy close to zero for $\epsilon=2.5$, with only a single run. Moreover, the $l_2$-version of SignHunter~\cite{AlDujaili2019ThereAN} fails to accurately assess the robustness because the method optimizes only over the extreme points of the $l_\infty$-ball of radius $\epsilon / \sqrt{d}$ embedded in the target $l_2$-ball.
	
	\begin{table}[t] \caption{\label{tab:adv_training_evaluation_l2}
			$l_2$-robustness  of the $l_\infty$-adversarially trained models of \cite{madry2018towards} at different thresholds $\epsilon$. PGD is shown with 1, 10, 100 random restarts. The black-box attacks are given a 10k queries budget (see the supplement for details)}
		\centering
		{\small
			\setlength{\tabcolsep}{4.3pt}
			\begin{tabular}{c| ccc | cccc}
				\hline
				\multirow{3}{*}{$\epsilon_2$}& \multicolumn{7}{c}{\textbf{Robust accuracy}}\\
				& \multicolumn{3}{c|}{\textbf{White-box}} & \multicolumn{4}{c}{\textbf{Black-box}}\\
				%\cmidrule{2-4}
				& PGD$_1$ &  PGD$_{10}$  &PGD$_{100}$ & SignHunter & Bandits & SimBA & \textbf{Square}\\
				%&1 & 10 & 100& &&\\
				\hline
				2.0&79.6\%&67.4\%&\textbf{59.8\%}&95.9\%&80.1\%&87.6\%&\textbf{16.7\%} \\
				2.5&69.2\%&51.3\%&\textbf{36.0\%}&94.9\%&32.4\%&75.8\%&\textbf{2.4\%} \\
				3.0&57.6\%&29.8\%&\textbf{12.7\%}&93.8\%&12.5\%&58.1\%&\textbf{0.6\%}\\
				\hline
	\end{tabular}}\end{table}

	\begin{table}[t] \small\centering 
		\setlength{\tabcolsep}{1.5pt}
		\caption{$l_\infty$-robustness of Clean Logit Pairing (CLP), Logit Squeezing (LSQ) \cite{kannan2018adversarial}. The Square Attack is competitive to white-box PGD with many restarts (R=10,000, R=100 on MNIST, CIFAR-10 resp.) 
			and more effective than black-box attacks  \cite{ilyas2019prior,AlDujaili2019ThereAN}
		} \label{tab:logit_squeezing}
		\begin{tabular}{cc | cc | ccc}
			\hline
			\multirow{3}{*}{$\epsilon_\infty$}&\multirow{3}{*}{\textbf{Model}} & \multicolumn{4}{c}{\textbf{Robust accuracy}} \\
			&& \multicolumn{2}{c|}{\textbf{White-box}} & \multicolumn{3}{c}{\textbf{Black-box}} \\
			&& PGD\textsubscript{1}   & PGD\textsubscript{R} & Bandits & SignHunter & \textbf{Square} \\
			\hline
			\multirow{2}{*}{$0.3$}&CLP\textsubscript{MNIST} &   62.4\% & \textbf{4.1\%} & 33.3\% & 62.1\% & \textbf{6.1\%} \\
			&LSQ\textsubscript{MNIST} &   70.6\% & \textbf{5.0\%}          &  37.3\% & 65.7\% &\textbf{2.6\%} \\
			%		\hline
			%\hdashline
			\multirow{2}{*}{$16/255$}&CLP\textsubscript{CIFAR} &   2.8\%  & \textbf{0.0\%} & 14.3\% & \textbf{0.1\%} & 0.2\% \\
			&LSQ\textsubscript{CIFAR} &   27.0\% & \textbf{1.7\%} & 27.7\% & 13.2\% & \textbf{7.2\%} \\
			\hline
		\end{tabular}
	\end{table}
	\customparagraph{Attacking Clean Logit Pairing and Logit Squeezing.}
	These two $l_\infty$ defenses proposed in \cite{kannan2018adversarial} were broken in \cite{MosEtAl18}. However, \cite{MosEtAl18} needed up to 10k restarts of PGD which is computationally prohibitive. Using the publicly available models from \cite{MosEtAl18}, we run the Square Attack with $p=0.3$ and 20k query limit (results in Table~\ref{tab:logit_squeezing}). We obtain robust accuracy similar to PGD\textsubscript{R} in most cases, but with a \textit{single run}, i.e. without additional restarts. %, outpeforming Bandits. 
	At the same time, although on some models Bandits and SignHunter outperform PGD\textsubscript{1}, they on average
	achieve significantly worse results than the Square Attack. % the CLP\textsubscript{MNIST} and LSQ\textsubscript{MNIST} models. 
	This again shows the utility of the Square Attack %robustness evaluation,
	%when even PGD fails
	to accurately assess robustness.

	\section{Conclusion}
	We have presented a simple black-box attack which outperforms by a large margin the state-of-the-art both in terms of query efficiency and success rate. Our results suggest that our attack is useful \textit{even in comparison to white-box attacks} to better estimate the robustness of models that exhibit gradient masking. 
	
	\ \\
%	In future work, it would be interesting to further analyze the sensitivity of CNNs to square-shaped and high-frequency perturbations.\\
	\textbf{Acknowledgements.} We thank L. Meunier and S. N. Shukla for providing the data for Figure~\ref{fig:success_rate_queries}. M.A. thanks A. Modas for fruitful discussions.
	% M.H. and F.C. acknowledge support from  Tue.AI Center (FKZ: 01IS18039A), the DFG TRR 248, project number 389792660 and the DFG Excellence Cluster “Machine Learning - New Perspectives for Science”, EXC 2064/1, project number 390727645.
	M.H and F.C. acknowledge support by the Tue.AI Center (FKZ: 01IS18039A), DFG TRR 248, project number 389792660 and DFG EXC 2064/1, project number 390727645. 
	
	% \section*{Acknowledgements} We are very grateful to Laurent Meunier and Satya Narayan Shukla for providing the data for Figure~\ref{fig:success_rate_queries}. M.A. also thanks Apostolos Modas for fruitful discussions.
	
	% M.H. and F.C. acknowledge support from the BMBF through the T{\"u}bingen AI Center (FKZ: 01IS18039A), the DFG TRR 248, project number 389792660 and the DFG Excellence Cluster “Machine Learning - New Perspectives for Science”, EXC 2064/1, project number 390727645.
	
	\bibliographystyle{splncs04}
	\bibliography{literature}

\begin{thebibliography}{10}
\providecommand{\url}[1]{\texttt{#1}}
\providecommand{\urlprefix}{URL }
\providecommand{\doi}[1]{https://doi.org/#1}

\bibitem{AkhtarARXIV2018}
Akhtar, N., Mian, A.: Threat of adversarial attacks on deep learning in
  computer vision: A survey. IEEE Access  \textbf{6},  14410--14430 (2018)

\bibitem{AlDujaili2019ThereAN}
Al-Dujaili, A., O'Reilly, U.M.: There are no bit parts for sign bits in
  black-box attacks. In: ICLR (2020)

\bibitem{alzantot2018genattack}
Alzantot, M., Sharma, Y., Chakraborty, S., Srivastava, M.: Genattack: practical
  black-box attacks with gradient-free optimization. In: Genetic and
  Evolutionary Computation Conference (GECCO) (2019)

\bibitem{AthEtAl2018}
Athalye, A., Carlini, N., Wagner, D.A.: Obfuscated gradients give a false sense
  of security: Circumventing defenses to adversarial examples. In: ICML (2018)

\bibitem{BasEtAl2016}
Bastani, O., Ioannou, Y., Lampropoulos, L., Vytiniotis, D., Nori, A.,
  Criminisi, A.: Measuring neural net robustness with constraints. In: NeurIPS
  (2016)

\bibitem{bhagoji2018practical}
Bhagoji, A.N., He, W., Li, B., Song, D.: Practical black-box attacks on deep
  neural networks using efficient query mechanisms. In: ECCV (2018)

\bibitem{BiggioPR2018}
Biggio, B., Roli, F.: Wild patterns: Ten years after the rise of adversarial
  machine learning. Pattern Recognition  \textbf{84},  317--331 (2018)

\bibitem{BoyVan2004}
Boyd, S., Vandenberghe, L.: Convex Optimization. Cambridge University Press,
  Cambridge (2004)

\bibitem{brendel2017decision}
Brendel, W., Rauber, J., Bethge, M.: Decision-based adversarial attacks:
  Reliable attacks against black-box machine learning models. In: ICLR (2018)

\bibitem{BroEtAl2017}
Brown, T.B., Man{\'e}, D., Roy, A., Abadi, M., Gilmer, J.: Adversarial patch.
  In: NeurIPS 2017 Workshop on Machine Learning and Computer Security (2017)

\bibitem{brunner2018guessing}
Brunner, T., Diehl, F., Le, M.T., Knoll, A.: Guessing smart: biased sampling
  for efficient black-box adversarial attacks. In: ICCV (2019)

\bibitem{CarWag2017}
Carlini, N., Wagner, D.: Adversarial examples are not easily detected:
  Bypassing ten detection methods. In: ACM Workshop on Artificial Intelligence
  and Security (2017)

\bibitem{chen2019boundary}
Chen, J., Jordan, M.I., J., W.M.: Hop{S}kip{J}ump{A}ttack: a query-efficient
  decision-based attack (2019), arXiv preprint arXiv:1904.02144

\bibitem{CheEtAl2018}
Chen, P., Sharma, Y., Zhang, H., Yi, J., Hsieh, C.: Ead: Elastic-net attacks to
  deep neural networks via adversarial examples. In: AAAI (2018)

\bibitem{chen2017zoo}
Chen, P.Y., Zhang, H., Sharma, Y., Yi, J., Hsieh, C.J.: Zoo: Zeroth order
  optimization based black-box attacks to deep neural networks without training
  substitute models. In: 10th ACM Workshop on Artificial Intelligence and
  Security - AISec ’17. ACM Press (2017)

\bibitem{cheng2018query}
Cheng, M., Le, T., Chen, P.Y., Yi, J., Zhang, H., Hsieh, C.J.: Query-efficient
  hard-label black-box attack: An optimization-based approach. In: ICLR (2019)

\bibitem{cheng2019improving}
Cheng, S., Dong, Y., Pang, T., Su, H., Zhu, J.: Improving black-box adversarial
  attacks with a transfer-based prior. In: NeurIPS (2019)

\bibitem{cohen2019certified}
Cohen, J.M., Rosenfeld, E., Kolter, Z.: Certified adversarial robustness via
  randomized smoothing. In: ICML (2019)

\bibitem{croce2019sparse}
Croce, F., Hein, M.: Sparse and imperceivable adversarial attacks. In: ICCV
  (2019)

\bibitem{CroHei2019}
Croce, F., Hein, M.: Minimally distorted adversarial examples with a fast
  adaptive boundary attack. In: ICML (2020)

\bibitem{croce2020reliable}
Croce, F., Hein, M.: Reliable evaluation of adversarial robustness with an
  ensemble of diverse parameter-free attacks. In: ICML (2020)

\bibitem{davis2019stochastic}
Davis, D., Drusvyatskiy, D.: Stochastic model-based minimization of weakly
  convex functions. SIAM Journal on Optimization  \textbf{29}(1),  207--239
  (2019)

\bibitem{du2019query}
Du, J., Zhang, H., Zhou, J.T., Yang, Y., Feng, J.: Query-efficient meta attack
  to deep neural networks. In: ICLR (2020)

\bibitem{duchi2015optimal}
Duchi, J., Jordan, M., Wainwright, M., Wibisono, A.: Optimal rates for
  zero-order convex optimization: The power of two function evaluations. IEEE
  Transactions on Information Theory  \textbf{61}(5),  2788--2806 (2015)

\bibitem{fawzi2016measuring}
Fawzi, A., Frossard, P.: Measuring the effect of nuisance variables on
  classifiers. In: British Machine Vision Conference (BMVC) (2016)

\bibitem{GuRig2015}
Gu, S., Rigazio, L.: Towards deep neural network architectures robust to
  adversarial examples. In: ICLR Workshop (2015)

\bibitem{guo2018low}
Guo, C., Frank, J.S., Weinberger, K.Q.: Low frequency adversarial perturbation.
  In: UAI (2019)

\bibitem{guo2019simple}
Guo, C., Gardner, J.R., You, Y., Wilson, A.G., Weinberger, K.Q.: Simple
  black-box adversarial attacks. In: ICML (2019)

\bibitem{haagerup1981best}
Haagerup, U.: The best constants in the {K}hintchine inequality. Studia Math.
  \textbf{70}(3),  231--283 (1981)

\bibitem{ilyas2018black}
Ilyas, A., Engstrom, L., Athalye, A., Lin, J.: Black-box adversarial attacks
  with limited queries and information. In: ICML (2018)

\bibitem{ilyas2019prior}
Ilyas, A., Engstrom, L., Madry, A.: Prior convictions: Black-box adversarial
  attacks with bandits and priors. In: ICLR (2019)

\bibitem{ilyas2019adversarial}
Ilyas, A., Santurkar, S., Tsipras, D., Engstrom, L., Tran, B., Madry, A.:
  Adversarial examples are not bugs, they are features. NeurIPS  (2019)

\bibitem{kannan2018adversarial}
Kannan, H., Kurakin, A., Goodfellow, I.: Adversarial logit pairing (2018),
  arXiv preprint arXiv:1803.06373

\bibitem{karmon2018lavan}
Karmon, D., Zoran, D., Goldberg, Y.: Lavan: Localized and visible adversarial
  noise. In: ICML (2018)

\bibitem{li2019nattack}
Li, Y., Li, L., Wang, L., Zhang, T., Gong, B.: Nattack: Learning the
  distributions of adversarial examples for an improved black-box attack on
  deep neural networks. In: ICML (2019)

\bibitem{lin2019bandlimiting}
Lin, Y., Jiang, H., Jiang, H.: Bandlimiting neural networks against adversarial
  attacks (2019), arXiv preprint arXiv:1905.12797

\bibitem{madry2018towards}
Madry, A., Makelov, A., Schmidt, L., Tsipras, D., Vladu, A.: Towards deep
  learning models resistant to adversarial attacks. In: ICLR (2018)

\bibitem{matyas1965random}
Matyas, J.: Random optimization. Automation and Remote control  \textbf{26}(2),
   246--253 (1965)

\bibitem{MeuEtAl2019}
{Meunier}, L., {Atif}, J., {Teytaud}, O.: Yet another but more efficient
  black-box adversarial attack: tiling and evolution strategies (2019), arXiv
  preprint, arXiv:1910.02244

\bibitem{MosEtAl18}
Mosbach, M., Andriushchenko, M., Trost, T., Hein, M., Klakow, D.: Logit pairing
  methods can fool gradient-based attacks. In: NeurIPS 2018 Workshop on
  Security in Machine Learning (2018)

\bibitem{narodytska2017simple}
Narodytska, N., Kasiviswanathan, S.: Simple black-box adversarial attacks on
  deep neural networks. In: CVPR Workshops (2017)

\bibitem{NemYud83}
Nemirovsky, A.S., Yudin, D.B.: {Problem Complexity and Method Efficiency in
  Optimization}. Wiley-Interscience Series in Discrete Mathematics, John Wiley
  \&\ Sons (1983)

\bibitem{nesterov2017random}
Nesterov, Y., Spokoiny, V.: Random gradient-free minimization of convex
  functions. Foundations of Computational Mathematics  \textbf{17}(2),
  527--566 (2017)

\bibitem{papernot2016transferability}
Papernot, N., McDaniel, P., Goodfellow, I.: Transferability in machine
  learning: from phenomena to black-box attacks using adversarial samples
  (2016), arXiv preprint arXiv:1605.07277

\bibitem{PapEtAl2016a}
Papernot, N., McDaniel, P., Wu, X., Jha, S., Swami, A.: Distillation as a
  defense to adversarial perturbations against deep networks. In: IEEE
  Symposium on Security \& Privacy (2016)

\bibitem{rastrigin1963convergence}
Rastrigin, L.: The convergence of the random search method in the extremal
  control of a many parameter system. Automaton \& Remote Control  \textbf{24},
   1337--1342 (1963)

\bibitem{schrack1976optimized}
Schrack, G., Choit, M.: Optimized relative step size random searches.
  Mathematical Programming  \textbf{10},  230--244 (1976)

\bibitem{schumer1968adaptive}
Schumer, M., Steiglitz, K.: Adaptive step size random search. IEEE Transactions
  on Automatic Control  \textbf{13}(3),  270--276 (1968)

\bibitem{MooEtAl2019}
Seungyong, M., Gaon, A., Hyun, O.S.: Parsimonious black-box adversarial attacks
  via efficient combinatorial optimization. In: ICML (2019)

\bibitem{shukla2019blackbox}
Shukla, S.N., Sahu, A.K., Willmott, D., Kolter, Z.: Black-box adversarial
  attacks with {B}ayesian optimization (2019), arXiv preprint arXiv:1909.13857

\bibitem{SuVarKou19}
Su, J., Vargas, D., Sakurai, K.: One pixel attack for fooling deep neural
  networks. IEEE Transactions on Evolutionary Computation  (2019)

\bibitem{suya2019hybrid}
Suya, F., Chi, J., Evans, D., Tian, Y.: Hybrid batch attacks: Finding black-box
  adversarial examples with limited queries (2019), arXiv preprint,
  arXiv:1908.07000

\bibitem{TraBon2019}
Tramèr, F., Boneh, D.: Adversarial training and robustness for multiple
  perturbations. In: NeurIPS (2019)

\bibitem{tsipras2019robustness}
Tsipras, D., Santurkar, S., Engstrom, L., Turner, A., Madry, A.: Robustness may
  be at odds with accuracy. In: ICLR (2019)

\bibitem{tu2019autozoom}
Tu, C.C., Ting, P., Chen, P.Y., Liu, S., Zhang, H., Yi, J., Hsieh, C.J., Cheng,
  S.M.: Autozoom: Autoencoder-based zeroth order optimization method for
  attacking black-box neural networks. In: AAAI Conference on Artificial
  Intelligence (2019)

\bibitem{uesato2018adversarial}
Uesato, J., O'Donoghue, B., Van~den Oord, A., Kohli, P.: Adversarial risk and
  the dangers of evaluating against weak attacks. In: ICML (2018)

\bibitem{yan2019subspace}
Yan, Z., Guo, Y., Zhang, C.: Subspace attack: Exploiting promising subspaces
  for query-efficient black-box attacks. In: NeurIPS (2019)

\bibitem{yin2019fourier}
Yin, D., Lopes, R.G., Shlens, J., Cubuk, E.D., Gilmer, J.: A {F}ourier
  perspective on model robustness in computer vision. In: NeurIPS (2019)

\bibitem{yu2017dilated}
Yu, F., Koltun, V., Funkhouser, T.: Dilated residual networks. In: CVPR (2017)

\bibitem{zabinsky2010random}
Zabinsky, Z.B.: Random search algorithms. Wiley encyclopedia of operations
  research and management science  (2010)

\bibitem{ZhaEtAl2019}
Zhang, H., Yu, Y., Jiao, J., Xing, E.P., Ghaoui, L.E., Jordan, M.I.:
  Theoretically principled trade-off between robustness and accuracy. In: ICML
  (2019)

\bibitem{ZheEtAl2016}
Zheng, S., Song, Y., Leung, T., Goodfellow, I.J.: Improving the robustness of
  deep neural networks via stability training. In: CVPR (2016)

\bibitem{ZheEtAl2019}
Zheng, T., Chen, C., Ren, K.: Distributionally adversarial attack. In: AAAI
  (2019)

\end{thebibliography}
	
	\clearpage

	\appendix
	\begin{center}
		\Large\textbf{Supplementary Material}
	\end{center}
	
	\section*{Organization of the Supplementary Material}
	
	In Section~\ref{app:secproof}, we present the missing proofs of Section~\ref{sec:attack} and Section~\ref{sec:theory} and slightly deepen our theoretical insights on the efficiency of the proposed $l_\infty$-attack.
	Section~\ref{app:exp_details} covers various implementation details and the hyperparameters we used. 
	We show a more detailed ablation study on different choices of the attack's algorithm in Section~\ref{app:ablation_study}.
	Since the Square Attack is a randomized algorithm, we show the variance of the main reported metric for different random seeds in Section~\ref{app:stability}.
	Finally, Section~\ref{app:additional_exp_results} presents results of the targeted attacks on ImageNet, additional results for the untargeted attacks, and an evaluation of the post-averaging defense \cite{lin2019bandlimiting} which we conclude is much less robust than claimed.
	
	% This part goes to the appendix
	\section{Proofs Omitted from Section~\ref{sec:attack} and Section~\ref{sec:theory}}
	In this section, we present the proofs omitted from Section~\ref{sec:attack} and Section~\ref{sec:theory}.
	\label{app:secproof}
	
	\subsection{Proof of Proposition~\ref{pro:l2}}
	\label{app:prol2}
	Let $\delta$ be the output of Algorithm~\ref{alg:l2_sampling_distribution}. We prove here that $\norm{\hat{x} + \delta - x}_2=\epsilon$.

	From Step 13 of Algorithm~\ref{alg:l2_sampling_distribution}, we directly have the equality $\norm{\hat{x} + \delta - x}_2=\norm{\nu}_2$.
	Let $\nu^\textrm{old}$ be the update at the previous iteration, defined in Step 1 and $\overline{W_1 \cup W_2}$ the indices not belonging to $W_1\cup W_2$.
	Then, \begin{equation*}
	\begin{split} \norm{\nu}_2^2 &= \sum_{i=1}^{c}\norm{\nu_{W_1 \cup W_2,i}}_2^2 + \sum_{i=1}^{c}\norm{\nu_{\overline{W_1 \cup W_2},i}}_2^2\\ & = \sum_{i=1}^{c}\norm{\nu_{W_1,i}}_2^2 + \sum_{i=1}^{c}\norm{\nu_{\overline{W_1 \cup W_2},i}}_2^2\\ 
	&= \sum_{i=1}^{c}(\epsilon^i_{\textrm{avail}})^2 + \sum_{i=1}^{c}\norm{\nu_{\overline{W_1 \cup W_2},i}}_2^2\\
	&=\sum_{i=1}^{c}\norm{\nu^\textrm{old}_{W_1 \cup W_2,i}}_2^2 + \epsilon_{\textrm{unused}}^2 + \sum_{i=1}^{c}\norm{\nu_{\overline{W_1 \cup W_2},i}}_2^2\\
	&\overset{(i)}{=} \sum_{i=1}^{c}\norm{\nu^\textrm{old}_{W_1 \cup W_2,i}}_2^2 + \epsilon_{\textrm{unused}}^2 + \sum_{i=1}^{c}\norm{\nu^\textrm{old}_{\overline{W_1 \cup W_2},i}}_2^2\\
	&= \norm{\nu^\textrm{old}}_2^2 +\epsilon_{\textrm{unused}}^2 \overset{(ii)}{=} \epsilon^2,
	\end{split}
	\end{equation*}
	where $(i)$ holds since $\nu^\textrm{old}_{\overline{W_1\cup W_2}} \equiv \nu_{\overline{W_1\cup W_2}}$ as the modifications affect only the elements in the two windows, and $(ii)$ holds by the definition of $\epsilon_{\textrm{unused}}$ in Step 4 of Algorithm~\ref{alg:l2_sampling_distribution}.
	
	\subsection{Proof of Proposition~\ref{pro:conv}}
	
	Using the $L$-smoothness of the function $g$, that is it holds for all $x,y \in \R^d$,
	\[ \norm{\nabla g(x)-\nabla g(y)}_2 \leq L \norm{x-y}_2.\]
	we obtain (see e.g. \cite{BoyVan2004}):
	\[
	g(x_t+\delta_t)\leq g(x_t)+\langle \nabla g(x_t),\delta_t\rangle +\frac{L}{2}\Vert \delta_t\Vert_2^2,
	\]
	and by definition of $x_{t+1}$ we have 
	\begin{align*}
	g(x_{t+1})&\leq \min\{ g(x_t),g(x_t+\delta_t) \}\\
	&\leq g(x_t)+\min \{ 0,\langle \nabla g(x_t),\delta_t\rangle +\frac{L}{2}\Vert \delta_t\Vert_2^2 \}.
	\end{align*}
	Using the definition of the $\min$ as a function of the absolute value ($2\min\{a,b\}=a+b-|a-b|$) yields
	\begin{align*}
	g(x_{t+1})&\leq g(x_t)+\frac{1}{2}\langle \nabla g(x_t),\delta_t\rangle +\frac{L}{4}\Vert \delta_t\Vert_2^2 %\\
	%&\quad
	-\frac{1}{2}|\langle \nabla g(x_t),\delta_t\rangle +\frac{L}{2}\Vert \delta_t\Vert_2^2 | .
	\end{align*}
	And using the triangular inequality ($|a+b|\geq |a|-|b|$), we have
	\begin{align*}
	g(x_{t+1})&\leq g(x_t)+\frac{1}{2}\langle \nabla g(x_t),\delta_t\rangle +\frac{L}{2}\Vert \delta_t\Vert_2^2 -\frac{1}{2}|\langle \nabla g(x_t),\delta_t\rangle|  .
	\end{align*}
	Therefore taking the expectation and using that $\mathbb E \delta_t=0$, we get 
	\[
	\mathbb E g(x_{t+1})\leq \mathbb E g(x_t) -\frac{1}{2} \mathbb E |\langle \nabla g(x_t),\delta_t\rangle|+\frac{L}{2}\mathbb E \Vert\delta_t\Vert_2^2.
	\]
	Therefore, together with the assumptions in Eq.~\eqref{eq:asspt} this yields to 
	\[
	\mathbb E g(x_{t+1})\leq \mathbb E g(x_t) -\frac{\tilde C \gamma_t}{2} \mathbb E\| \nabla g(x_t)\|_2+\frac{LC\gamma_t^2}{2}.
	\]
	and thus
	\[ \mathbb E \norm{\nabla g(x_t)}_2 \leq \frac{2}{\gamma_t \tilde{C}}\Big(\mathbb E g(x_t) - \mathbb E g(x_{t+1})+\frac{LC\gamma_t^2}{2}\Big).\]
	Thus for $\gamma_t=\gamma$ we have summing for $t=0:T$
	%
	%\begin{align*}
	%\min_{0\leq i \leq T}  \| \nabla g(x_i)\|_2 &\leq %\frac{1}{T}\sum_{t=0}^T
	%\| \nabla g(x_t)\|_2 \\
	%&\leq \frac{2}{\tilde C \gamma T}\big[g(x_0)- g(x_\ast)+\frac{T L %C\gamma^2}{2}\big].
	%\end{align*}
	\begin{align*}
	\min_{0\leq i \leq T} \mathbb E \| \nabla g(x_i)\|_2 &\leq \frac{1}{T}\sum_{t=0}^T
	\mathbb E\| \nabla g(x_t)\|_2 \\
	&\leq \frac{2}{\tilde C \gamma T}\big[g(x_0)- \mathbb E g(x_{T+1})+\frac{T L C\gamma^2}{2}\big].
	\end{align*}
	We conclude setting the step-size to $\gamma=\Theta(1/\sqrt{T})$.

	\subsection{Assumptions in Eq.~\eqref{eq:asspt} Do Not Hold for the Sampling Distribution $P$}
	
	\label{app:assno}
	
	Let us consider an update $\delta$ with a window size $h=2$ and the direction $v\in\{-1,1\}^{w\times w \times c}$ defined as 
	\[
	v_{k,l}^i=(-1)^{kl} \quad \text{for all } i,k,l.
	\]
	It is easy to check that any update $\delta$ drawn from the sampling distribution $P$ is orthogonal to this direction $v$:
	\[\langle v,\delta\rangle =\sum_{i=1}^c \sum_{k=r+1}^{r+2}\sum_{l=s+1}^{s+2}(-1)^{kl}=c(-1+1-1+1)=0.\]
	Thus, $\mathbb E |\langle v,\delta\rangle|=0$ and the assumptions in Eq.~\eqref{eq:asspt} do not hold.
	This means that the convergence analysis does not directly hold for the sampling distribution $P$.

	\subsection{Assumptions in Eq.~\eqref{eq:asspt}  Hold for the Sampling Distribution $P^{\text{multiple}}$}
	
	\label{app:assyes}
	
	Let us consider the sampling distribution $P^{\text{multiple}}$  where different Rademacher $\rho_{k,l,i}$ are drawn for each pixel of the update window $\delta_{r+1: r+h,\;s+1:s+h,\;i}$. We present it in Algorithm~\ref{alg:mult} with the convention that any subscript $k>w$ should be understood as $k-w$. This technical modification is greatly helpful to avoid side effect.
	
	\begin{algorithm}[b]
		
		\caption{Sampling distribution $P^{\text{multiple}}$ for $l_\infty$-norm} 
		
		\label{alg:mult}
		
		\KwIn{maximal norm $\epsilon$, window size $h$, image size $w$, color channels $c$}
		
		\KwOut{New update $\delta$}
		
		$\delta \gets $ array of zeros of size $w\times w\times c$\\
		
		sample uniformly $r, s \in \{0, \ldots, w\} \subset \N$\\
		
		\For{$i=1,\ldots, c$}
		{%sample value
			$\delta_{r+1: r+h,\;s+1:s+h,\;i} \gets  Uniform(\{-2\epsilon, 2\epsilon\}^{h \times h})$ %\tcc{assign the same value to all locations}
		}
		
	\end{algorithm}
	
	Let $v\in \mathbb{R}^{w\times w \times c}$ for which we have using the Khintchine inequality~\cite{haagerup1981best}:
	
	\begin{align*}
	\mathbb E | \langle \delta,v\rangle| &=  \mathbb E | \sum_{k=r+1}^{r+h} \sum_{l=s+1}^{s+h}\sum_{i=1}^c  \delta_{k,l}^i v_{k,l}^i| \\
	&\overset{(i)}{=}  \mathbb E_{(r,s)} \mathbb E_\rho | \sum_{k=r+1}^{r+h} \sum_{l=s+1}^{s+h}\sum_{i=1}^c  \delta_{k,l}^i v_{k,l}^i| \\
	&\overset{(ii)}{\geq}  \frac{2\varepsilon}{\sqrt{2}} \mathbb E_{(r,s)} \| V_{(r,s)}\|_2 \\
	&\overset{(iii)}{\geq} \sqrt{2}\varepsilon\| \mathbb E_{(r,s)} V_{(r,s)}\|_2 \\
	&\geq  \frac{\sqrt{2}\varepsilon h^2}{w^2} \|  v\|_2,
	\end{align*}
	where we define by $V_{(r,s)}=\{v_{k,l}^i\}_{ k \in \{r+1,\dots,r+h\}, l\in\{s+1,\dots,s+h\}, i\in\{1,\dots,c\}}$ and $(i)$ follows from the decomposition between the randomness of the Rademacher and the random window, $(ii)$ follows from the Khintchine inequality and  $(iii)$ follows from Jensen inequality.
	
	In addition we have for the variance:
	\begin{align*}
	\mathbb E  \|\delta\|_2^2&=  \mathbb E_{(r,s)} \sum_{k=r+1}^{r+h} \sum_{l=s+1}^{s+h}\sum_{i=1}^c  \mathbb E_\rho (\delta_{k,l}^i)^2 \\
	&=  \mathbb E_{(r,s)} \sum_{k=r+1}^{r+h} \sum_{l=s+1}^{s+h}\sum_{i=1}^c 4\varepsilon^2 \\
	&= 4 c \varepsilon^2  h^2.
	\end{align*}
	Thus the assumptions in Eq.~\eqref{eq:asspt} hold for the sampling distribution $P^{\text{multiple}}$.
	
	\subsection{Why Updates of Equal Sign?}
	\label{app:why_equal_sign}
	Proposition~\ref{pro:conv} underlines the importance of a large inner product $\mathbb{E} [\vert \langle  \delta_t, \nabla g (x_t)\rangle \vert]$ in the direction of the gradients. This provides some intuition explaining why the update $\delta^{\text{single}}$, where a single Rademacher is drawn for each window,
	is more efficient than the update $\delta^{\text{multiple}}$ where different Rademacher values are drawn. Following the observation that the gradients are often approximately piecewise constant \cite{ilyas2019prior}, we consider, as a heuristic, a piecewise constant direction $v$ for which we will show that
	\[
	\mathbb{E} [\vert \langle  \delta^{\text{single}}, v\rangle \vert]=\Theta(\Vert v\Vert_1) \ \text{ and } \ \mathbb{E} [\vert \langle  \delta^{\text{multiple}}, v\rangle \vert] =\Theta(\Vert v\Vert_2 ).
	\]
	Therefore the directions sampled by our proposal are more correlated with the gradient direction and help the algorithm to converge faster. This is also verified empirically in our experiments (see the ablation study in Sup.~\ref{app:ablation_study}). 
	
	\customparagraph{Analysis.}
	Let us consider the direction $v\in \mathbb R ^{w\times w}$ composed of different blocks $\{V_{(r,s)}\}_{(r,s)\in \{0,\dots, w/h\}}$ of constant sign.
	
	For this direction $v$ we compare two different proposal $P^{\text{multiple}}$ and $P^{\text{single}}$ where we choose uniformly one random block $(r,s)$ and we either assign a single Rademacher $\rho_{(r,s)}$ to the whole block (this is $P^{\text{single}}$) or we assign multiple Rademacher values $\{\rho_{(k,l)}\}_{k\in \{rh+1,\dots, (r+1)h\}, l\in \{sh+1,\dots, (s+1)h\}}$ (this is $P^{\text{multiple}}$). Using the Khintchine and Jensen inequalities similarly to Sec.~\ref{app:assyes}, we have
	\begin{align*}
	\mathbb E | \langle \delta^{\text{multiple}},v\rangle| &=  \mathbb E | \sum_{k=rh+1}^{(r+1)h} \sum_{l=sh+1}^{(s+1)h}\delta_{k,l} v_{k,l}| \\
	&{\geq}  \frac{2\varepsilon}{\sqrt{2}} \mathbb E_{(r,s)} \| V_{(r,s)}\|_2\\
	% &{\geq}  \frac{\sqrt{2}\varepsilon h^2}{w^2} \sum_{r=1}^{w/h} \sum_{s=1}^{w/h}\| V_{(r,s)}\|_2.
	&\geq  \frac{\sqrt{2}\varepsilon h^2}{w^2} \|  v\|_2.
	\end{align*} 
	% Therefore we obtain the $l_1/l_2$-norm of the different groups.
	Moreover, we can show the following upper bound using the Khintchine inequality and the inequality between the $l_1$- and $l_2$-norms:
	\begin{align*}
	\mathbb E | \langle \delta^{\text{multiple}},v\rangle| &=  \mathbb E | \sum_{k=rh+1}^{(r+1)h} \sum_{l=sh+1}^{(s+1)h}\delta_{k,l} v_{k,l}| \\
	&\leq  2\varepsilon \mathbb E_{(r,s)} \| V_{(r,s)}\|_2\\
	&= \frac{2\varepsilon h^2}{w^2} \sum_{r=1}^{w/h} \sum_{s=1}^{w/h}\| V_{(r,s)}\|_2\\
	&\leq  \frac{2\varepsilon h}{w} \|  v\|_2
	\end{align*} 
	Thus, $\mathbb{E} [\vert \langle  \delta^{\text{multiple}}, v\rangle \vert] =\Theta(\Vert v\Vert_2 )$.
	
	For the update $\delta^{\text{single}}$ we obtain
	\begin{align*}
	\mathbb E | \langle \delta^{\text{single}},v\rangle| &=  \mathbb E | \sum_{k=rh+1}^{(r+1)h} \sum_{l=sh+1}^{(s+1)h}\delta_{r,s} v_{k,l}| \\
	&=  \mathbb E | \delta_{r,s}  \sum_{k=rh+1}^{(r+1)h} \sum_{l=sh+1}^{(s+1)h}v_{k,l}| \\
	&\overset{(i)}{=} 2\varepsilon  \mathbb E_{(r,s)} \| V_{(r,s)}\|_1   \\
	&=  \frac{2\varepsilon h^2}{w^2} \| v\|_1
	\end{align*} 
	where $(i)$ follows from the fact the $V_{(r,s)}$ has a constant sign. We recover then the $l_1$-norm of the direction $v$, i.e. we conclude that $\mathbb{E} [\vert \langle  \delta^{\text{single}}, v\rangle \vert]=\Theta(\Vert v\Vert_1)$.
	
	This implies that for an approximately constant block
	% For a $v$ that has a single approximately constant block, we have that 
	$\mathbb E | \langle \delta^{\text{single}},v\rangle|$ will be larger than $\mathbb E | \langle \delta^{\text{multiple}},v\rangle|$. For example, in the extreme case of constant binary block $|V_{(r,s)}|=11^\top$, we have 
	\[
	\mathbb E | \langle \delta^{\text{single}},v\rangle|=2\varepsilon h^2 >> \mathbb E | \langle \delta^{\text{multiple}},v\rangle| \asymp 2\varepsilon h.
	\]
	
	\subsection{Proof of Proposition~\ref{prop:squares_1}}
	\label{app:square}
	Let $x\in\F$, and $N(x)$ the number of elements of $\G$ that $x$ contains. Let initialize $x$ as a square of size $s\times s$, so that $N(x)=1$. We then add iteratively the remaining $k-s^2$ unitary squares to $x$ so to maximize $N(x)$.\\
	In order to get $N(x)=2$ it is necessary to increase $x$ to have size $s\times (s + 1)$. At this point, again to get $N(x)=3$ we need to add $s$ squares to one side of $x$. However, if we choose to glue them so to form a rectangle $s\times (s+2)$, then $N(x)=3$ and once more we need other $s$ squares to increase $N$, which means overall $s^2 + 3s$ to achieve $N(x)=4$. If instead we glue $s$ squares along the longer side, with only one additional unitary square we get $N(x)=4$ using $s^2 + 2s + 1 < s^2 + 3s$ unitary squares (as $s\geq 2$), with $x=(s+1)\times (s+1)$.\\
	Then, if the current shape of $x$ is $a\times b$ with $a\geq b$, the optimal option is adding $a$ unitary squares to have shape $a\times (b+1)$, increasing the count $N$ of $a-s+1$. This strategy can be repeated until the budget of $k$ unitary squares is reached.\\
	Finally, since we start from the shape $s\times s$, then at each stage $b-a\in\{0,1\}$, which means that the final $a$ will be $\floor{\sqrt{k}}$.
	A rectangle $a\times b$ in $\F$ contains $(a-s+1)(b-s+1)$ elements of $\G$.
	The remaining $k - ab$ squares can be glued along the longer side, contributing to $N(x)$ with $(k-ab-s+1)^+$.

	\section{Experimental Details}
	\label{app:exp_details}
	In this section, we list the main hyperparameters and various implementation details for the experiments done in the main experiments (Sec.~\ref{sec:exp}).
	
	\subsection{Experiments on ImageNet}
	For the untargeted Square Attack on the ImageNet models, we used $p=0.05$ and $p=0.1$ for the $l_\infty$- and $l_2$- versions respectively.
	For Bandits, we used their code with their suggested hyperparameters (specified in the configuration files) for both $l_\infty$ and $l_2$. For SignHunter, we used directly their code which does not have any hyperparameters (assuming that the finite difference probe $\delta$ is set to $\epsilon$).
	For SimBA-DCT, we used the default parameters of the original code apart from the following, which are the suggested ones for each model: for ResNet-50 and VGG-16-BN "freq\_dims" = $28$, "order" = "strided" and "stride" = $7$, for Inception v3 "freq\_dims" = $38$, "order" = "strided" and "stride" = $9$. Notice that SimBA tries to minimize the $l_2$-norm of the perturbations but it does not have a bound on the size of the changes. Then we consider it successful when the adversarial examples produced have norm smaller than the fixed threshold $\epsilon$. 
	The results for all other methods were taken directly from the corresponding papers.
	
	\customparagraph{Evaluation of Bandits.} %\cite{ilyas2019prior}.
	The code of Bandits \cite{ilyas2019prior} does not have image standardization at the stage where the set of correctly points is determined (see \url{https://github.com/MadryLab/blackbox-bandits/issues/3}). As a result, the attack is run only on the set of points correctly classified by the network \textit{without standardization}, although the network was trained on standardized images. We fix this bug, and report the results in Table~\ref{tab:main_linf} based on the fixed version of their code. We note that the largest difference of our evaluation compared to the $l_\infty$ results reported in Appendix~E of \cite{ilyas2019prior} is obtained for the VGG-16-BN network: we get $2.0\%$ failure rate while they reported $8.4\%$ in their paper. 
	%$2.0\%$ failure rate that we received versus $8.4\%$ from their paper. 
	Also, we note that the query count for Inception~v3 we obtain is also better than reported in \cite{ilyas2019prior}: 957 instead of 1117 with a slightly better failure rate. Our $l_2$ results also differ -- we obtain a significantly lower failure rate ($9.8\%$, $6.8\%$, $10.2\%$ instead of $15.5\%$, $9.7\%$, $17.2\%$ for the Inception~v3, ResNet-50, VGG-16-BN networks respectively) with improved average number of queries ($1486$, $939$, $511$ instead of $1858$, $993$, $594$).

	\subsection{Square Attack Can be More Accurate than White-box Attacks}
	For the $l_\infty$-Square Attack, we used $p=0.3$ for all models on MNIST and CIFAR-10. %and also on the post-averaging model \cite{lin2019bandlimiting} on ImageNet.
	For Bandits on MNIST and CIFAR-10 adversarially trained models we used "exploration" = $0.1$, "tile size" = $16$, "gradient iters" = $1$ following \cite{MooEtAl2019}.
	
	For the comparison of $l_2$-attacks on the $l_\infty$-adversarially trained model of \cite{madry2018towards} we used the Square Attack with the usual parameter $p=0.1$. For Bandits we used the parameters "exploration" = 0.01, "tile size" = $28$, "gradient iters" = $1$, after running a grid search over the three of them (all the other parameters are kept as set in the original code). For SimBA we used the "pixel attack" with parameters "order" = "rand", "freq\_dims" = 28, step size of $0.50$, after a grid search on all the parameters.

	\section{Ablation Study}
	\label{app:ablation_study}
	
\begin{figure*} \centering \setlength{\tabcolsep}{0.5pt}{\small\setlength{\newl}{0.29\columnwidth}
			\begin{tabular}{c c c c}
				& failure rate & avg. queries & median queries\\
				$l_\infty$ - $\epsilon=0.05$&
				\includegraphics[align=c, width=\newl]{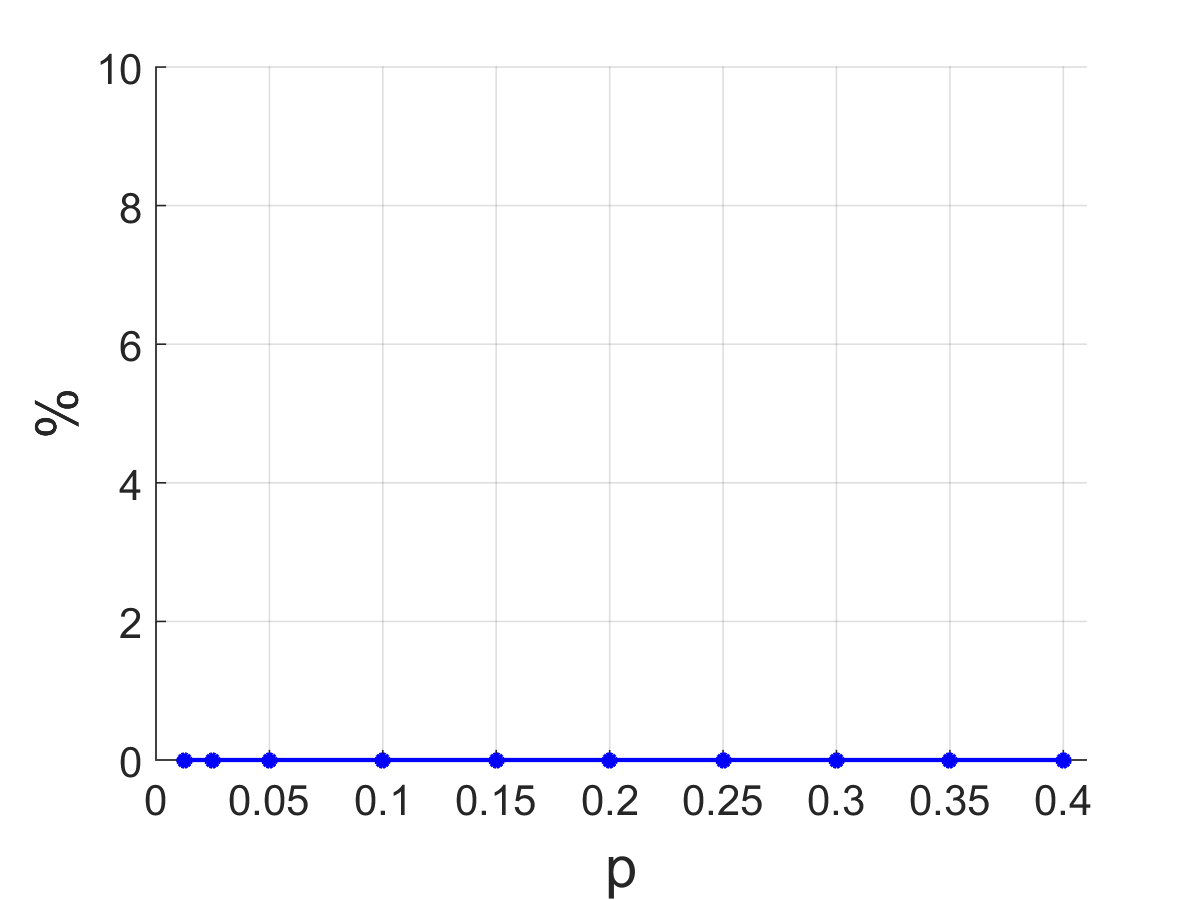}&
				\includegraphics[align=c, width=\newl]{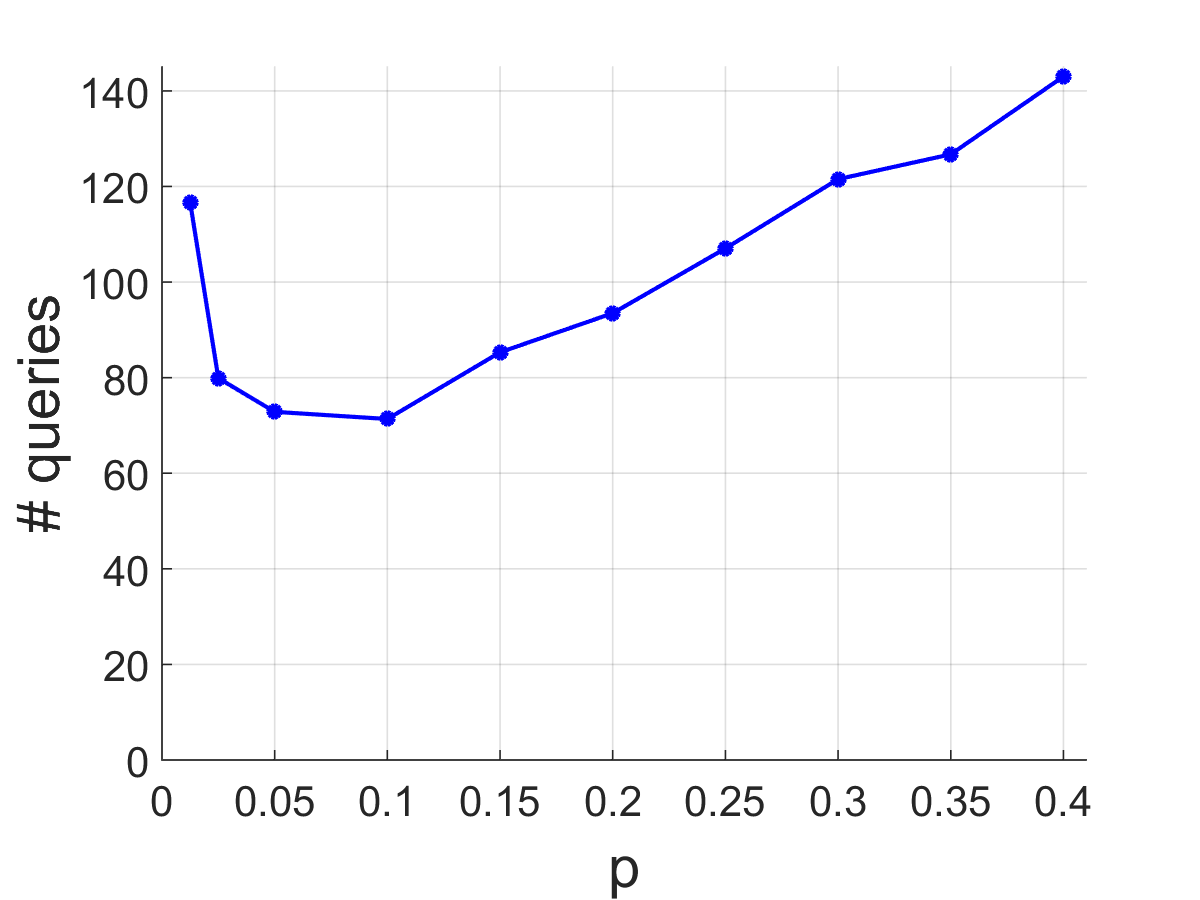}&
				\includegraphics[align=c, width=\newl]{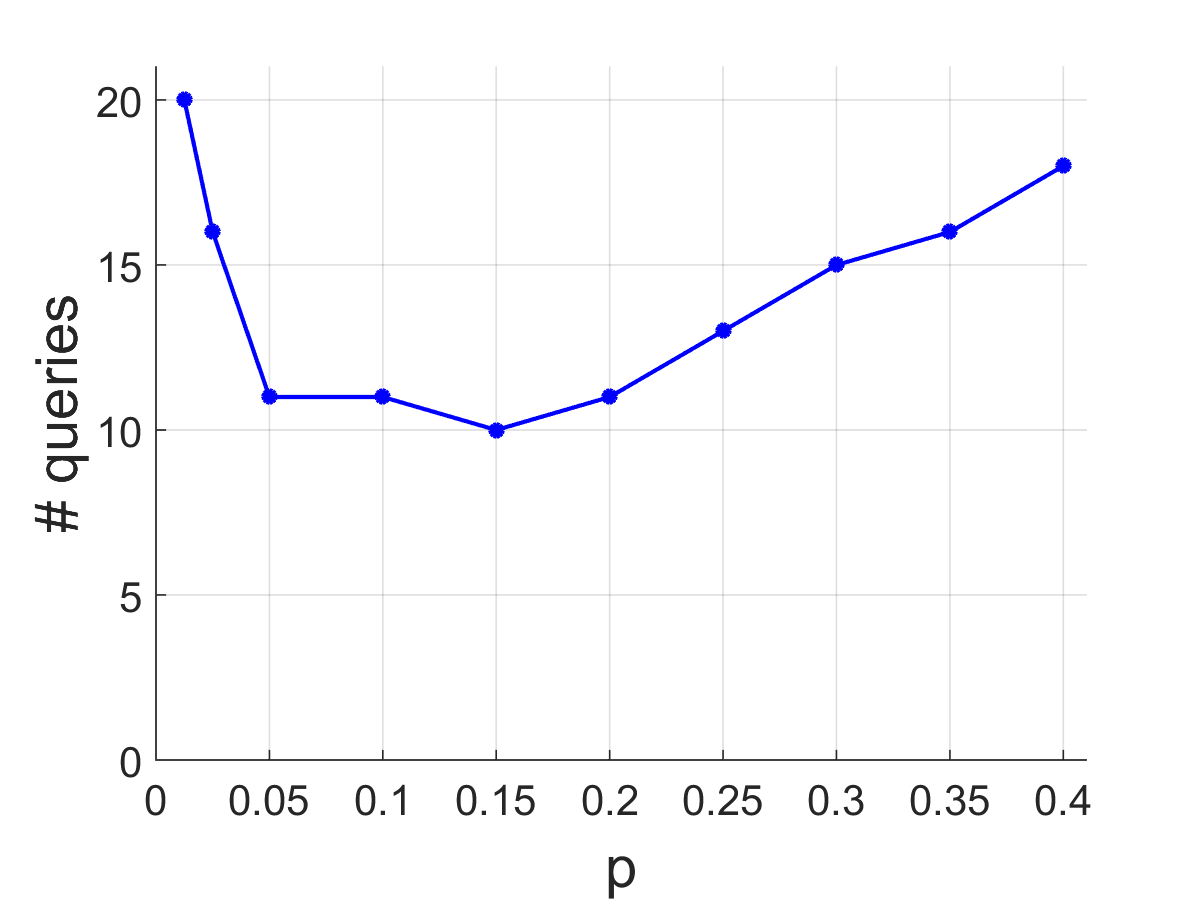}\\ 
				$l_2$ - $\epsilon=5.0$ & \includegraphics[align=c, width=\newl]{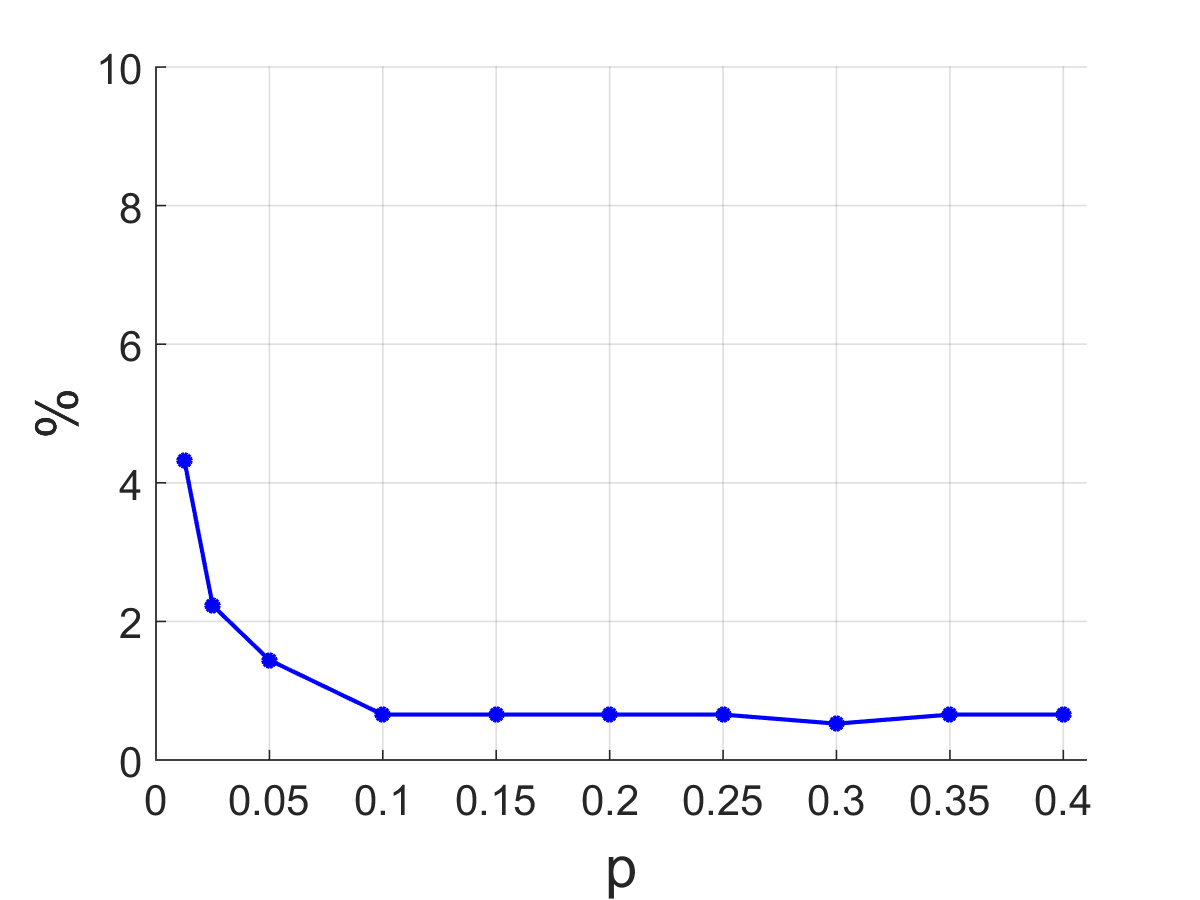}&
				\includegraphics[align=c, width=\newl]{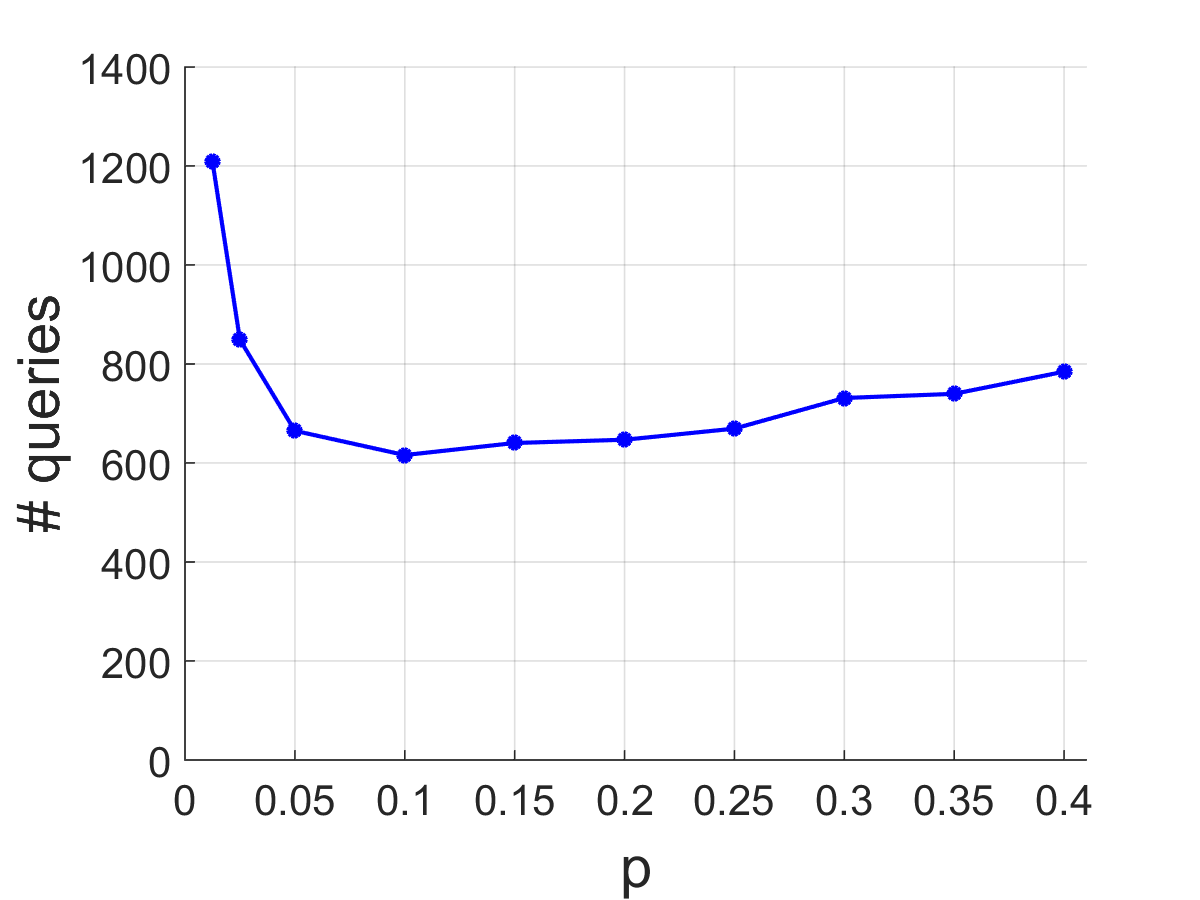}&
				\includegraphics[align=c, width=\newl]{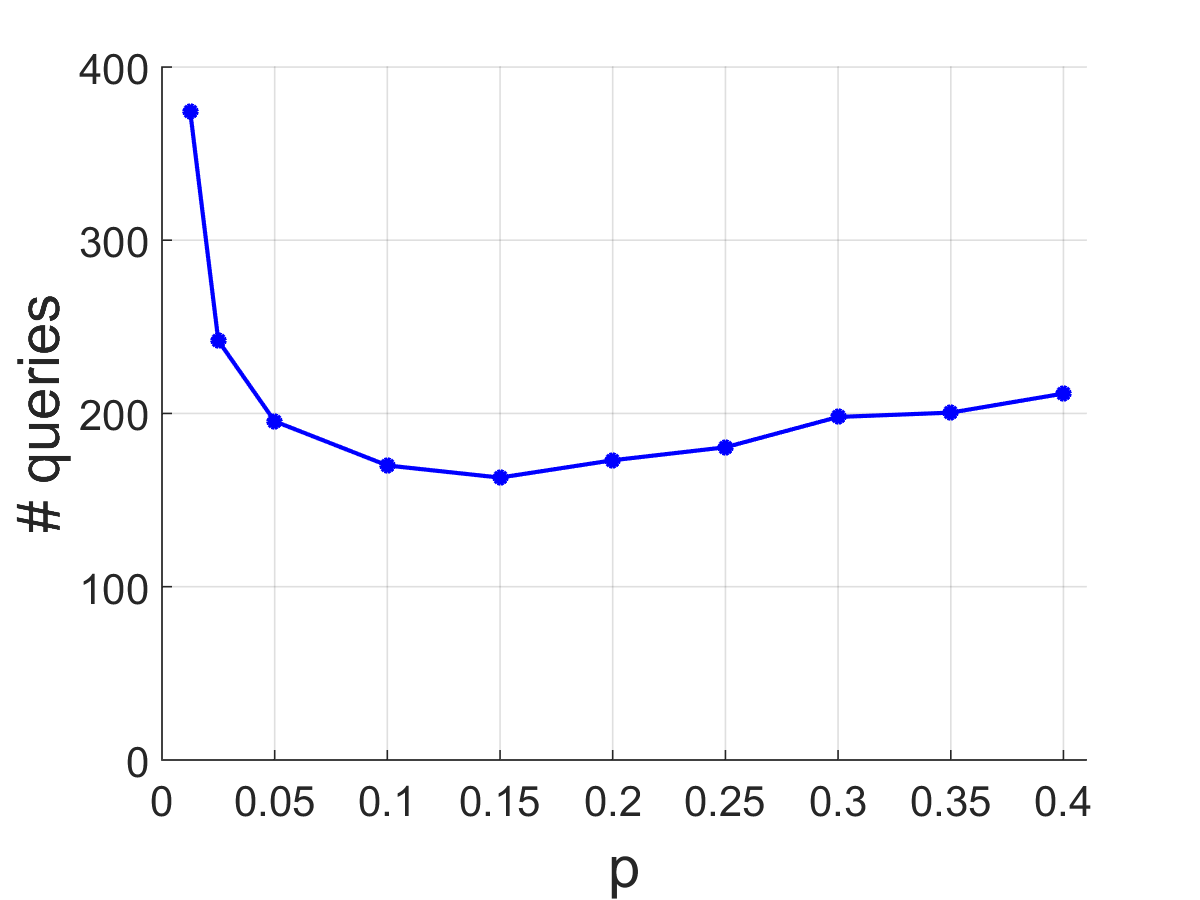}
		\end{tabular}}
		\caption{Sensitivity of the Square Attack to different choices of $p \in \{0.0125, 0.025, 0.05, 0.1, 0.15, 0.2, 0.25, 0.3, 0.35, 0.4\}$, i.e. the initial fraction of pixels changed by the attack, on ImageNet for a ResNet-50 model}\label{fig:ablation_study_p}
	\end{figure*}

	Here we discuss in more detail the ablation study which justifies the algorithmic choices made for our $l_\infty$- and $l_2$-attacks. Additionally, we discuss the robustness of the attack to the hyperparameter $p$, i.e. the initial fraction of pixels changed by the attack (see Fig.~\ref{fig:ablation_study_p}). We perform all these experiments on ImageNet with a standardly trained ResNet-50 model from the PyTorch repository.
	%where we plot the various statistics for $p \in \{0.0125, 0.025, 0.05, 0.1, 0.15, 0.2, 0.25, 0.3, 0.35, 0.4\}$.
	
	\subsection{$l_\infty$-Square Attack}
	
	\customparagraph{Sensitivity to the hyperparameter $p$.} First of all, we note that for \textit{all} values of $p$ we achieve $0.0\%$ failure rate. Moreover, we achieve state-of-the-art query efficiency with \textit{all} considered values of p (from $0.0125$ to $0.4$), i.e. we have the average number of queries below 140, and the median below 20 queries. Therefore, we conclude that the attack is robust to a wide range of $p$, which is an important property of a black-box attack -- since the target model is unknown, and one aims at minimizing the number of queries needed to fool the model, doing even an approximate grid search over $p$ is prohibitively expensive.

	\customparagraph{Algorithmic choices.}
	In addition to the results presented in Sec. \ref{sec:ablation}, we show in Table~\ref{tab:ablation_study} the results of a few more variants of the Square Attack. We recall that ``\# random signs'' indicates how many different signs we sample to build the updates, with $c$ being the number of color channels and $h$ the current size of the square-shaped updates.
	Specifically, we test the performance of using a single random sign for all the elements for the update, "square-$1$", which turns out to be comparable to ``square-$c\cdot h^2$'', i.e. every component of the update has sign independently sampled, but worse than keeping the sign constant within each color channel (``square-$c$'').
	
	In order to implement update shape ``rectangle'', on every iteration and for every image we sample $\alpha, \beta \sim \text{Exp}(1)$ and take a rectangle with sides $\alpha \cdot s$ and $\beta \cdot s$, so that in expectation its area is equal to $s^2$, i.e. to the area of the original square. This update scheme performs significantly better than changing a random subset of pixels (93 vs 339 queries on average), but worse than changing squares (73 queries on average) as discussed in Sec.~\ref{sec:why_squares}.
	
	Finally, we show the results with two more initialization schemes: horizontal stripes (instead of vertical), as well as initialization with randomly placed squares. While both solutions lead to the state-of-the-art query efficiency (83 and 90 queries on average) compared to the literature, they achieve worse results than the vertical stripes we choose for our Square Attack.
	
	%The last row represents the setting used in Sec. \ref{sec:exps_imagenet}.
	\definecolor{Gray}{rgb}{0.9, 0.9, 1}
	\begin{table}[t] \centering \caption{An ablation study for the performance of the $l_\infty$- and $l_2$-Square Attack under various algorithmic choices of the attack. The metrics are calculated on 1,000 ImageNet images for a ResNet-50 model. The last row represents our recommended setting. For all experiments we used the best performing $p$ ($0.05$ for $l_\infty$ and $0.1$ for $l_2$)}
		\label{tab:ablation_study}
		{\small
			
			\textbf{$l_\infty$ ablation study}
			
			\setlength{\tabcolsep}{4.0pt}
			\begin{tabular}{c c c | c c c}
				\hline
				Update & \# random & \multirow{2}{*}{Initialization} & Failure & Avg. & Median\\
				shape  & signs     &                                 & rate    & queries & queries\\
				\hline
				random & $c \cdot h^2$  & vert. stripes  &  0.0\% & 401 & 48\\
				random & $c \cdot h^2$  & uniform rand.  &  0.0\% & 393 & 132\\
				random & $c$            & vert. stripes  &  0.0\% & 339 & 53\\
				square & $c \cdot h^2$ & vert. stripes   &  0.0\% & 153 & 15\\
				square & $1$ & vert. stripes             &  0.0\% & 129 & 18\\
				rectangle & $c$ & vert. stripes          &  0.0\% & 93 & 16\\
				square & $c$ & uniform rand.             &  0.0\% & 91 & 26\\
				square & $c$ & rand. squares             &  0.0\% & 90 & 20\\
				square & $c$ & horiz. stripes            &  0.0\% & 83 & 18\\
				\rowcolor{Gray}
				square & $c$ & vert. stripes             &  0.0\% & 73 & 11\\
				\hline
			\end{tabular}
			
			\ \\ \ \\
			
			\textbf{$l_2$ ablation study}
			
			\begin{tabular}{c c | c c c}
				\hline
				\multirow{2}{*}{Update} & \multirow{2}{*}{Initialization} & Failure& Avg. & Median\\
				&  &rate & queries & queries\\
				\hline
				$\eta$\textsuperscript{rand}&
				$\eta$\textsuperscript{rand}-grid & 3.3\%&1050 & 324\\
				$\eta$\textsuperscript{single}&
				$\eta$\textsuperscript{single}-grid & 0.7\%&650&171\\
				$\eta$ & gaussian & 0.4\% & 696 & 189\\
				$\eta$ & uniform  & 0.8\% & 660 & 187\\
				$\eta$ & vert. stripes & 0.8\% & 655 &186\\
				\rowcolor{Gray}
				$\eta$ & $\eta$-grid & 0.7\% & 616 & 170\\
				\hline
			\end{tabular}
		}
		%\caption{An ablation study for the performance of the $l_\infty$ and $l_2$ Square Attack under various algorithmic choices of the attack. The statistics are calculate on 1,000 ImageNet images for a ResNet-50 model. The last row represents our recommended setting used in Sec.~\ref{sec:exps_imagenet}. For all experiments we used the best performing $p$ ($0.05$ for $l_\infty$ and $0.1$ for $l_2$)
		%}\label{tab:ablation_study} 
	\end{table}

	\subsection{$l_2$-Square Attack}
	
	\customparagraph{Sensitivity to the hyperparameter $p$.}
	We observe that the $l_2$-Square Attack is robust to different choices in the range between $0.05$ and $0.4$
	%while its performance degrades for $p\in\{0.0125, 0.025\}$
	showing approximately the same failure rate and query efficiency for all values of $p$ in this range, while its performance degrades slightly for very small initial squares $p\in\{0.0125, 0.025\}$.

	\customparagraph{Algorithmic choices.}
	We analyze in Table~\ref{tab:ablation_study} the sensitivity of the $l_2$-attack to different choices of the shape of the update and initialization.

	In particular, we test an update with only one "center" instead of two, namely $\eta^\textrm{single}=\eta^{h,h}$ (following the notation of Eq. \ref{eq:l2_pgp}) and one, $\eta^\textrm{rand}$, where the step 7 in Algorithm \ref{alg:l2_sampling_distribution} is $\rho~\gets~Uniform(\{-1, 1\}^{h\times h})$ instead of $\rho \gets  Uniform(\{-1, 1\})$, which means that each element of $\eta$ is multiplied randomly by either $-1$ or $1$ independently (instead of all elements multiplied by the same value). 
	We can see that using different random signs in the update and initialization ($\eta^\textrm{rand}$) significantly ($1.5\times$ factor) degrades the results for the $l_2$-attack, which is similar to the observation made for the $l_\infty$-attack.
	
	Alternatively to the grid described in Sec.~\ref{sec:sampl_l2}, we consider as starting perturbation
	i) a random point sampled according to $Uniform(\{-\epsilon/\sqrt{d}, \epsilon/\sqrt{d}\}^{w\times w\times c})$, that is on the corners of the largest $l_\infty$-ball contained in the $l_2$-ball of radius $\epsilon$ (uniform initialization),
	ii) a random position on the $l_2$-ball of radius $\epsilon$ (Gaussian initialization) or iii) vertical stripes similarly to what done for the $l_\infty$-Square Attack, but with magnitude $\epsilon/\sqrt{d}$ to fulfill the constraints on the $l_2$-norm of the perturbation.
	We note that different initialization schemes do not have a large influence on the results of our $l_2$-attack, unlike for the $l_\infty$-attack.

	\section{Stability of the Attack under Different Random Seeds}\label{app:stability}
	
	\begin{table}[t] 
		\caption{Mean and standard deviation of the main performance metrics of the Square Attack across 10 different runs with different random seeds} \label{tab:stability}
		\centering {\small
			\setlength{\tabcolsep}{5.0pt}
			\textbf{ImageNet, ResNet-50}
			
			\begin{tabular}{cc|ccc}
				\hline
				Norm & $\epsilon$ & Failure rate & Avg. queries & Median queries\\
				\hline
				$l_\infty$ & $0.05$ & $0.0\% \pm 0.0$\% & $72  \pm  2$ & $11  \pm 1$ \\
				$l_2$      & $5$    & $0.6\% \pm 0.1$\% & $638 \pm 12$ & $163 \pm 8$ \\
				\hline
			\end{tabular}
			
			\ \\ \ \\
			
			\textbf{MNIST, adversarially trained LeNet from \cite{madry2018towards}}
			
			\begin{tabular}{cc|ccc}
				\hline
				Norm & $\epsilon$ & Robust accuracy & Avg. queries & Median queries\\
				\hline
				$l_\infty$ & $0.3$ & $87.0\% \pm 0.1$\% & $299  \pm  47$ & $52  \pm 7$ \\
				$l_2$      & $2$   & $16.0\% \pm 1.4$\% & $1454 \pm 71$ & $742\pm 78$ \\
				\hline
			\end{tabular}
		} 
	\end{table}
	
	Here we study the stability of the Square Attack over the randomness in the algorithm, i.e. in the initialization, in the choice of the locations of square-shaped regions, and in the choice of the values in the updates $\delta$. We repeat 10 times experiments similar to the ones reported in Sec.~\ref{sec:exps_imagenet} and Sec.~\ref{sec:challenging_tasks} with different random seeds for our attack, and report all the metrics with standard deviations in Table~\ref{tab:stability}. On ImageNet, we evaluate the \textit{failure rate} (over initially correctly classified points) and query efficiency on 1,000 images using ResNet-50. On MNIST, we evaluate the \textit{robust accuracy} (i.e. the failure rate over all points) and query efficiency on 1,000 images using the $l_\infty$-adversarially trained LeNet from \cite{madry2018towards}. Note that unlike in Sec.~\ref{sec:challenging_tasks} in both cases we use a single restart for the attack on MNIST, and we compute the statistics on 1,000 points instead of 10,000, thus the final results will differ. 
	
	On the ImageNet model, all these metrics are very concentrated for both the $l_\infty$- and the $l_2$-norms. Moreover, we note that the standard deviations are much smaller than the gap between the Square Attack and the competing methods reported in Table~\ref{tab:main_linf}. Thus we conclude that the results of the attack are stable under different random seeds.
	
	On the adversarially trained MNIST model from \cite{madry2018towards}, the robust accuracy is very concentrated showing only $0.1\%$ and $1.4\%$ standard deviations for the $l_\infty$- and the $l_2$-norms respectively. Importantly, this is much less than the gaps to the nearest competitors reported in Tables~\ref{tab:adv_training_evaluation_linf} and \ref{tab:adv_training_evaluation_l2}. 
	We also show query efficiency for this model, although for models with non-trivial robustness it is more important to achieve lower robust accuracy, and query efficiency on successful adversarial examples is secondary.
	We note that the standard deviation of the mean and median number of queries is higher than for ImageNet, particularly for the $l_\infty$-ball of radius $\epsilon=0.3$ where the robust accuracy is much higher than for the $l_2$-ball of radius $\epsilon=2$.
	This is possibly due to the fact that attacking more robust models (within a certain threat model) is a more challenging task than, e.g., attacking standardly trained classifiers, as those used on ImageNet, which means that a favorable random initialization or perturbation updates can have more influence on the query efficiency.

	\section{Additional Experimental Results}
	\label{app:additional_exp_results}
	
	This section contains results on targeted attacks, and also additional results on untargeted attacks that complement the ImageNet results from Table~\ref{tab:main_linf}. 
	Moreover, we show that the Square Attack is useful for evaluating the robustness of newly proposed defenses (see Sec.~\ref{sec:post-averaging_app}).
	
	\subsection{Targeted Attacks}\label{sec:targeted_app}
	While in Sec.~\ref{sec:exp} we considered only untargeted attacks, here we report the results of the different attacks in the \textit{targeted} scenario.
	
	\customparagraph{Targeted Square Attack.} In order to adapt our scheme to targeted attacks, where one first choose a target class $t$ and then tries to get the model $f$ to classify a point $x$ as $t$, we need to modify the loss function $L$ which is minimized (see Eq.~\eqref{eq:opt_problem}). For the untargeted attacks we used the margin-based loss $L(f(x), y) = f_y(x) - \max_{k\neq y}f_k(x)$, with $y$ the correct class of $x$. This loss could be straightforwardly adapted to the targeted case as $L(f(x), t) = -f_t(x) +\max_{k\neq t}f_k(x)$. However, in practice we observed that this loss leads to suboptimal query efficiency. We hypothesize that the drawback of the margin-based loss in this setting is that the maximum over $k \neq t$ is realized by different $k$ at different iterations, and then the changes applied to the image tend to cancel each other. We observed this effect particularly on ImageNet which has a very high number of classes.
	
	Instead, we use here as objective function the cross-entropy loss on the target class, defined as \begin{equation} L(f(x), t) = - f_t(x) + \log\left(\sum_{i=1}^K e^{f_i(x)}\right).\label{eq:xent_targeted}
	\end{equation}
	Minimizing $L$ is then equivalent to maximizing the confidence of the classifier in the target class. Notice that in Eq.~\eqref{eq:xent_targeted} the scores of all the classes are involved, so that it increases the \textit{relative} weight of the target class respect to the others, making the targeted attacks more effective.
	\commentout{
		%\customparagraph{Experiments.}
		We present the results for targeted attacks on ImageNet for Inception-v3 model in Table~\ref{tab:targeted}. We calculate the statistics on 1,000 images (the target class is randomly picked for each image) with query limit of 100,000 for $l_\infty$ and on 200 points and with query limit 60,000 for $l_2$, as this one is more expensive computationally because of the lower success rate, with the same norm bounds $\epsilon$ used in the untargeted case. The targeted $l_\infty$-Square Attack achieves 100\% success rate and requires $1.5$ times fewer queries on average than the nearest competitor \cite{MeuEtAl2019}.}
	\begin{table}[h]
		\caption{Results of \textbf{targeted} attacks on ImageNet for Inception-v3 model using 100k query limit for $l_\infty$, 60k for $l_2$. The results for the competing methods for $l_\infty$ are taken from 
			\cite{MeuEtAl2019}, except SignHunter~\cite{AlDujaili2019ThereAN} which we evaluated using their code }\label{tab:targeted}
		\centering\small
		\setlength{\tabcolsep}{4pt}
		\begin{tabular}{cc ccc}
			\hline
			\textbf{Norm} & \textbf{Attack} & \textbf{Failure rate} & \textbf{Avg. queries} & \textbf{Median queries}\\
			\hline
			\multirow{6}{*}{$l_\infty$}& Bandits \cite{ilyas2019prior} & 7.5\%& 25341 & 18053 \\
			&SignHunter \cite{AlDujaili2019ThereAN} & 1.1\% & 8814 & 5481 \\
			&Parsimonious \cite{MooEtAl2019} & \textbf{0.0\%} & 7184 & 5116 \\
			&DFO$_c$ -- Diag. CMA \cite{MeuEtAl2019} & 6.0\% & 6768 & 3797  \\
			&DFO$_c$ -- CMA \cite{MeuEtAl2019} & \textbf{0.0\%} & 6662 & 4692 \\
			&\textbf{Square Attack} & \textbf{0.0\%} & \textbf{4584} & \textbf{2859} \\ 
			\hline
			\multirow{3}{*}{$l_2$}& Bandits \cite{ilyas2019prior}&\textbf{24.5\%}& 20489 & 17122 \\
			% &SignHunter \cite{AlDujaili2019ThereAN} & 36.8\% & 31457 & 29438
			& SimBA-DCT \cite{guo2019simple} & 25.5\% & 30576 & 30180 \\
			&\textbf{Square Attack} & 33.5\% & \textbf{19794} & \textbf{15946} 
			\\ \hline
	\end{tabular}\end{table}
	
	\customparagraph{Experiments.} We present the results for targeted attacks on ImageNet for Inception-v3 model in Table~\ref{tab:targeted}. We calculate the statistics on 1,000 images (the target class is randomly picked for each image) with query limit of 100,000 for $l_\infty$ and on 200 points and with query limit 60,000 for $l_2$, as this one is more expensive computationally because of the lower success rate, with the same norm bounds $\epsilon$ used in the untargeted case.
	We use the Square Attack with $p=0.01$ for $l_\infty$ and $p=0.02$ for $l_2$.
	The results for the competing methods for $l_\infty$ are taken from \cite{MeuEtAl2019}, except SignHunter~\cite{AlDujaili2019ThereAN} which was not evaluated in the targeted setting before, thus we performed the evaluation using their code on 100 test points using the cross-entropy as the loss function. For $l_2$, we use Bandits \cite{ilyas2019prior} with the standard parameters used in the untargeted scenario, while we ran a grid search over the step size of SimBA (we set it to $0.03$) and keep the other hyperparameters as suggested for the Inception-v3 model.
	
	The targeted $l_\infty$-Square Attack achieves 100\% success rate and requires $1.5$ times fewer queries on average than the nearest competitor \cite{MeuEtAl2019}, showing that even in the \textit{targeted} scenario our simple scheme outperforms the state-of-the-art methods.
	On the other hand, our $l_2$-attack suffers from worse higher failure rate than the competitors, but achieves lower average and median number of queries (although on a different number of successful points).

	\subsection{Success Rate on ImageNet for Different Number of Queries}
	\label{app:additional_exp_results_imagenet}
	\begin{figure*}[t] \centering 
		\setlength{\newl}{0.25\columnwidth}
		\setlength{\tabcolsep}{1.5pt}{\small
			\begin{tabular}{c c c c}
				& \textbf{Inception v3} & \textbf{ResNet-50} & \textbf{VGG-16-BN}\\
				\textbf{$l_\infty$-attacks} & \includegraphics[align=c, width=\newl]{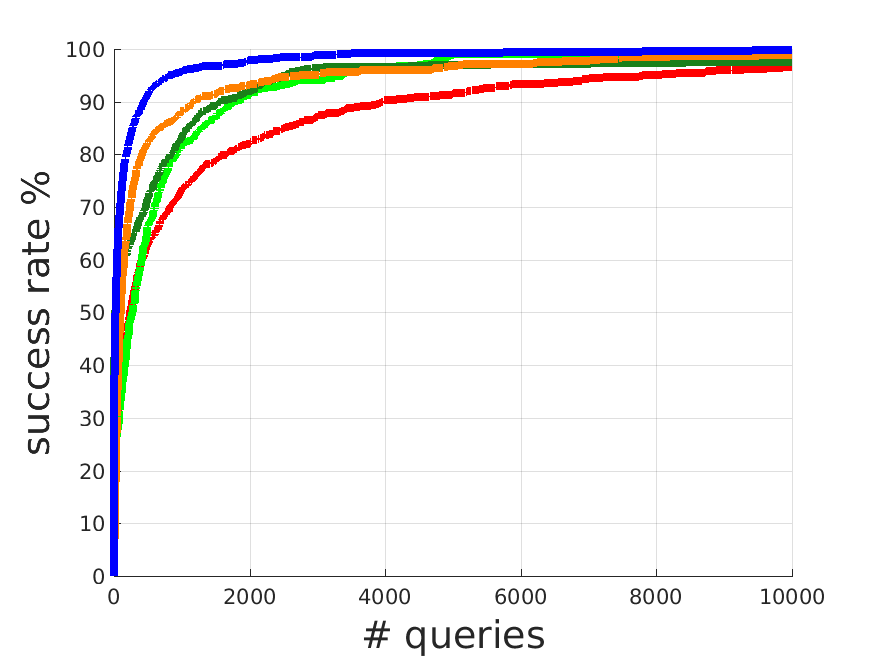}&
				\includegraphics[align=c, width=\newl]{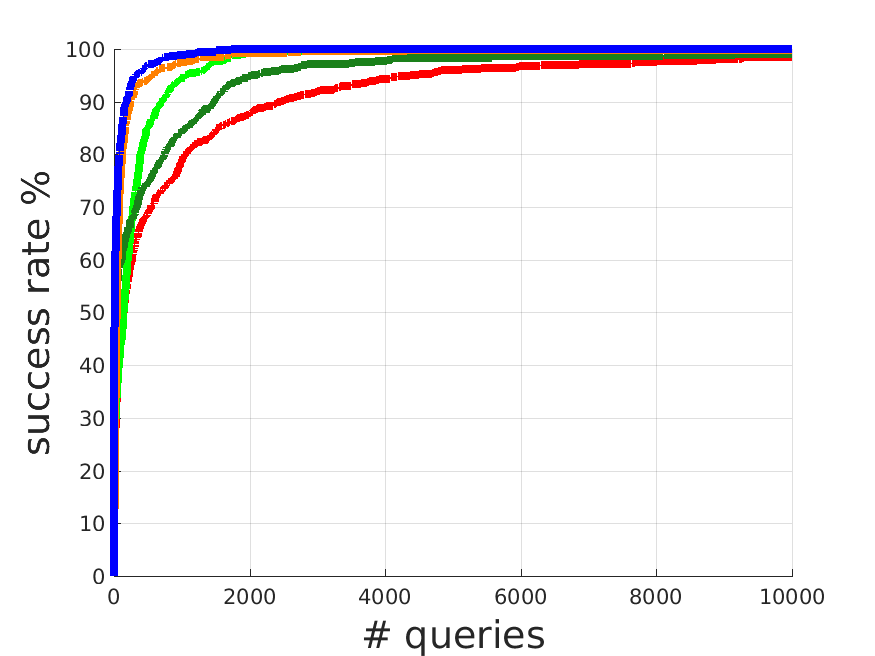}&
				\includegraphics[align=c, width=\newl]{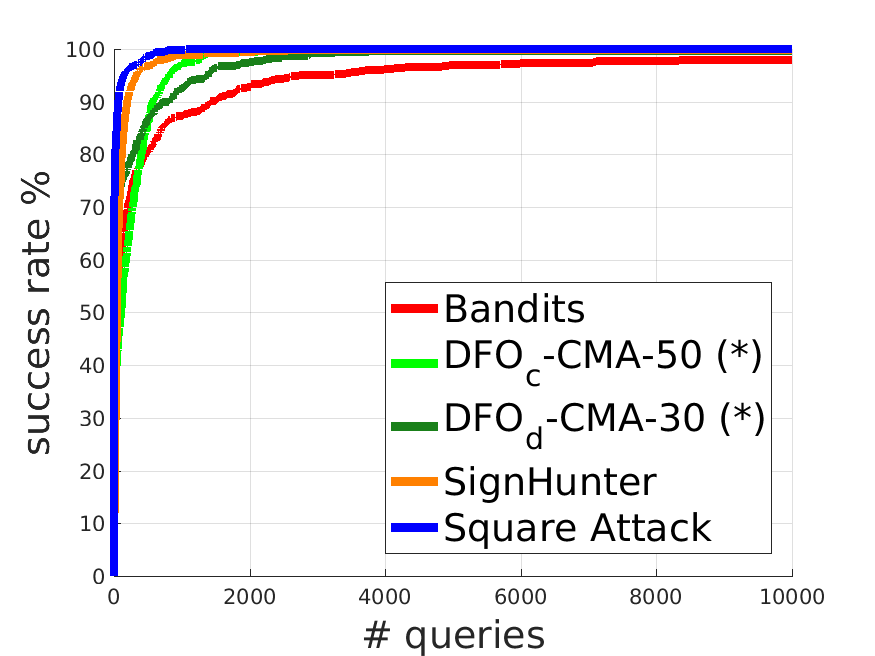}\\ 
				\makecell{\textbf{$l_\infty$-attacks} \\ \textbf{low query regime}}&
				\includegraphics[align=c, width=\newl]{pl_success_rate_linf_qrmax_200_inceptionv3_new}&
				\includegraphics[align=c, width=\newl]{pl_success_rate_linf_qrmax_200_resnet_new}&
				\includegraphics[align=c, width=\newl]{pl_success_rate_linf_qrmax_200_vgg16_new}\\
				\textbf{$l_2$-attacks}&
				\includegraphics[align=c, width=\newl]{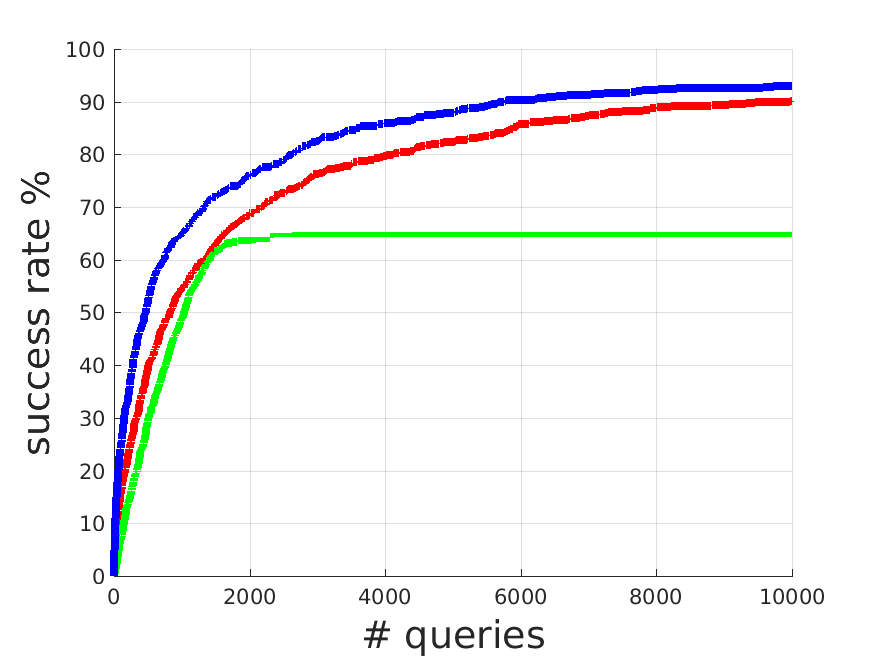}&
				\includegraphics[align=c, width=\newl]{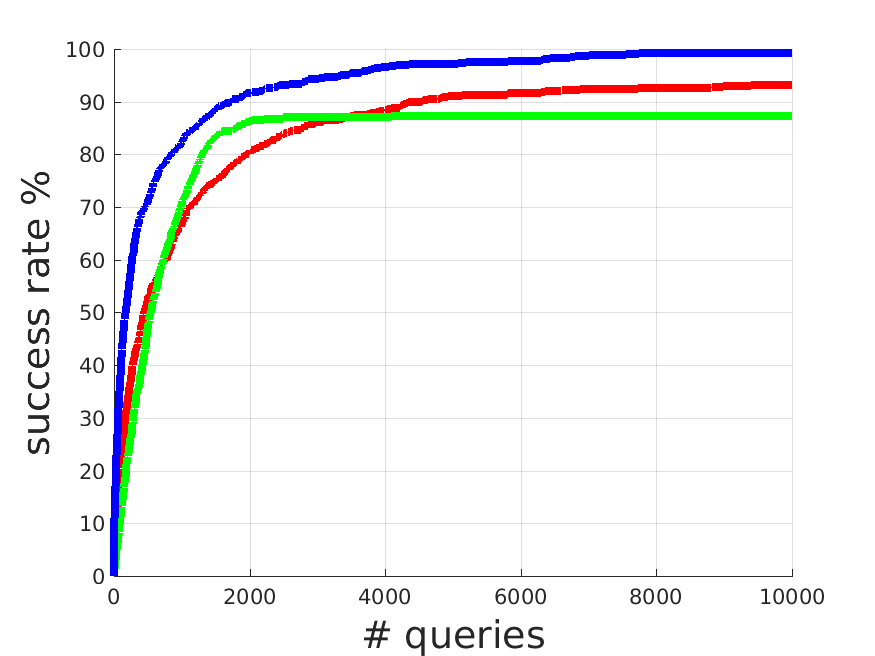}&
				\includegraphics[align=c, width=\newl]{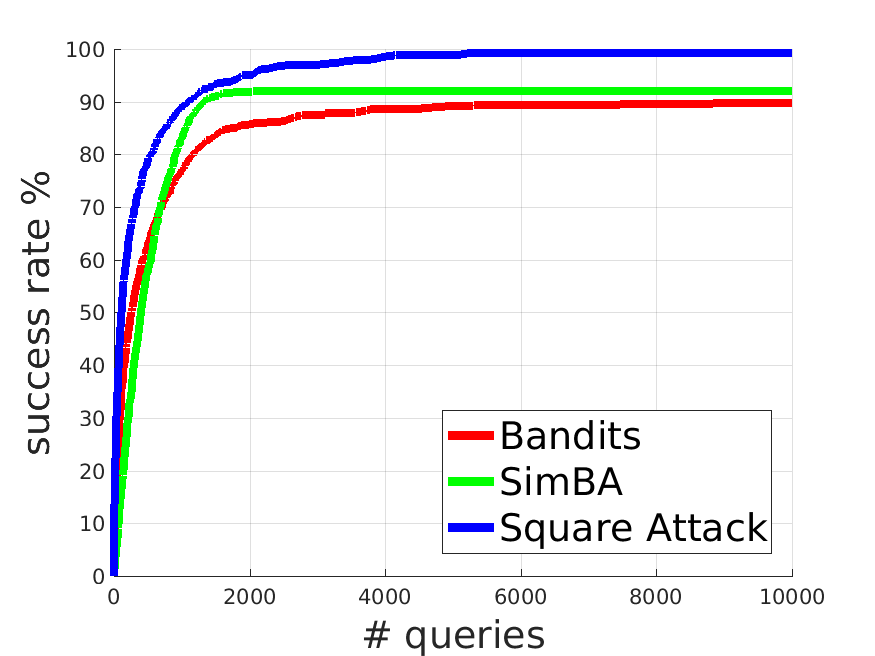}\\ 
				\makecell{\textbf{$l_2$-attacks} \\ \textbf{low query regime}}&
				\includegraphics[align=c, width=\newl]{pl_success_rate_qrmax_200_inceptionv3_new}&
				\includegraphics[align=c, width=\newl]{pl_success_rate_qrmax_200_resnet_new}&
				\includegraphics[align=c, width=\newl]{pl_success_rate_qrmax_200_vgg16_new.png}
			\end{tabular}
		}
		\caption{Success rate vs number of queries for different attacks on ImageNet on three standardly trained models. The low query regime corresponds to up to 200 queries, while the standard regime corresponds to 10,000 queries. $^*$ denotes the results obtained via personal communication with the authors and evaluated on 500 and 10,000 randomly sampled points for DFO \cite{MeuEtAl2019} and BayesAttack \cite{shukla2019blackbox} methods, respectively
		}\label{fig:success_rate_queries}
	\end{figure*}

	In this section, we provide a more detailed comparison to the competitors from Table~\ref{tab:main_linf} under different query budgets
	and more comments about the low query regime experiment in Fig.~\ref{fig:low_query}. 
	We show in Fig.~\ref{fig:success_rate_queries} the behaviour of the success rate for each attack depending on the number queries.
	The success rates of the attacks from \cite{MeuEtAl2019} (DFO$_c$--CMA--50 and DFO$_d$--Diag. CMA--30) and \cite{shukla2019blackbox} (BayesAttack) for different number of queries were obtained via personal communication directly from the authors, and were calculated on 500 and 10,000 randomly sampled points, respectively. For the other attacks, as mentioned above, the success rate is calculated on 1,000 randomly sampled points.
	
	%\textbf{$l_\infty$ results:}
	\customparagraph{$l_\infty$-results.}
	First, we observe that the Square Attack outperforms all other methods in the standard regime with 10,000 queries. 
	The gap in the success rate gets larger in the range of $100$-$1000$ queries for the more challenging Inception-v3 model, where we observe over $10\%$ improvement in the success rate over all other methods including SignHunter.
	Our method also outperforms the BayesAttack in the low query regime, i.e. less than $200$ queries, by approximately $20\%$ on every model. We note that DFO$_d$--Diag. CMA--30 method is also quite effective in the low query regime showing results close to BayesAttack. However, it is also outperformed by our Square Attack.
	
	%\textbf{$l_2$ results:}
	\customparagraph{$l_2$-results.}
	First, since the $l_2$-version of SignHunter~\cite{AlDujaili2019ThereAN} is not competitive to Bandits on ImageNet (see Fig.~2 in \cite{AlDujaili2019ThereAN}), we do not compare to them here.
	The $l_2$-Square Attack outperforms both Bandits and SimBA, and the gap is particularly large in the low query regime. 
	We note that the success rate of SimBA plateaus after some iteration. This happens due to the fact that their algorithm only adds orthogonal updates to the perturbation, and does not have any way to correct the greedy decisions made earlier. Thus, there is no progress anymore after the norm of the perturbation reaches the $\epsilon=5$
	(note that we used for SimBA the same parameters of the comparison between SimBA and Bandits in \cite{guo2019simple}).
	Contrary to this, both Bandits and our attack constantly keep improving the success rate, although with a different speed.

	\subsection{Performance on Architectures with Dilated Convolutions} \label{sec:dilated_convolutions}
	In Sec.~\ref{sec:why_squares}, we provided justifications for square-shaped updates for \textit{convolutional} networks. Thus, a reasonable question is whether the Square Attack still works equally well on less standard convolutional networks such as, for example, networks with dilated convolutions.
	For this purpose, we evaluate three different architectures introduced in \cite{yu2017dilated} that involve \textit{dilated} convolutions: DRN-A-50, DRN-C-42 and DRN-D-38. We use $\epsilon_\infty=0.05$ as in the main ImageNet experiments from Table~\ref{tab:main_linf}. We present the results in Table~\ref{tab:dilated_convolutions} and observe that for all the three model the Square Attack achieves 100\% success rate and both average and median number of queries stay comparable to that of VGG or ResNet-50 from Table~\ref{tab:main_linf}. Thus, this experiment suggests that our attack can be applied not only to standard convolutional networks, but also to more recent neural network architectures.
	% DRN-A-50: 2504: acc=0.00% acc_corr=0.00% avg#q_ae=85.96 med#q=12.0, avg_margin=-1.06 (p=0.05 s=50->6.00, n_ex=763, eps=0.050, 735.87s)
	% DRN-C-42: 3185: acc=0.00% acc_corr=0.00% avg#q_ae=56.79 med#q=7.0, avg_margin=-1.51 (p=0.05 s=50->6.00, n_ex=784, eps=0.050, 763.08s)
	% DRN-D-38: 2147: acc=0.00% acc_corr=0.00% avg#q_ae=48.35 med#q=5.5, avg_margin=-1.32 (p=0.05 s=50->6.00, n_ex=776, eps=0.050, 597.67s)
	\begin{table}[t!]
		\caption{Results of untargeted $l_\infty$-perturbations produced by the Square Attack on architectures with dilated convolutions}
		\label{tab:dilated_convolutions}
		\centering
		{\setlength{\tabcolsep}{4pt}
			\small
			\begin{tabular}{cc|ccc}
				\hline
				$\epsilon_\infty$ & \textbf{Model} & \textbf{Failure rate} & \textbf{Avg. queries} & \textbf{Median queries} \\ 
				\hline
				\multirow{3}{*}{$0.05$}  & DRN-A-50 & 0.0\% & 86 & 12 \\
				& DRN-C-42 & 0.0\% & 57 & 7 \\
				& DRN-D-38 & 0.0\% & 48 & 6 \\
				\hline
		\end{tabular}}
	\end{table}

	\subsection{Imperceptible Adversarial Examples with the Square Attack}\label{sec:imperceptible_adv_ex}
\begin{figure}[h!] 
		\centering 
		\small 
		\setlength{\newl}{0.30\columnwidth}
		\begin{tabular}{ccc}
			\textbf{Original image} & \textbf{Perturbation} & \textbf{Adversarial image} \\ 
			\includegraphics[width=\newl]{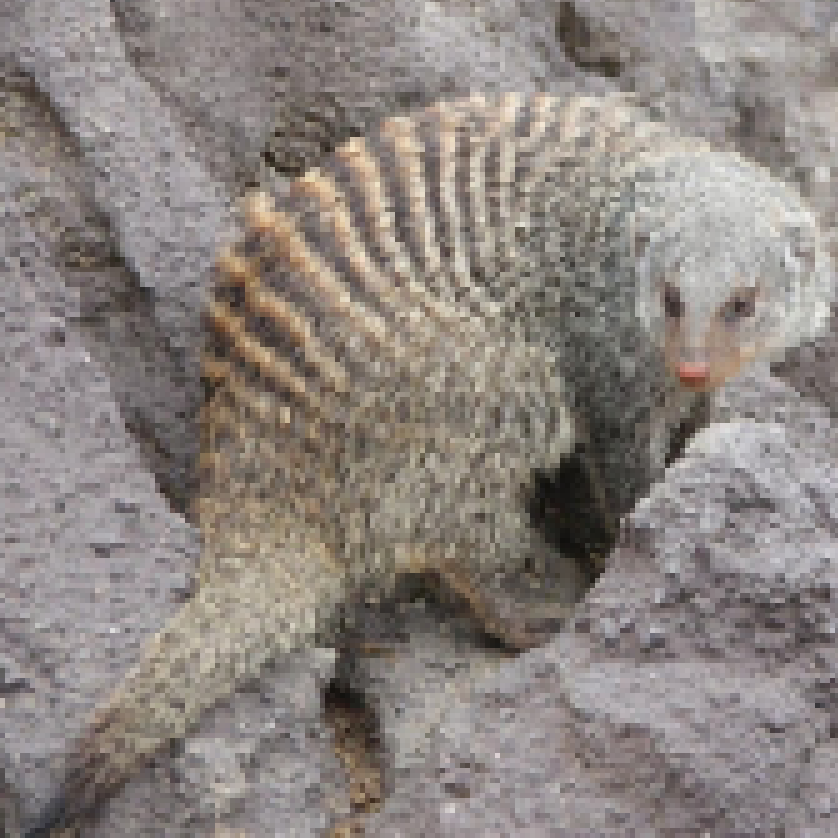} & 
			\includegraphics[width=\newl]{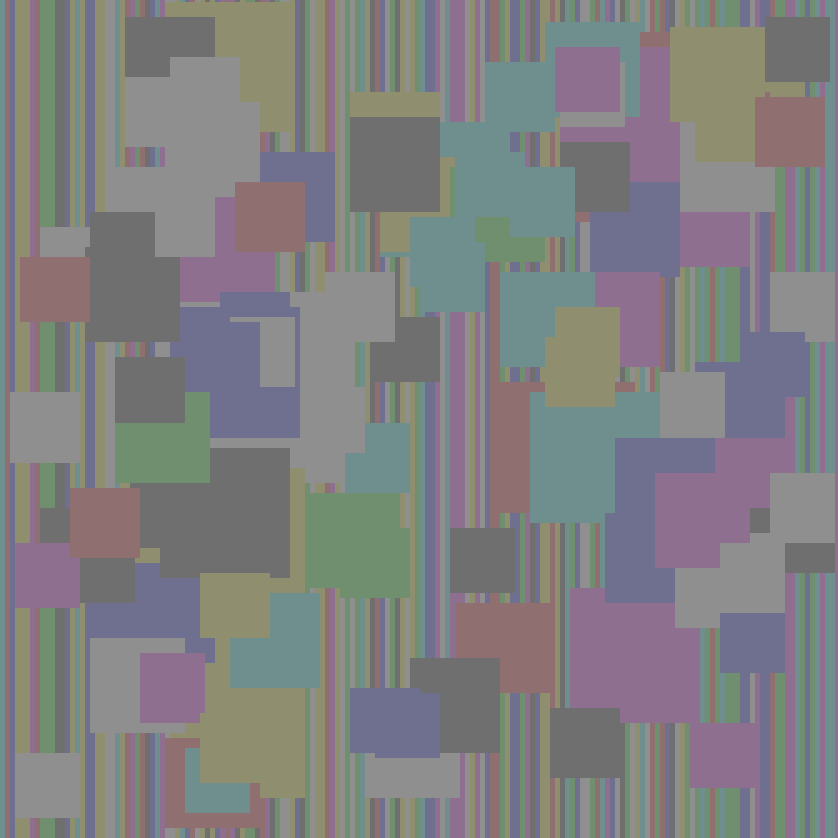} &
			\includegraphics[width=\newl]{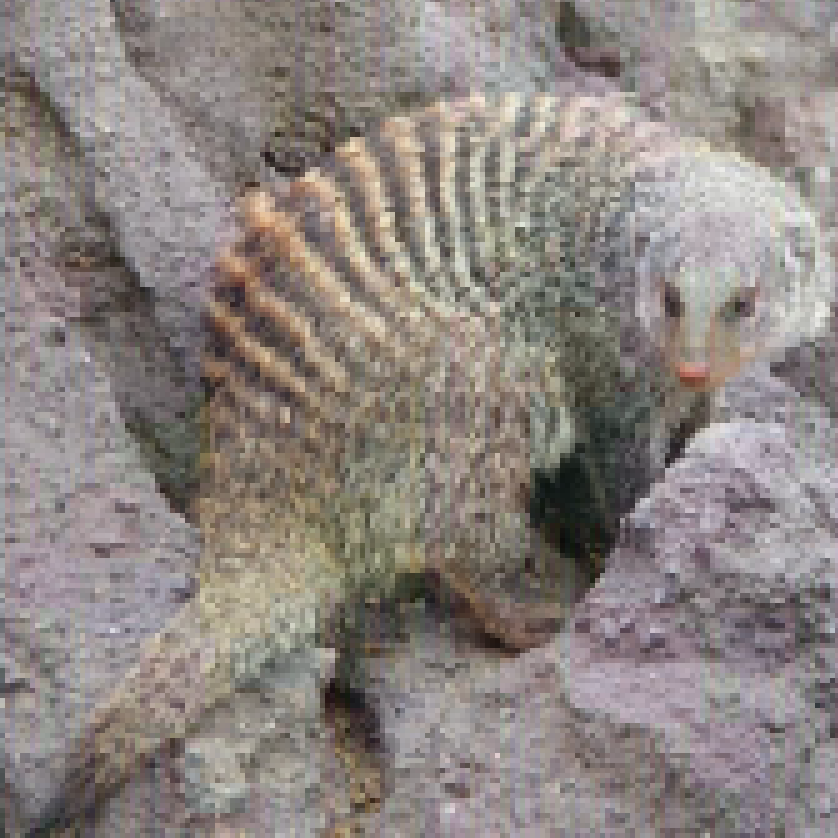} \\
			\includegraphics[width=\newl]{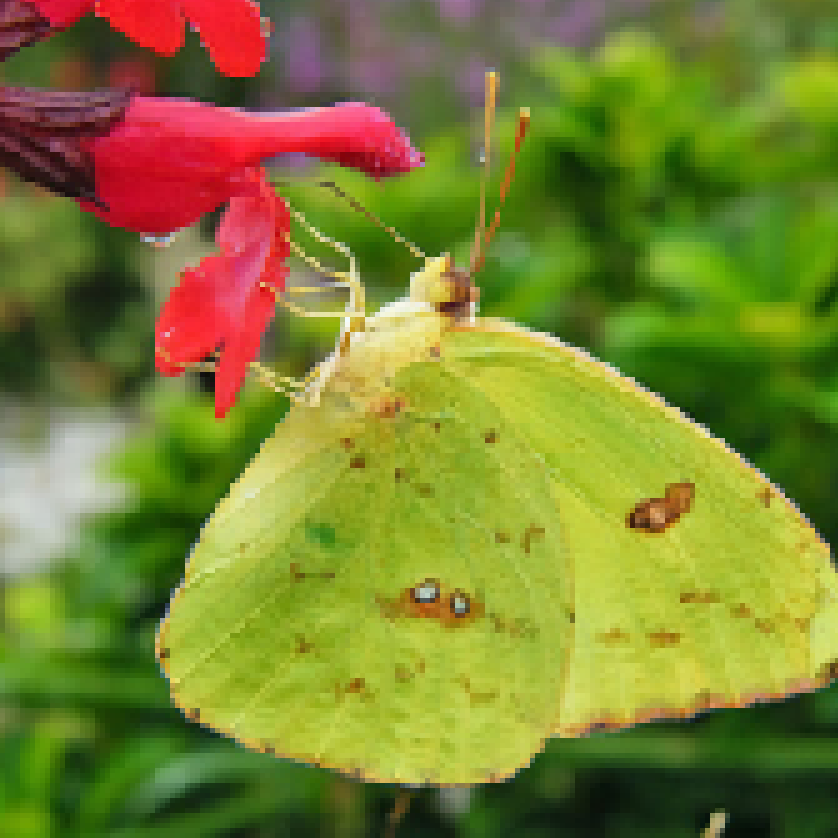} & 
			\includegraphics[width=\newl]{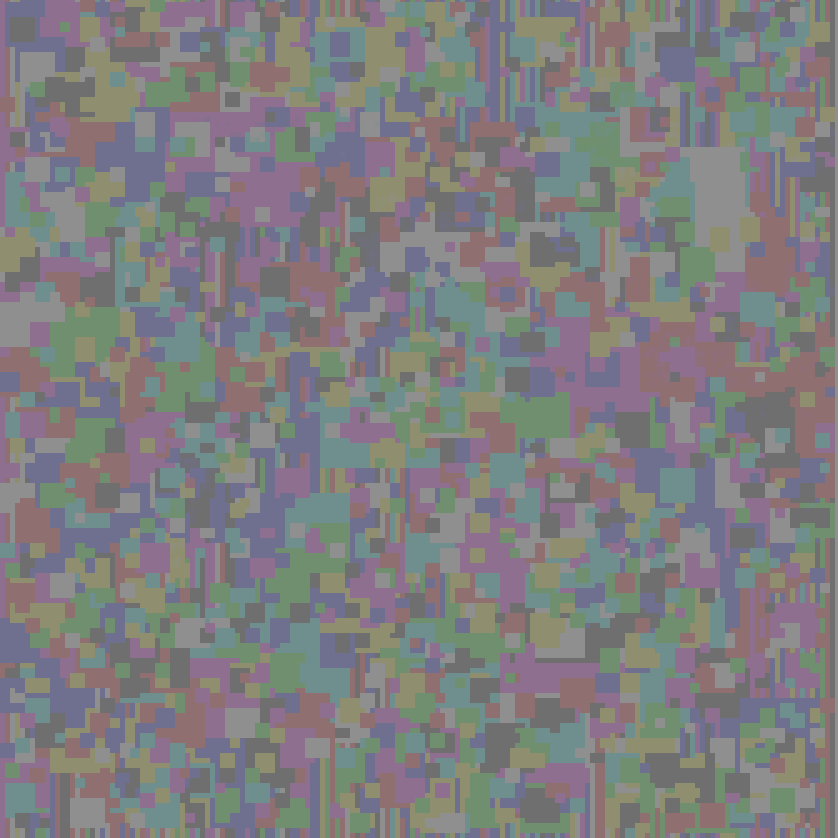} &
			\includegraphics[width=\newl]{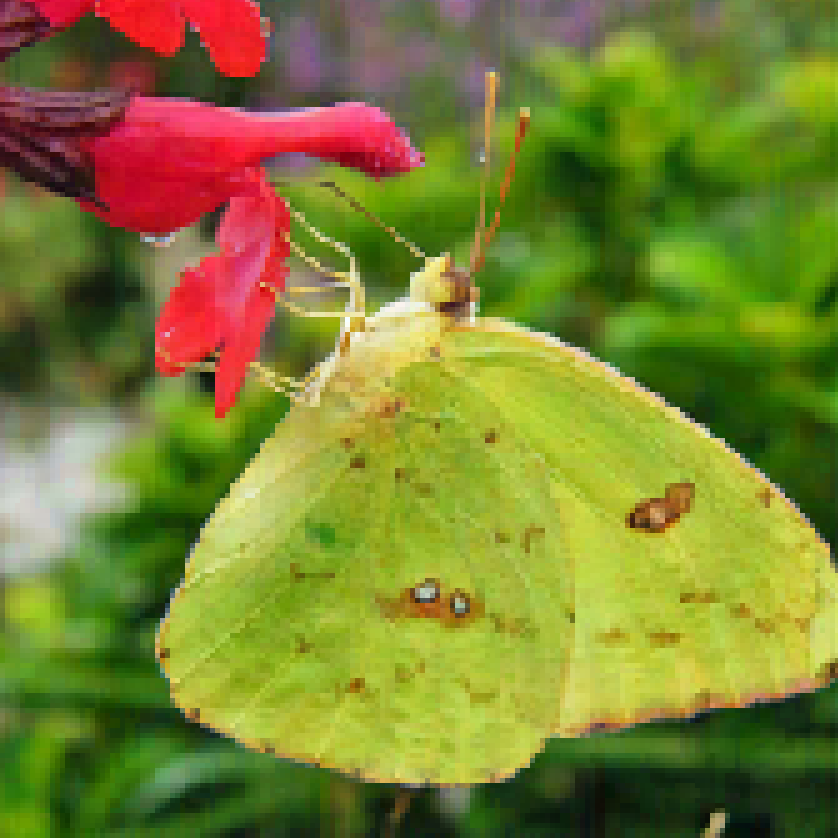} \\
			\includegraphics[width=\newl]{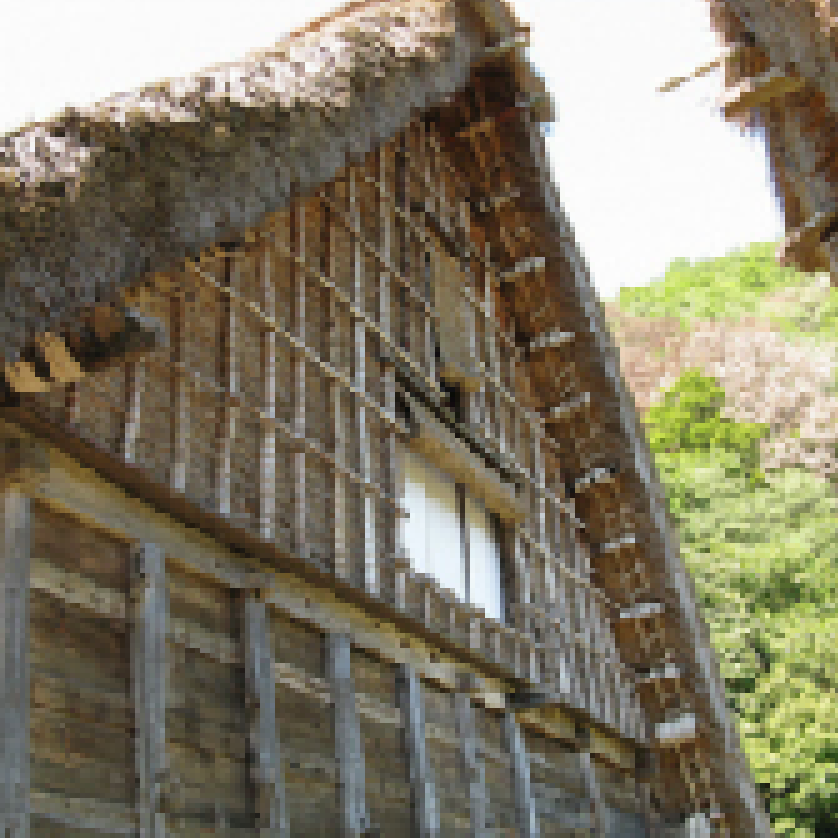} & 
			\includegraphics[width=\newl]{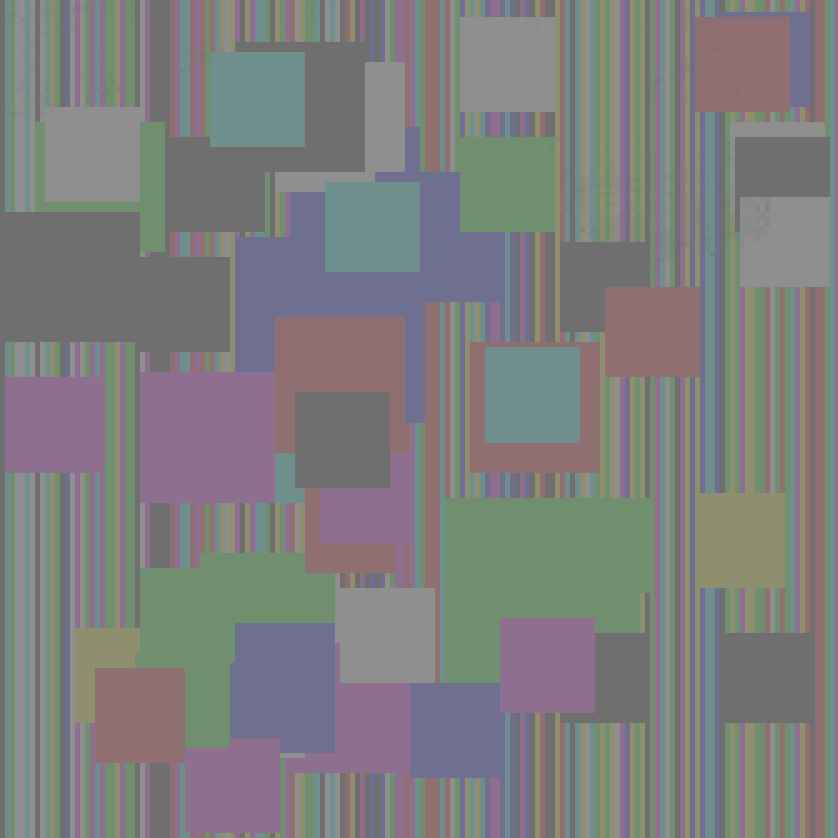} &
			\includegraphics[width=\newl]{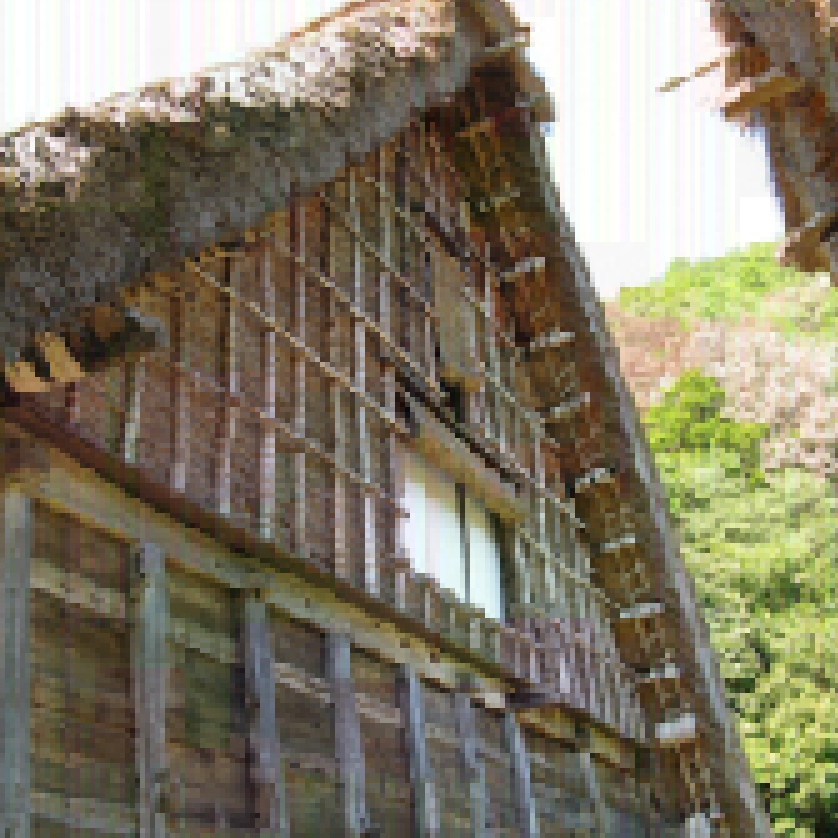} 
		\end{tabular}
		\caption{Visualization of the \textit{imperceptible} adversarial examples found by the $l_\infty$ Square Attack on ImageNet using ResNet-50 for $\epsilon_\infty=4/255$. All the original images were correctly classified while the adversarial images are misclassified by the model. The perturbations are amplified for the visualization purpose} 
		\label{fig:imperceptible_adv_ex}
	\end{figure}

	Adversarial examples in general need not be imperceptible, for example adversarial patches \cite{BroEtAl2017,karmon2018lavan} are clearly visible, and yet can be used to attack machine learning systems deployed in-the-wild. However, if imperceptibility is the goal, it can be easily ensured by adjusting the size of the allowed perturbations. In the main ImageNet experiments in Table~\ref{tab:main_linf} we used $\epsilon_\infty = 0.05 = 12.75/255$ since this is standard in the literature \cite{AlDujaili2019ThereAN,ilyas2019prior,MeuEtAl2019,MooEtAl2019}, and for which all the considered attacks produce visible perturbations. Below we additionally provide results on the more than $3\times$ smaller threshold $\epsilon_\infty = 4/255$ which leads to imperceptible perturbations (see Fig.~\ref{fig:imperceptible_adv_ex}). Our attack still achieves almost perfect success rate requiring only a limited number of queries as shown in Table~\ref{tab:imperceptible_adv_ex}. Thus, one can also generate imperceptibile adversarial examples with the Square Attack simply by adjusting the perturbation size.
	\begin{table}[b!]
		\caption{Results of untargeted imperceptible $l_\infty$-perturbations produced by the Square Attack on standard architectures}
		\label{tab:imperceptible_adv_ex}
		\centering
		{\setlength{\tabcolsep}{4pt}
			\small
			\begin{tabular}{cc|ccc}
				\hline
				$\epsilon_\infty$ & \textbf{Model} & \textbf{Failure rate} & \textbf{Avg. queries} & \textbf{Median queries} \\ 
				\hline
				\multirow{3}{*}{$4/255$} & VGG          & 0.5\% & 424  & 115 \\
				& ResNet-50    & 0.3\% & 652  & 213 \\
				& Inception v3 & 5.4\% & 1013 & 391 \\
				\hline
		\end{tabular}}
	\end{table}
	
	\subsection{Analysis of Adversarial Examples that Require More Queries} \label{sec:high_query_adv}
\begin{figure}[t!] 
		\centering 
		\small 
		\setlength{\newl}{0.2\columnwidth}
		\setlength{\newlloss}{0.36\columnwidth}
		\begin{tabular}{cccc}
			\textbf{Margin loss} & \textbf{10 iterations} & \textbf{100 iterations} & \textbf{500 iterations} \\ 
			\hline
			\ \\
			& Class: limpkin & Class: limpkin & Class: {\color{red} dowitcher} \\ 
			\multirow[c]{2}{*}[14.5mm]{
				\includegraphics[width=\newlloss]{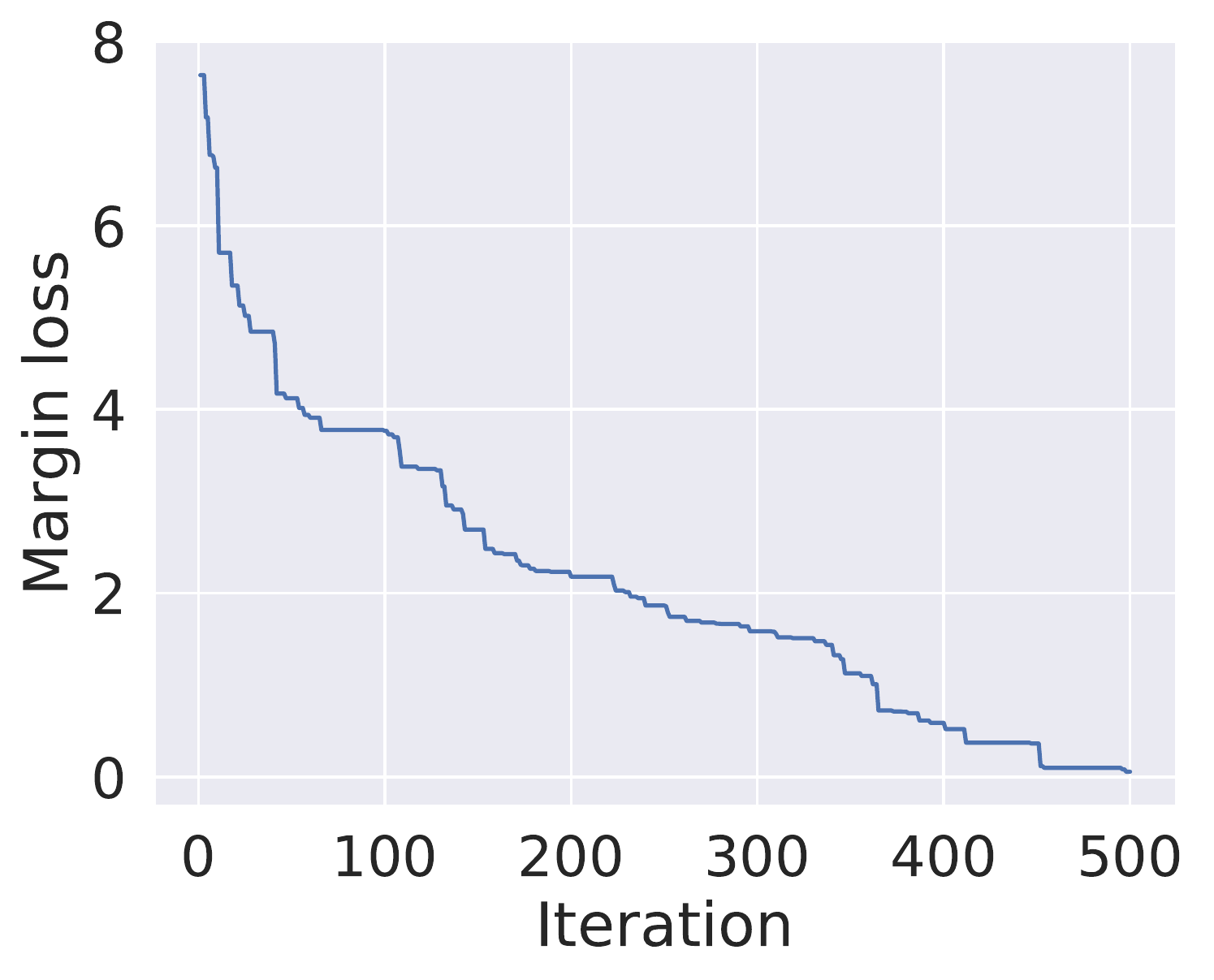}} & 
			\includegraphics[width=\newl]{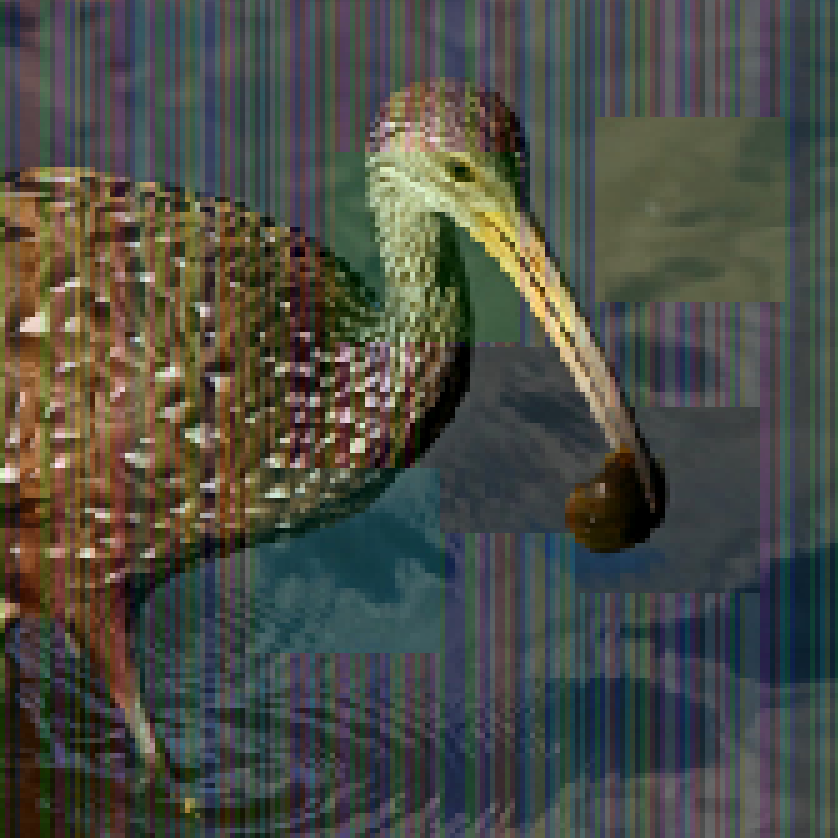} &
			\includegraphics[width=\newl]{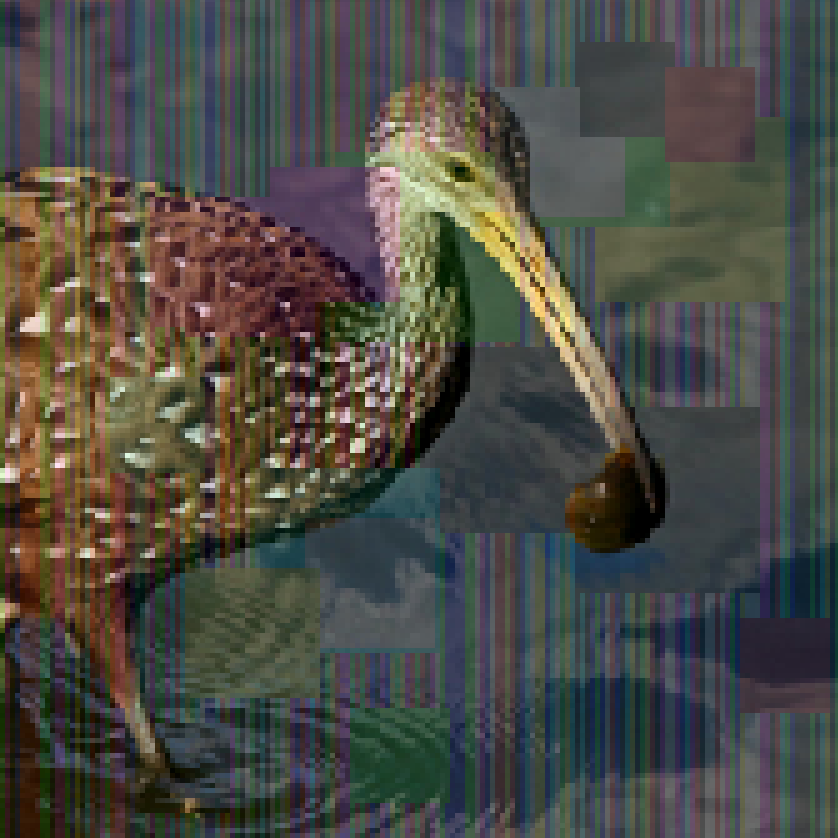} &
			\includegraphics[width=\newl]{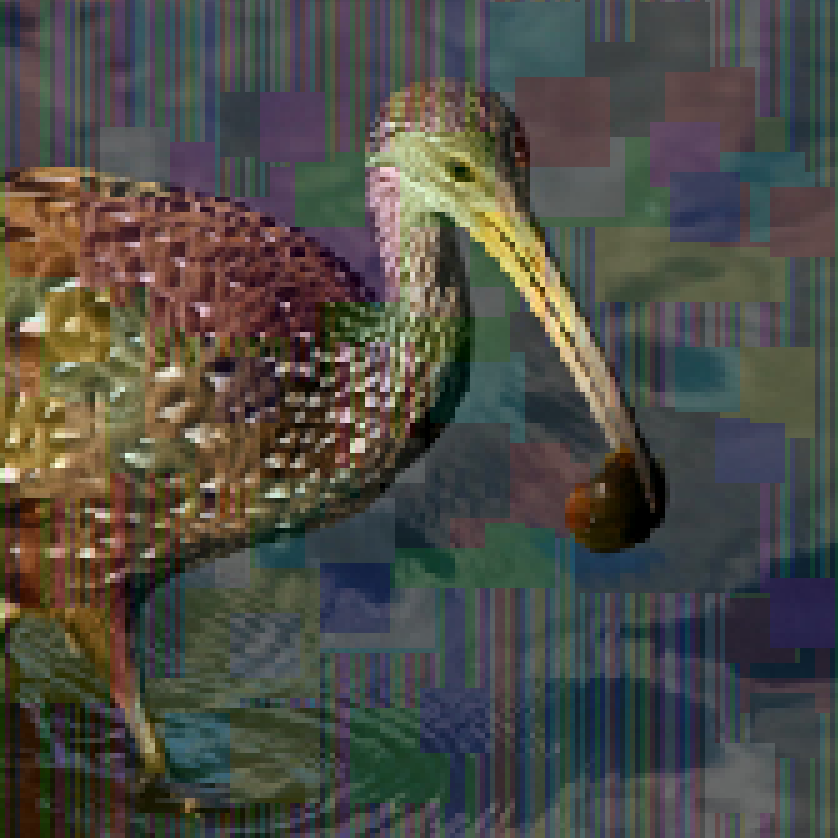} \\
			& 
			\includegraphics[width=\newl]{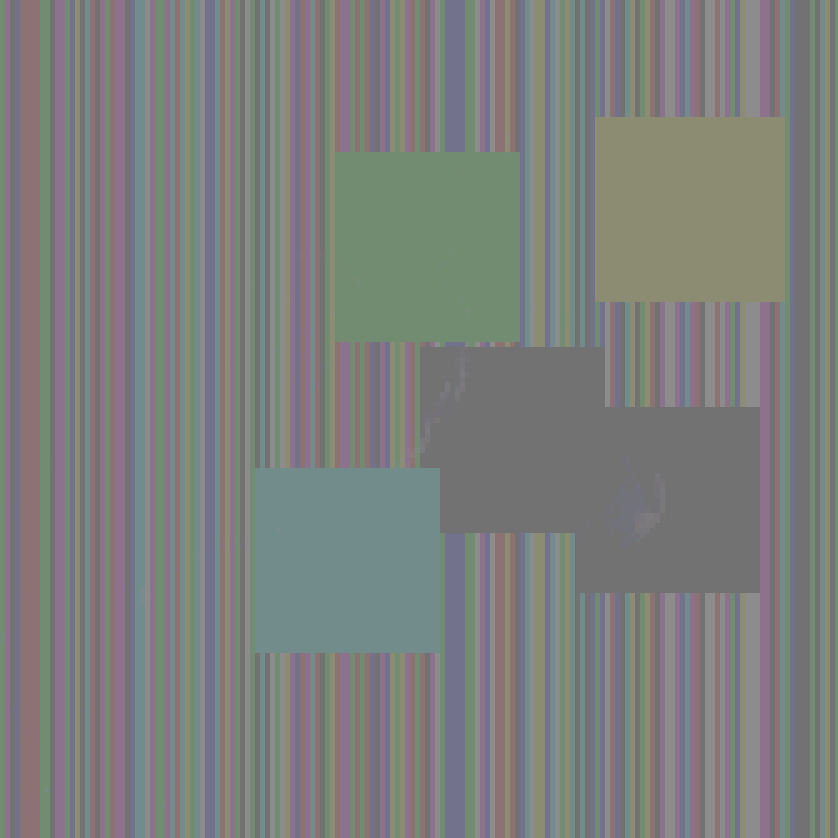} &
			\includegraphics[width=\newl]{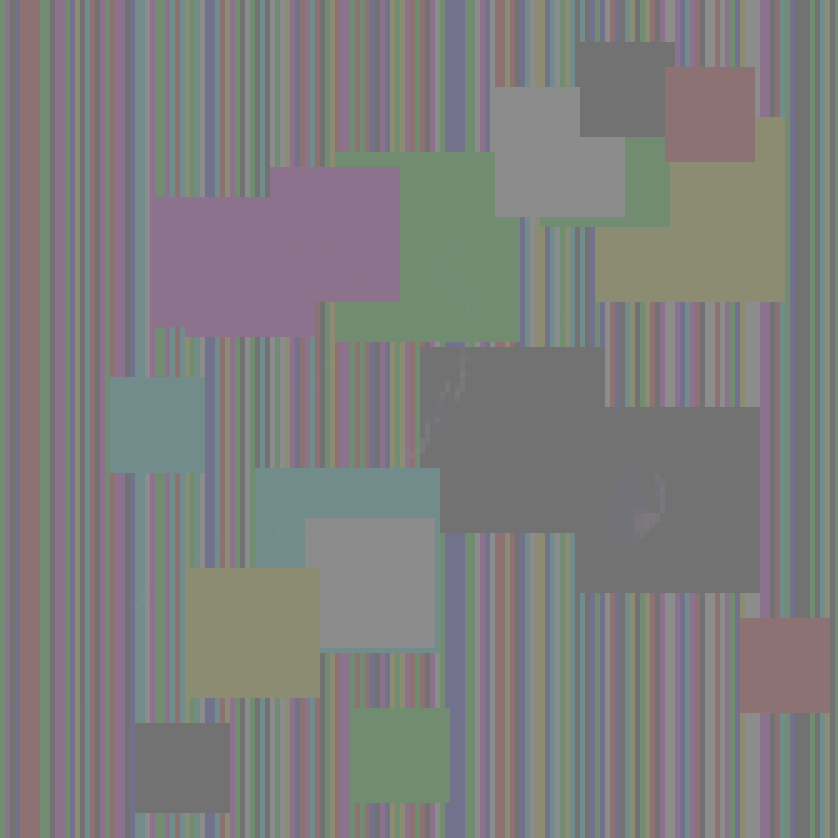} &
			\includegraphics[width=\newl]{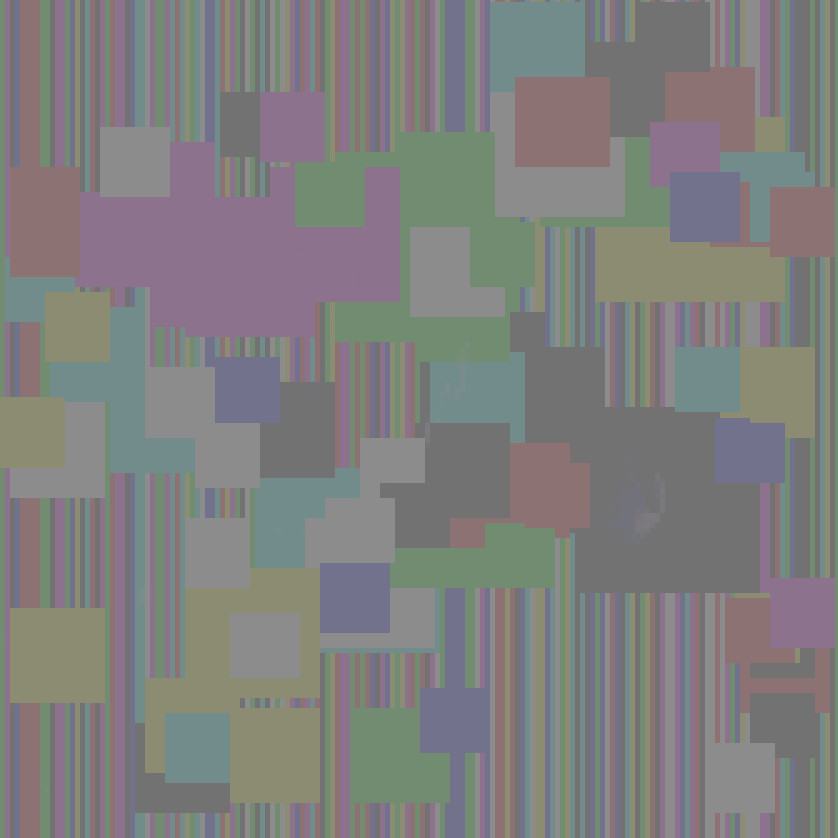} \\
			\ \\
			& Class: tennis ball & Class: tennis ball & Class: {\color{red} fig} \\ 
			\multirow[c]{2}{*}[14.5mm]{\includegraphics[width=\newlloss]{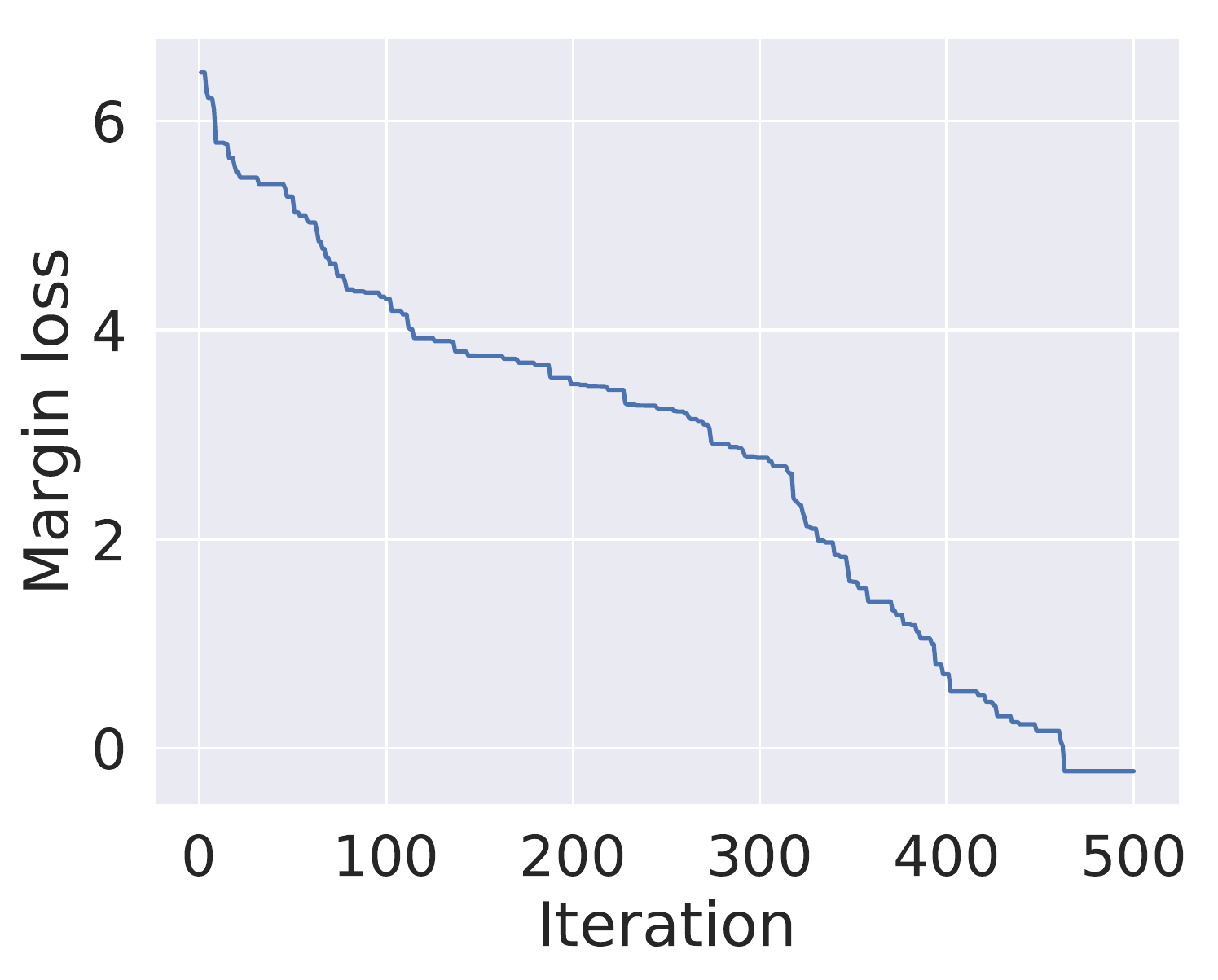}} & 
			\includegraphics[width=\newl]{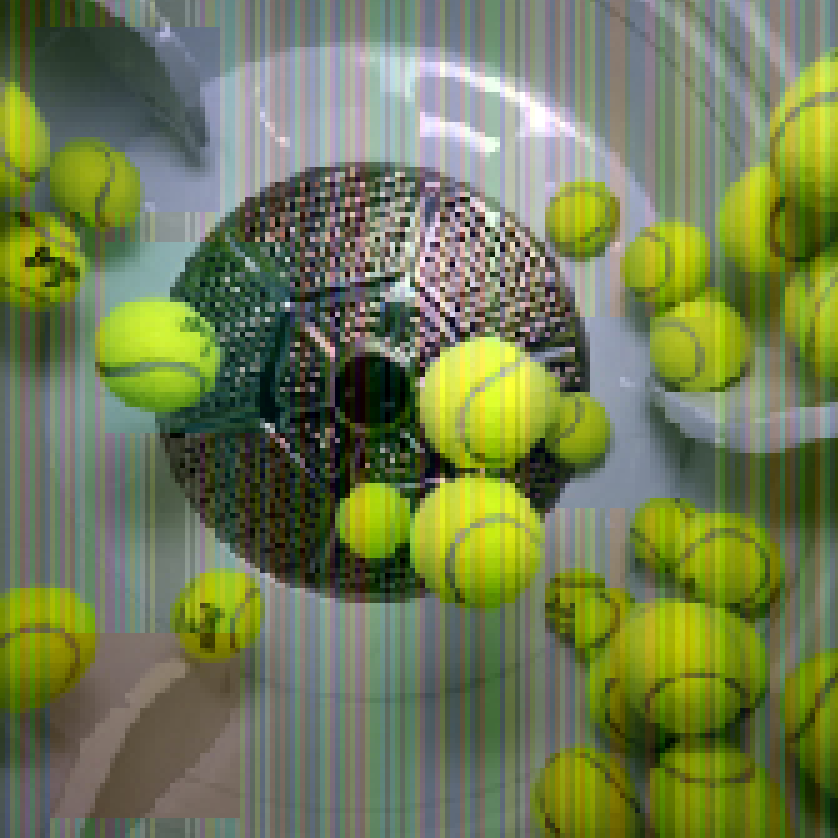} &
			\includegraphics[width=\newl]{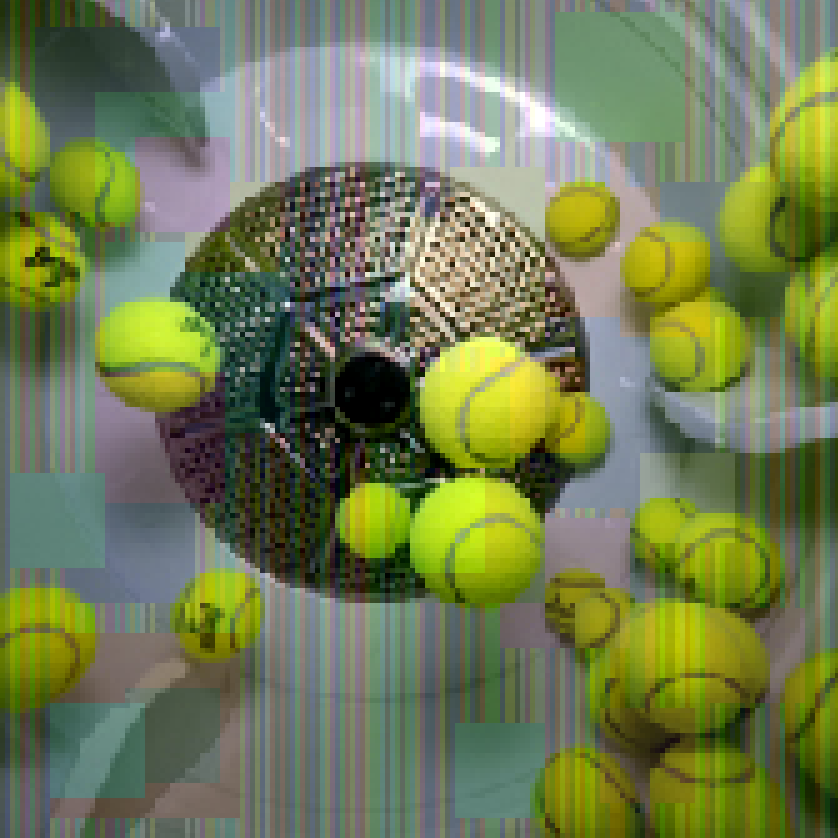} &
			\includegraphics[width=\newl]{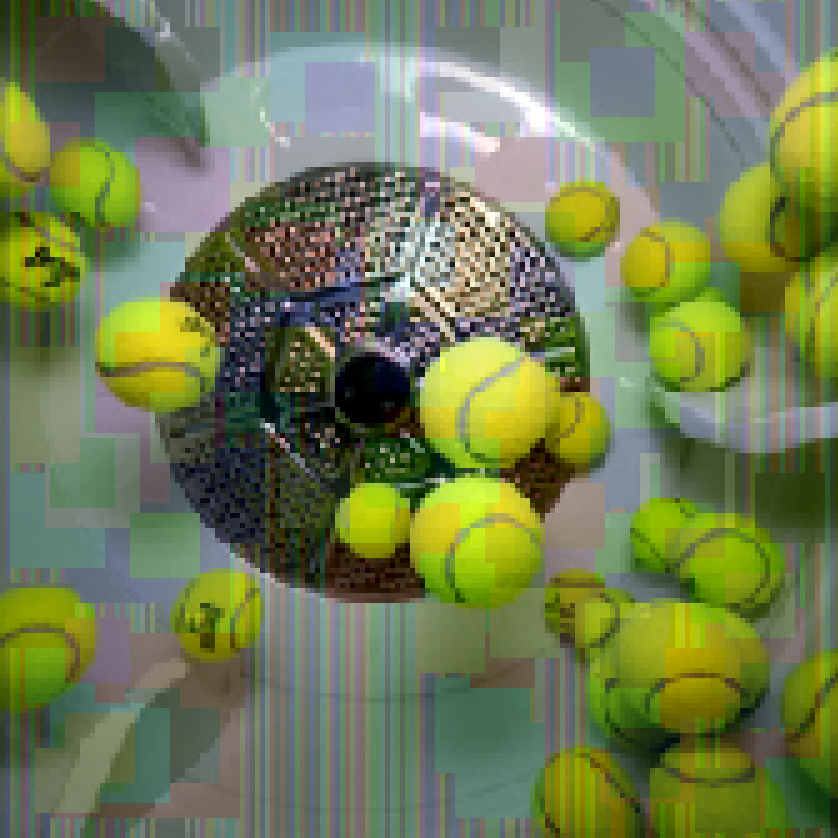} \\
			& 
			\includegraphics[width=\newl]{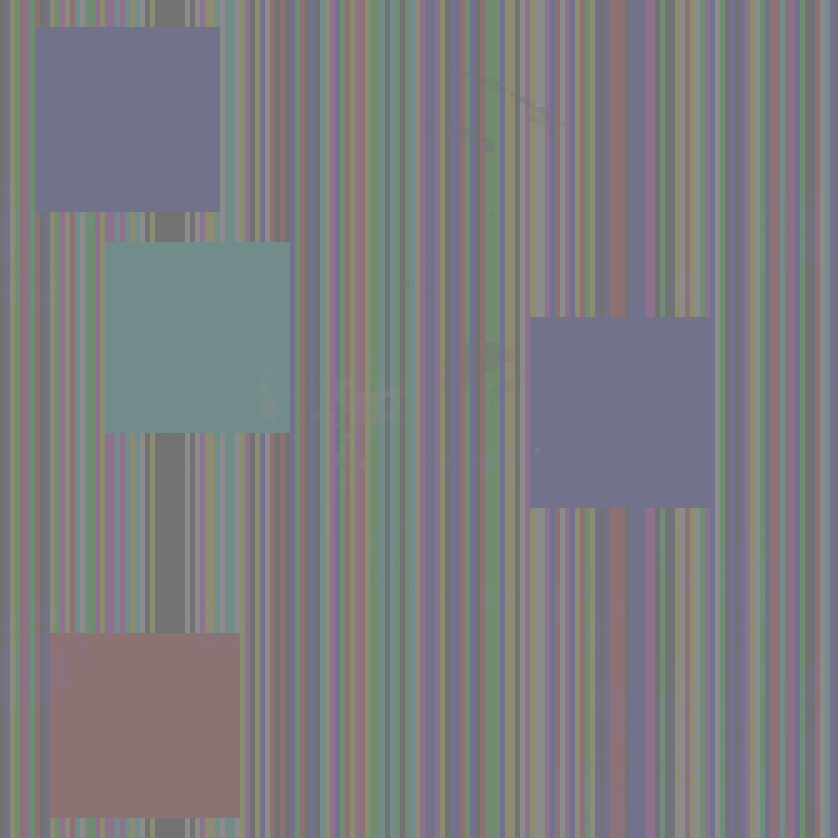} &
			\includegraphics[width=\newl]{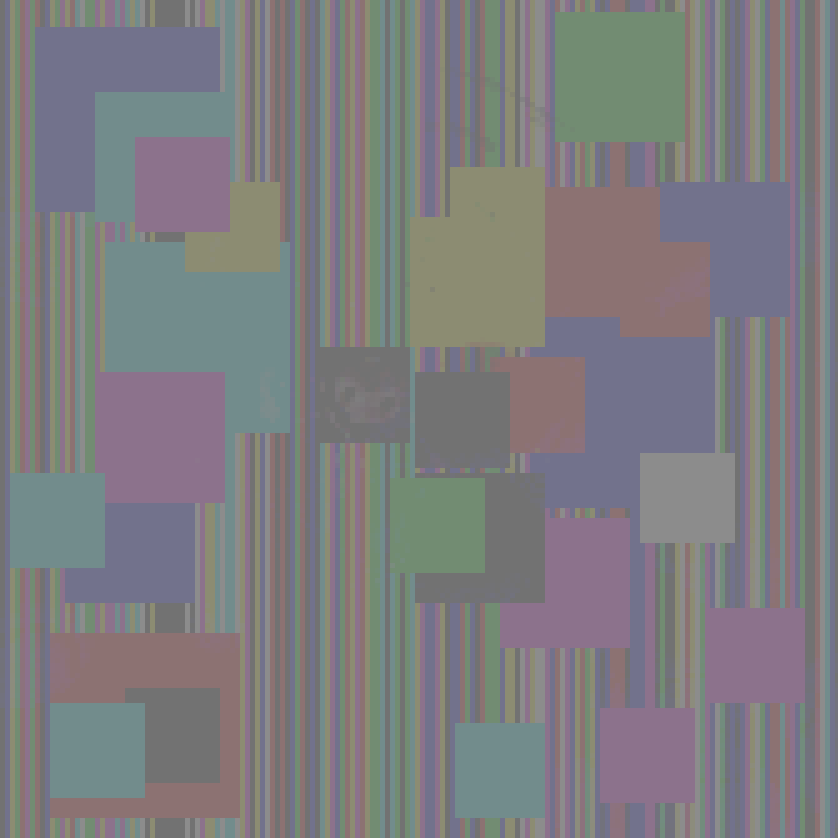} &
			\includegraphics[width=\newl]{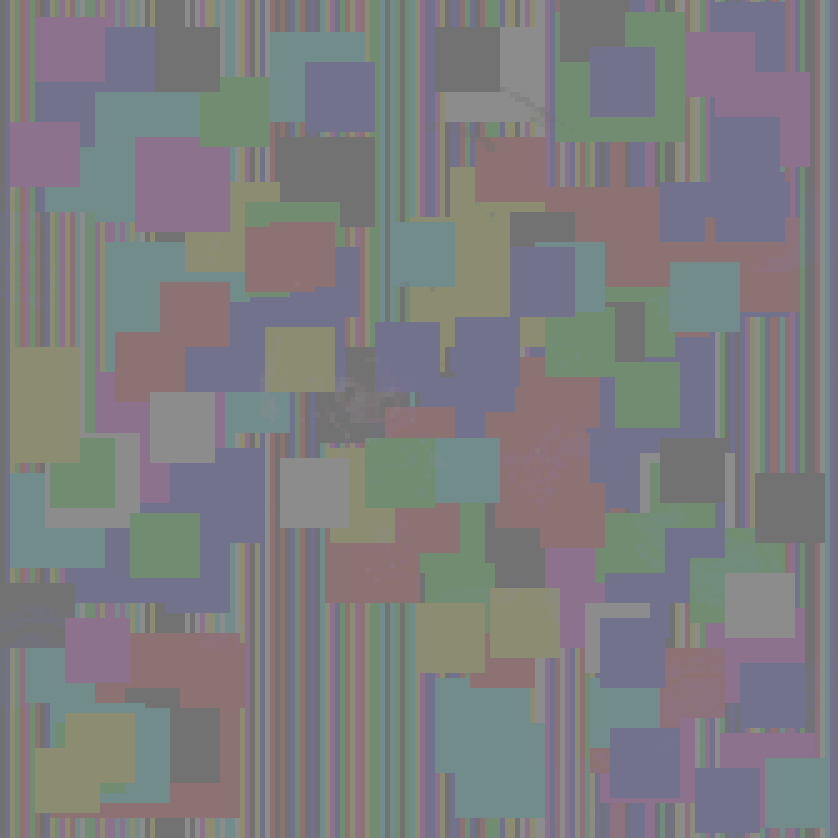} \\
			\ \\
			& Class: jinrikisha & Class: jinrikisha & Class: {\color{red} tricycle} \\ 
			\multirow[c]{2}{*}[14.5mm]{\includegraphics[width=\newlloss]{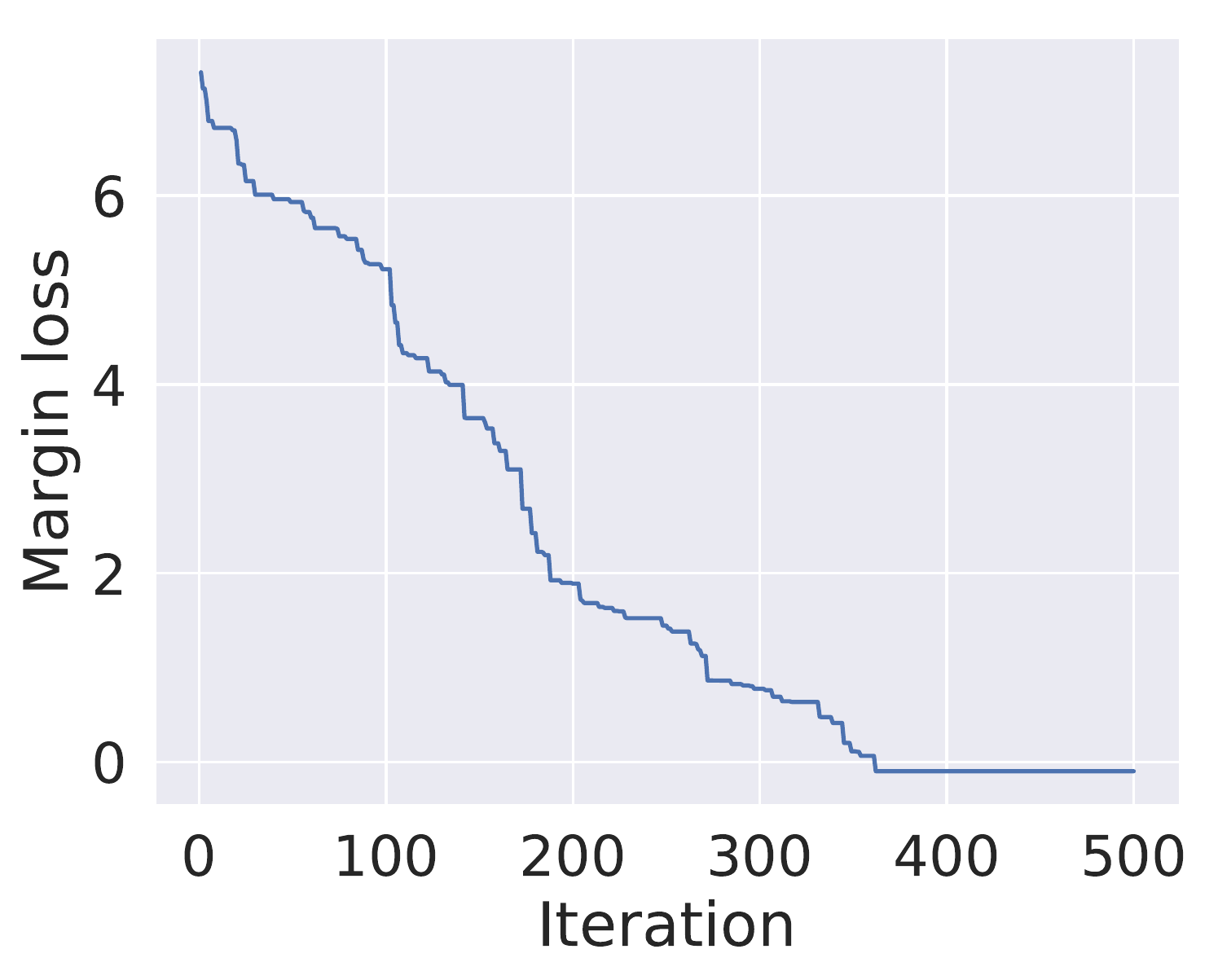}} & 
			\includegraphics[width=\newl]{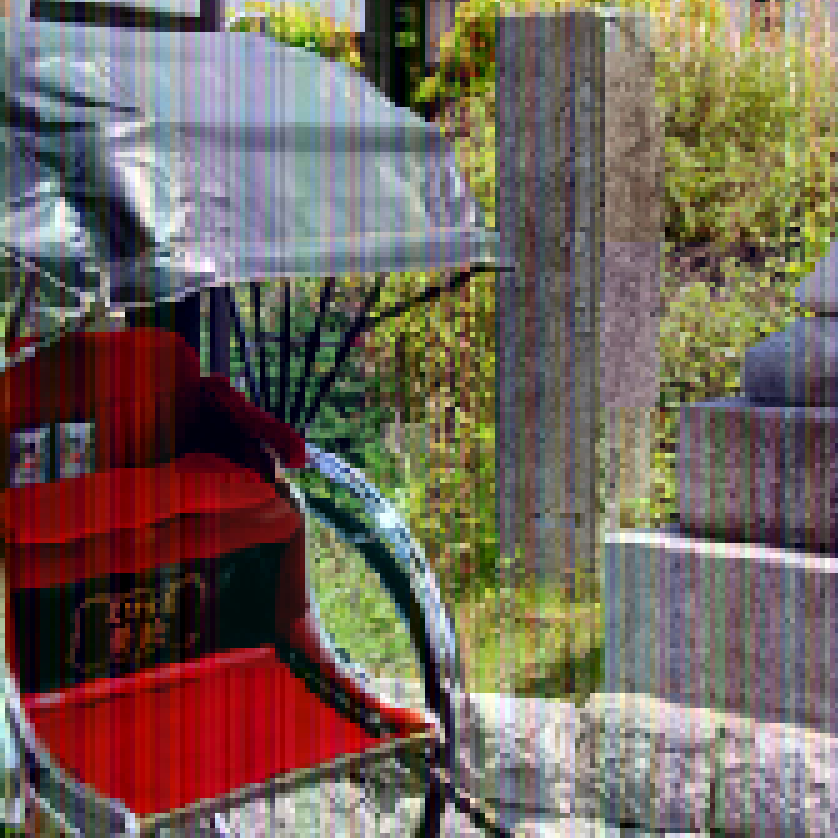} &
			\includegraphics[width=\newl]{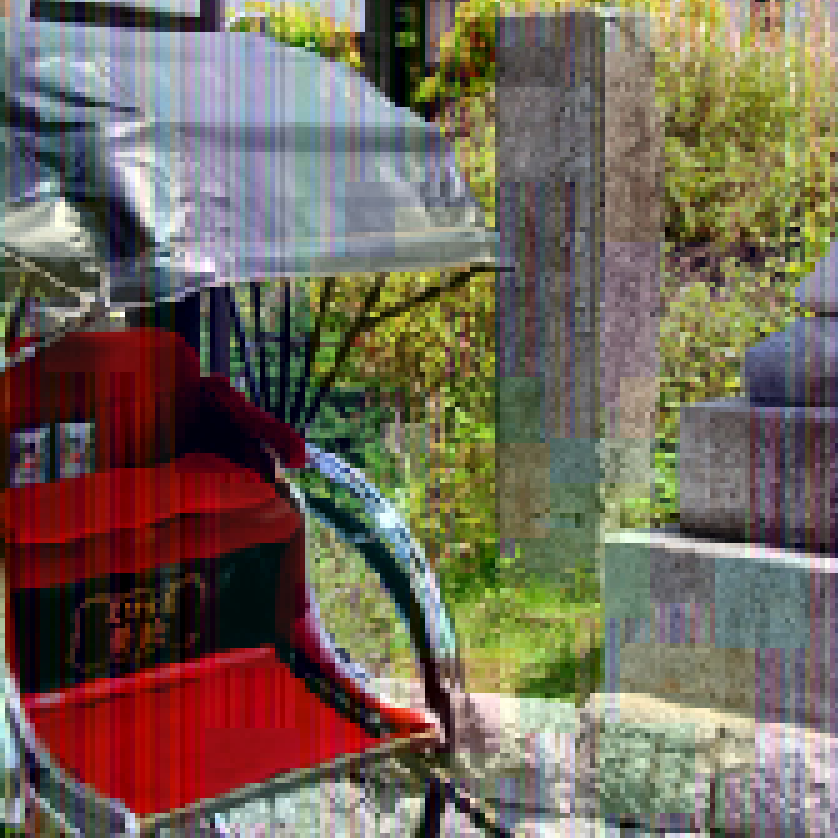} &
			\includegraphics[width=\newl]{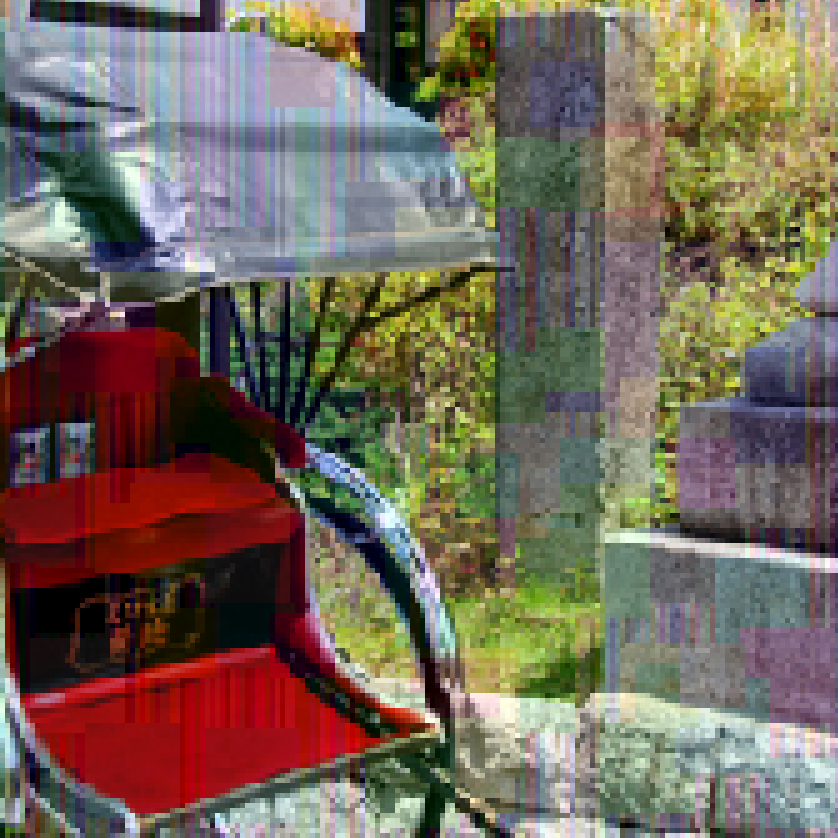} \\
			& 
			\includegraphics[width=\newl]{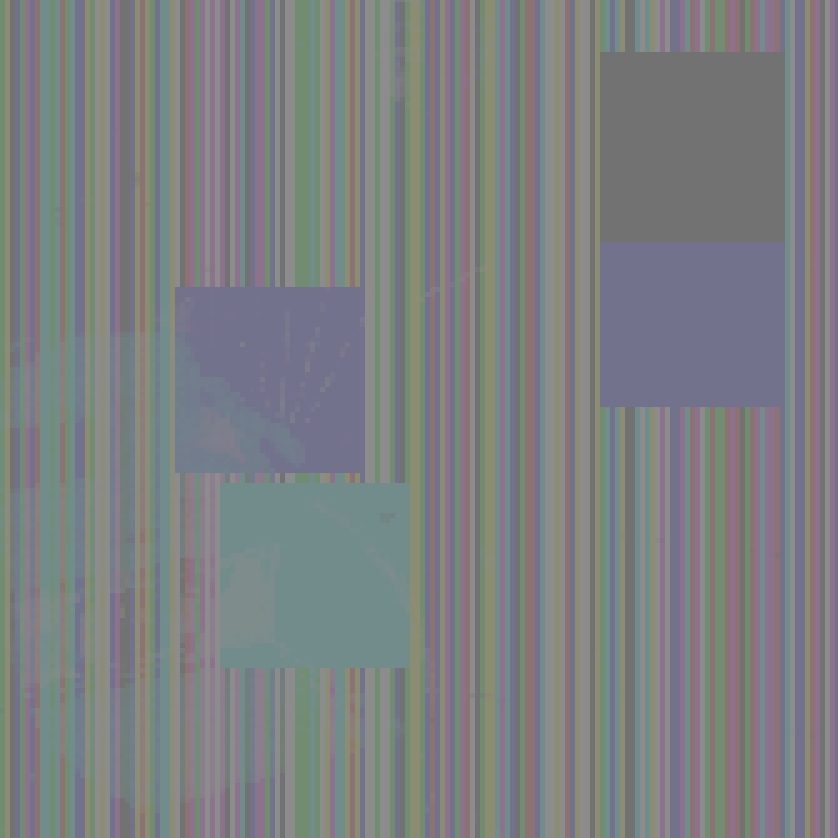} &
			\includegraphics[width=\newl]{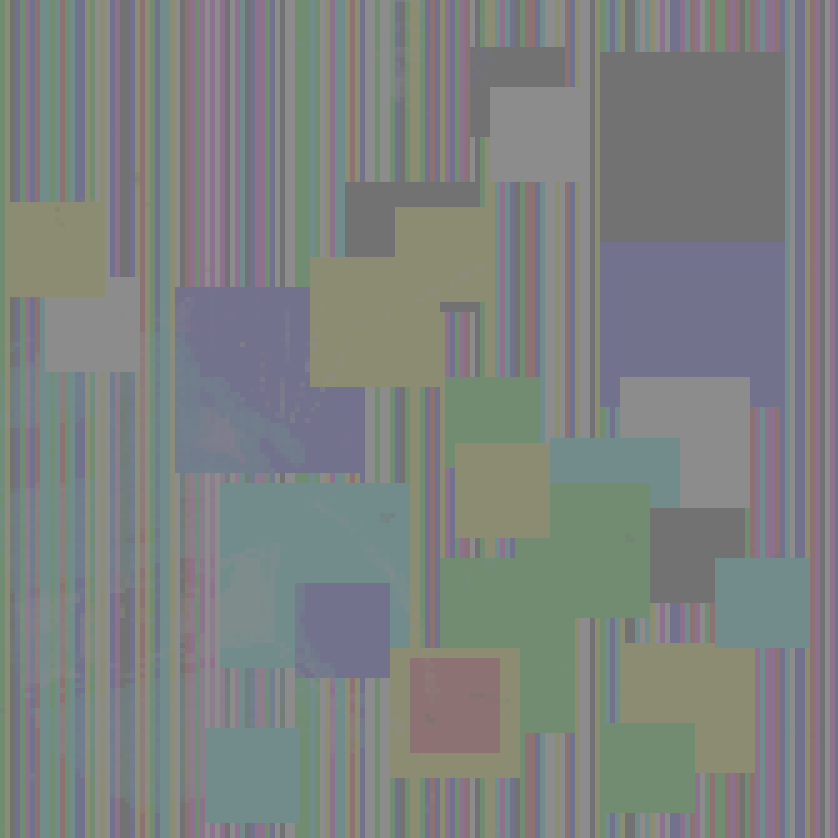} &
			\includegraphics[width=\newl]{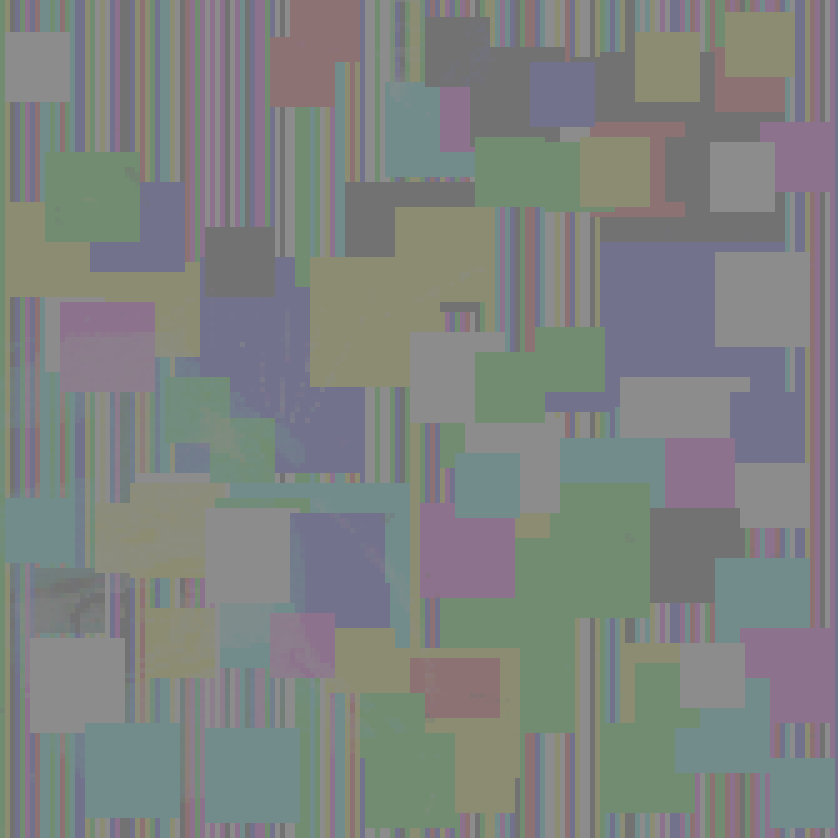} \\
			% \includegraphics[width=\newlloss]{near_miss-loss-img_id=39.pdf} & 
			% \includegraphics[width=\newl]{near_miss-images-iter=0-img_id=39_adv.pdf} &
			% \includegraphics[width=\newl]{near_miss-images-iter=1-img_id=39_adv.pdf} &
			% \includegraphics[width=\newl]{near_miss-images-iter=2-img_id=39_adv.pdf} \\
			%  & 
			% \includegraphics[width=\newl]{near_miss-images-iter=0-img_id=39_delta.pdf} &
			% \includegraphics[width=\newl]{near_miss-images-iter=1-img_id=39_delta.pdf} &
			% \includegraphics[width=\newl]{near_miss-images-iter=2-img_id=39_delta.pdf} \\
		\end{tabular}
		\caption{Visualization of adversarial examples for which the untargeted $l_\infty$ Square Attack requires more queries. We visualize adversarial examples and perturbations after 10, 100 and 500 iterations of the attack. The experiment is done on ImageNet using ResNet-50 for $\epsilon_\infty=0.05$. Note that a misclassification is achieved when the margin loss becomes negative} 
		\label{fig:high_query_adv}
	\end{figure}

	Here we provide more visualizations of adversarial perturbations generated by the untargeted Square Attack for $\epsilon=0.05$ on ImageNet. We analyze here the inputs that require more queries to be misclassified. We present the results in Fig.~\ref{fig:high_query_adv} where we plot adversarial examples after 10, 100 and 500 iterations of our attack.
	First, we note that a misclassification is achieved when the margin loss becomes negative. We can observe that the loss decreases gradually over iterations, and a single update only rarely leads to a significant decrease of the loss. As the attack progresses over iterations, the size of the squares is reduced according to our piecewise-constant schedule leading to more refined perturbations since the algorithm accumulates a larger number of square-shaped updates.
	% limpkin 99.95% -> dowitcher 46.21% (525 queries)
	% tennis_ball 99.83% -> fig 20.60% (463 queries)
	% jinrikisha 99.96% -> tricycle 37.26% (362 queries)
	
	\clearpage

	\subsection{Breaking the Post-averaging Defense} \label{sec:post-averaging_app}
	We investigate whether the $l_\infty$-robustness claims of \cite{lin2019bandlimiting} hold (as reported in \url{https://www.robust-ml.org/preprints/}).
	Their defense method is a randomized averaging method similar in spirit to \cite{cohen2019certified}. The difference is that \cite{lin2019bandlimiting} sample from the surfaces of several $d$-dimensional spheres instead of the Gaussian distribution, and they do not derive any robustness certificates, but rather measure robustness by the PGD attack. We use the hyperparameters specified in their code (K=15, R=6 on CIFAR-10 and K=15, R=30 on ImageNet). We show in Table \ref{tab:post-averaging_app} that the proposed defense can be broken by the $l_\infty$-Square Attack, which is able to reduce the robust accuracy suggested by PGD from 88.4\% to 15.8\% on CIFAR-10 and from 76.1\% to 0.4\% on ImageNet (we set $p=0.3$ for our attack). This again highlights that straightforward application of gradient-based white-box attacks may lead to inaccurate robustness estimation, and usage of the Square Attack can prevent false robustness claims.
	\ \\
	\begin{table}[h]
		\caption{$l_\infty$-robustness of the post-averaging randomized defense \cite{lin2019bandlimiting}. The Square Attack shows that these models are not robust}
		\label{tab:post-averaging_app}
		\centering
		{\setlength{\tabcolsep}{4pt}
			\small
			\begin{tabular}{c c|c|cc}
				\hline
				\multirow{2}{*}{$\epsilon_\infty$}&\multirow{2}{*}{\textbf{Dataset}} & \multicolumn{3}{c}{\textbf{Robust accuracy}} \\
				&& Clean & PGD & \textbf{Square} \\
				\hline
				\multirow{2}{*}{$8/255$}&CIFAR-10 & 92.6\% & 88.4\% & \textbf{15.8\%} \\
				&ImageNet & 77.3\% & 76.1\% & \textbf{0.4\%} \\
				\hline
		\end{tabular}}
	\end{table}

\end{document}